\pdfoutput=1

\documentclass[10pt,twocolumn,letterpaper]{article}

\usepackage[pagenumbers]{cvpr}      %

\usepackage{cuted} %

\usepackage{currfile} %

\usepackage{caption} %

\usepackage{microtype}

\usepackage{booktabs,multirow,amsmath,amssymb,graphicx,xspace,tikz,pgfplots}
\usepackage{pifont}
\usepackage{fontawesome5}

\usepackage{xr-hyper}
\makeatletter
\def\XR@@dURL#1#2#3#4#5\\{%
  \if\relax\detokenize{#3#4}\relax
    \unexpanded{{#1}{#2}}%
  \else
    \unexpanded{{#1}{#2}{#3}{#4}}{\XR@URL}%
  \fi
}
\makeatother
\usetikzlibrary{arrows.meta,positioning,fit,calc,shapes.misc,shadows.blur}
\usepgfplotslibrary{groupplots}
\pgfplotsset{compat=1.18}
\definecolor{cA}{HTML}{2E5EAA}   %
\definecolor{cB}{HTML}{23907F}   %
\definecolor{cC}{HTML}{D9792B}   %
\definecolor{cGrid}{HTML}{D5DAE1}
\pgfplotsset{paper/.style={
  font=\sffamily\scriptsize,
  axis line style={black!60, line width=0.5pt}, tick style={black!60, line width=0.5pt}, major tick length=2pt,
  tick label style={font=\sffamily\scriptsize}, label style={font=\sffamily\scriptsize},
  /pgf/number format/assume math mode=true,
  title style={font=\sffamily\scriptsize\bfseries, yshift=-3pt},
  legend style={font=\sffamily\tiny, draw=none, fill=white, fill opacity=0.85, text opacity=1,
                inner sep=1pt, row sep=-1pt},
  legend cell align=left,
  cycle list={{cA,mark=*},{cC,mark=square*},{cB,mark=triangle*},{black!55,mark=diamond*}},
  grid=major, grid style={cGrid, line width=0.3pt},
  every axis plot/.append style={line width=1pt},
  scaled y ticks=false,
}}
\definecolor{srv}{HTML}{2E5EAA}
\definecolor{srvL}{HTML}{EAF0F9}
\definecolor{cli}{HTML}{D9792B}
\definecolor{cliL}{HTML}{FCF1E6}
\definecolor{pkt}{HTML}{23907F}
\definecolor{pktL}{HTML}{E2F2EF}
\definecolor{ses}{HTML}{7A5BA6}
\definecolor{sesL}{HTML}{F1ECF8}
\definecolor{ink}{HTML}{262A30}
\definecolor{mute}{HTML}{7D8590}
\definecolor{hair}{HTML}{D5DAE1}
\definecolor{stMat}{HTML}{93BDE6}
\definecolor{stRec}{HTML}{2E5EAA}
\definecolor{stHand}{HTML}{AEB8C6}
\definecolor{stEnc}{HTML}{23907F}
\definecolor{stDec}{HTML}{F4C9A2}
\definecolor{stCor}{HTML}{E8995A}
\definecolor{stHead}{HTML}{BF5E1C}
\definecolor{hot}{HTML}{C8323C}
\definecolor{tokkeep}{RGB}{235,150,160}
\definecolor{tokdrop}{RGB}{40,120,200}
\newcommand{\hcsmall}{\fontsize{6.5}{7.5}\selectfont}
\tikzset{
  hc/.style={font=\sffamily\scriptsize, text=ink, line cap=round, line join=round,
             >={Stealth[length=3.6pt,width=3pt]}},
  hcarrow/.style={->, line width=0.7pt, draw=ink!75},
  hcpanel/.style={rounded corners=4pt, fill=#1, draw=none},
  hctitle/.style={font=\sffamily\scriptsize\bfseries, text=ink, inner sep=1pt},
  hcnote/.style={font=\sffamily\hcsmall, text=mute, inner sep=1pt},
  hcbox/.style={rounded corners=2.5pt, draw=#1, fill=white, line width=0.6pt, align=center,
                inner sep=2.5pt, font=\sffamily\scriptsize, text=ink},
  hctag/.style={font=\sffamily\bfseries\hcsmall, text=white, fill=#1, rounded corners=1.5pt,
                inner sep=1.4pt},
}
\pgfplotsset{hcplot/.style={
  font=\sffamily\scriptsize,
  axis line style={ink!55, line width=0.5pt, -}, tick style={ink!55, line width=0.5pt},
  major tick length=2pt, axis lines=left,
  tick label style={font=\sffamily\scriptsize, text=ink},
  /pgf/number format/assume math mode=true,
  label style={font=\sffamily\scriptsize, text=ink},
  title style={font=\sffamily\scriptsize\bfseries, text=ink},
  grid=major, grid style={hair, line width=0.35pt},
  every axis plot/.append style={line width=1pt},
  scaled y ticks=false, scaled x ticks=false,
}}
\pgfdeclareplotmark{hcstar}{%
  \pgfpathmoveto{\pgfqpointpolar{90}{1.3\pgfplotmarksize}}%
  \pgfpathlineto{\pgfqpointpolar{126}{0.55\pgfplotmarksize}}%
  \pgfpathlineto{\pgfqpointpolar{162}{1.3\pgfplotmarksize}}%
  \pgfpathlineto{\pgfqpointpolar{198}{0.55\pgfplotmarksize}}%
  \pgfpathlineto{\pgfqpointpolar{234}{1.3\pgfplotmarksize}}%
  \pgfpathlineto{\pgfqpointpolar{270}{0.55\pgfplotmarksize}}%
  \pgfpathlineto{\pgfqpointpolar{306}{1.3\pgfplotmarksize}}%
  \pgfpathlineto{\pgfqpointpolar{342}{0.55\pgfplotmarksize}}%
  \pgfpathlineto{\pgfqpointpolar{18}{1.3\pgfplotmarksize}}%
  \pgfpathlineto{\pgfqpointpolar{54}{0.55\pgfplotmarksize}}%
  \pgfpathclose\pgfusepathqfillstroke}

\newcommand{\sys}{HoloSplat\xspace}

\newcommand{\ProKB}{208.2}

\newcommand{\CompressionRatio}{22.8}
\newcommand{\UncDelta}{+0.000}
\newcommand{\ProDelta}{1.050}
\newcommand{\ServerDrop}{23.6}
\newcommand{\ByteDrop}{23.5}
\newcommand{\ServerLoss}{+0.012}

\newcommand{\ClientLoss}{+0.002}

\newcommand{\FuseDelta}{0.002}

\newcommand{\StateProKB}{201}

\newcommand{\HiFiGainMin}{1.34}
\newcommand{\HiFiGainMax}{1.50}
\newcommand{\HiFiRatioMin}{11.7}
\newcommand{\HiFiRatioMax}{39.5}

\newcommand{\AHQGainMin}{1.28}
\newcommand{\AHQGainMax}{1.42}
\newcommand{\AHQRatioMin}{16.9}
\newcommand{\AHQRatioMax}{59.8}

\newcommand{\XHGainMin}{0.54}
\newcommand{\XHGainMax}{1.08}
\newcommand{\XHRatioMin}{11.7}
\newcommand{\XHRatioMax}{39.3}
\newcommand{\ProdSceneCount}{16}

\newcommand{\SingleFPSMean}{8.0}

\newcommand{\FourFPSMean}{32.2}

\newcommand{\SixFPSMean}{51.3}
\newcommand{\ScalingEffMin}{97}
\newcommand{\ScalingEffMax}{115}
\newcommand{\ThirtyFPSGPUs}{4}
\newcommand{\ThirtyFPSGPUsWord}{four}
\newcommand{\ThirtyFPSMean}{32.2}

\newcommand{\GeometryKB}{35.0}
\newcommand{\ControlSPZOverUltra}{3.38}

\newcommand{\PruneBDPSNR}{0.12}
\newcommand{\PruneBDRate}{25.1}
\newcommand{\TeaserKB}{323.0}
\newcommand{\TeaserKept}{8,038}
\newcommand{\TeaserTotal}{13,324}
\newcommand{\TeaserGaussians}{102,886}
\newcommand{\TeaserDropped}{5,286}
\newcommand{\TeaserHeadGaussians}{128,608}
\newcommand{\TeaserHeaderKB}{1.4}

\newcommand{\DriftLastWithin}{0.26}

\newcommand{\LDDeltaLite}{+0.822}
\newcommand{\LDRecoveredLite}{66.7}

\newcommand{\LDDeltaPro}{+0.483}
\newcommand{\LDRecoveredPro}{80.4}

\newcommand{\LDDeltaUltra}{+0.191}
\newcommand{\LDRecoveredUltra}{68.5}

\newcommand{\NSCHiFiDeployedIncrementPro}{+1.419}

\newcommand{\NSCAHQDeployedIncrementPro}{+1.436}

\newcommand{\NSCKB}{44.5}
\newcommand{\NSCParams}{20{,}480}
\newcommand{\NSCPayloadKB}{41.0}
\newcommand{\NSCShareOfPro}{21}
\newcommand{\SessionStateProKB}{840}

\newcommand{\NSCFullHeadRatio}{814}
\newcommand{\NSCVsAdapterHiFi}{95.7}
\newcommand{\NSCColorHiFiPro}{+0.076}
\newcommand{\NSCVsAdapterAHQ}{88.9}
\newcommand{\NSCColorAHQPro}{+0.119}

\newcommand{\NSCGapProStreamMax}{0.12}
\newcommand{\NSCGapProPlainMax}{0.60}
\newcommand{\NSCHiFiLiteOverUnc}{+1.327}
\newcommand{\NSCAHQLiteOverUnc}{+1.282}
\newcommand{\NSCCodecCostMax}{0.76}
\newcommand{\NSCHyperCodecCostMax}{0.13}

\newcommand{\LDLaptopClientLite}{21.7}
\newcommand{\LDLaptopDecodePro}{0.76}
\newcommand{\LDLaptopOverheadPro}{5.55}
\newcommand{\LDLaptopHeadPro}{15.80}

\newcommand{\LDLaptopClientPro}{22.1}

\newcommand{\LDLaptopClientUltra}{22.8}
\newcommand{\ClientRealTokensMean}{5{,}628}
\newcommand{\ClientRealScenes}{3}
\newcommand{\HyperFoldMs}{525.3}

\newcommand{\LDLaptopLoRAFold}{0.36}
\newcommand{\HyperUpdateKB}{593.8}
\newcommand{\HyperTrainScenes}{160}
\newcommand{\HyperValActors}{three}
\newcommand{\SessionReadySeconds}{24}
\newcommand{\NSCLRSessions}{48}
\newcommand{\NSCLRLossExcessMin}{5}
\newcommand{\NSCLRLossExcessMax}{17}
\newcommand{\GoLiveRenderMinMs}{204}
\newcommand{\GoLiveRenderMaxMs}{304}

\newcommand{\GoLiveServerMinMs}{171}
\newcommand{\GoLiveServerMaxMs}{261}
\newcommand{\GoLiveWiFiMinMs}{10}
\newcommand{\GoLiveWiFiMaxMs}{14}
\newcommand{\GoLiveClientMinMs}{21}
\newcommand{\GoLiveClientMaxMs}{33}
\newcommand{\GoLiveGsplatMinMs}{1.8}
\newcommand{\GoLiveGsplatMaxMs}{2.0}
\newcommand{\LaptopFPCheckFrames}{1{,}488}
\newcommand{\LaptopFPExplainedMax}{$8\times10^{-6}$}
\newcommand{\HyperFitSecondsMin}{2.5}
\newcommand{\HyperFitSecondsMax}{4.2}
\newcommand{\HyperGenericRecoveredLite}{31.9}
\newcommand{\HyperSessionDeployedLite}{+0.296}
\newcommand{\HyperGenericRecoveredPro}{38.7}
\newcommand{\HyperSessionDeployedPro}{+0.193}
\newcommand{\HyperGenericRecoveredUltra}{24.1}
\newcommand{\HyperSessionDeployedUltra}{+0.098}
\newcommand{\MechNeighGainLitedB}{+0.107}
\newcommand{\MechPermGainLitedB}{0.109}
\newcommand{\MechMetricExpGainLite}{2.0}
\newcommand{\MechMetricLPIPSDeltaLite}{0.0024}
\newcommand{\MechNeighGainProdB}{+0.024}
\newcommand{\MechPermGainProdB}{0.019}
\newcommand{\MechMetricExpGainPro}{2.6}
\newcommand{\MechMetricLPIPSDeltaPro}{0.0015}
\newcommand{\NSCProFullMLPHeldDrop}{0.13}
\newcommand{\NSCProFullMLPTempDrop}{0.35}

\newcommand{\NSCProNoLPIPSWorse}{0.012}

\newcommand{\NSCDNADeltaPlainLite}{+1.890}
\newcommand{\NSCDNADeltaLearnedLite}{+1.068}

\newcommand{\NSCDNADeltaPlainPro}{+1.636}
\newcommand{\NSCDNADeltaLearnedPro}{+1.154}

\newcommand{\NSCDNADeltaPlainUltra}{+1.369}
\newcommand{\NSCDNADeltaLearnedUltra}{+1.178}

\newcommand{\NSCBudgetFrames}{30}
\newcommand{\NSCBudgetSteps}{300}
\newcommand{\NSCBudgetOneFrameSteps}{300}
\newcommand{\NSCBudgetOneFrameShareHiFi}{80}
\newcommand{\NSCBudgetOneFrameShareAHQ}{93}

\newcommand{\CmpUncKB}{3630.7}

\newcommand{\CmpMinCodecOverPro}{3.47}

\newcommand{\BudgetFormatMinMbps}{173}

\newcommand{\CmpSameCalMaxGainRounded}{0.14}

\newcommand{\CmpSameCalMinBytesOverPro}{3.47}
\newcommand{\CmpSameCalMinBytesOverUltra}{2.24}

\newcommand{\CmpDynHiCoMSparseBelowLite}{5.31--10.14}
\newcommand{\CmpDynQUEENSparseBelowLite}{7.35--12.62}
\newcommand{\CmpDynGSStreamSparseBelowLite}{0.62--4.79}

\newcommand{\CmpDynHiCoMDenseLiteGainMin}{3.81}
\newcommand{\CmpDynHiCoMDenseLiteGainMax}{8.45}
\newcommand{\CmpDynQUEENDenseBelowLite}{1.96--8.42}

\newcommand{\CmpDynGSStreamDenseMeanBelowLite}{0.37}

\newcommand{\CmpDynGSStreamDenseBytesOverUltra}{17.6--31.8}
\newcommand{\CmpDynPairedLPIPSGain}{0.211}

\newcommand{\CmpDynFirstS}{50--611}
\newcommand{\CmpDynLaterS}{1.2--23.4}

\newcommand{\BudgetDynMinMbps}{154}
\newcommand{\CmpDynMedianOverLite}{62}
\newcommand{\CmpDynMedianOverPro}{30}
\newcommand{\CmpDynMedianOverUltra}{19}
\newcommand{\CmpDynSpeedupMean}{16}

\newcommand{\CmpDynOursLaterS}{0.13--0.18}

\newcommand{\CmpDynSparseAttempted}{16}

\newcommand{\CmpDynBaselineRuns}{96}

\newcommand{\CmpDynOursLitePSNR}{25.95}

\newcommand{\CmpDynOursProPSNR}{26.30}

\newcommand{\CmpDynOursUltraPSNR}{26.38}

\newcommand{\CmpDynFourCProGain}{1.62}
\newcommand{\CmpDynFourCUltraGain}{1.70}
\newcommand{\CmpDynFourCLiteGain}{1.27}
\newcommand{\CmpDynFourCOverPro}{58}
\newcommand{\CmpDynFourCGaussiansM}{0.4--3.6}
\newcommand{\CmpIdleFitMinS}{0.87}
\newcommand{\CmpSparseTotal}{16}
\newcommand{\CmpDynHullSparse}{16}
\newcommand{\CmpDynHullDense}{4}

\newcommand{\CmpSparseMoreTotal}{13}

\newcommand{\CmpSparseMorePairs}{39}

\newcommand{\CmpSparseLiteDecoderGainMin}{0.696}
\newcommand{\CmpSparseLiteDecoderGainMax}{3.273}
\newcommand{\CmpSparseLiteGainMin}{0.62}
\newcommand{\CmpSparseLiteGainMax}{12.56}
\newcommand{\CmpLitFastestS}{2.67}

\newcommand{\HumanSeqFitMin}{6\,s}
\newcommand{\HumanSeqFitMax}{12\,min}
\newcommand{\HumanAvatarDays}{2}

\newcommand{\CmpCodecSplatSendMs}{276}
\newcommand{\CmpCodecSplatRecvMs}{199}
\newcommand{\CmpOursClientSameGPUMs}{3.9}
\newcommand{\CmpCodecSplatRecvRatio}{51}

\newcommand{\CmpHiReFFSameGPUFPS}{4.0}
\newcommand{\CmpCodecSplatRecvOverBudget}{6.0}

\newcommand{\CmpCodecSplatTopPSNR}{23.1}
\newcommand{\CmpCodecSplatTopPSNRFine}{23.14}
\newcommand{\CmpCodecSplatLiteGap}{3.7}
\newcommand{\CmpCodecSplatProGap}{4.1}
\newcommand{\CmpSharedIdealOneMax}{0.81}
\newcommand{\CmpSharedIdealFourMax}{3.25}

\newcommand{\CmpIdleRatio}{17.0}
\newcommand{\CmpIdleClosestMethod}{3DGStream}

\newcommand{\CmpThirtyGPUsMin}{27}

\newcommand{\CmpExtFullFormatSettings}{24}

\newcommand{\CmpExtHiFiMinCodecOverUltraAmort}{2.02}

\newcommand{\CmpExtAHQMinCodecOverUltraAmort}{2.01}

\newcommand{\ServerPairedRatio}{1.002}
\newcommand{\CmpReconForwardMs}{85--126}
\newcommand{\LiteKBCmp}{94.8}
\newcommand{\ProKBCmp}{208.2}
\newcommand{\UltraKBCmp}{321.9}
\newcommand{\BudgetProMbps}{50}

\newcommand{\BudgetVideoMinMbps}{20}
\newcommand{\BudgetVideoMaxMbps}{40}

\newcommand{\StageMattingMs}{45}
\newcommand{\StageReconMs}{109}
\newcommand{\StageHandMs}{10}
\newcommand{\StageEncodeMs}{7.5}
\newcommand{\StageTotalMs}{172}

\newcommand{\GanttFullPct}{99.8}

\newcommand{\GanttGPUsWord}{four}
\newcommand{\GanttWorkersWord}{twelve}

\newcommand{\CutAttrBestPSNR}{23.30}
\newcommand{\CutAttrBestKB}{653}
\newcommand{\CutLiteOverAttrBest}{1.65}
\newcommand{\CutAttrBestOverLiteBytes}{6.9}
\newcommand{\CutTOverAttrMin}{2.8}
\newcommand{\CutTOverAttrMax}{10.3}
\newcommand{\CutVarTokens}{81.9}
\newcommand{\CutVarAttrMin}{97.9}
\newcommand{\CutVarAttrMax}{99.9}
\newcommand{\CutHGainMin}{0.04}
\newcommand{\CutHGainMax}{0.12}

\newcommand{\MatRVMProLoss}{1.84}

\newcommand{\ReviewNoNSCUniformZstdSaving}{86}

\newcommand{\ReviewNoNSCDracoSaving}{75}

\newcommand{\ReviewNoNSCFCGSSaving}{72}

\newcommand{\ReviewNoNSCSavingMin}{72}
\newcommand{\ReviewNoNSCSavingMax}{86}

\newcommand{\ReviewNSCUniformZstdSaving}{85}

\newcommand{\ReviewNSCDracoSaving}{67}

\newcommand{\ReviewNSCFCGSSaving}{82}

\newcommand{\ReviewNSCSavingMin}{67}
\newcommand{\ReviewNSCSavingMax}{85}

\newcommand{\LongSeqCount}{10}
\newcommand{\LongMaxFrames}{195}

\newcommand{\LongNoRefreshLastFirstNine}{+0.04}

\newcommand{\LongRefreshPCAOneTwentyDeltaNine}{$-$0.07}

\newcommand{\LongRefreshPCAOneTwentyAddedBytesNine}{1.4}
\newcommand{\LongLoRAOneTwentyDeltaFive}{+0.00}

\newcommand{\LongNoRefreshSlopeMin}{$-$0.11}
\newcommand{\LongNoRefreshSlopeMax}{+0.27}

\newcommand{\LongNSCDeltaLastHiFi}{+1.19}

\newcommand{\LongNSCDeltaLastAHQ}{+1.69}

\newcommand{\DSBypassPSNR}{15.28}
\newcommand{\DSTBestPSNR}{15.64}
\newcommand{\DSTBestKB}{310.1}

\newcommand{\DSABestPSNR}{15.39}

\newcommand{\DSABestOverTBestBytes}{9.6}

\newcommand{\DSLearnedTopPSNR}{23.14}

\newcommand{\DSScenes}{16}

\newcommand{\VidBestPSNR}{16.36}

\newcommand{\VidBestMbps}{185.69}

\newcommand{\VidCPUMinMs}{112.79}
\newcommand{\VidCPUMaxMs}{143.86}
\newcommand{\VidExcludedScenes}{8}
\newcommand{\VidDiagnosticBestPSNR}{23.65}

\definecolor{cvprblue}{rgb}{0.21,0.49,0.74}
\usepackage[pagebackref,breaklinks,colorlinks,allcolors=cvprblue]{hyperref}

\def\confName{CVPR}
\def\confYear{2027}

\title{\sys: Real-Time Token Streaming for Feed-Forward Human Reconstruction}

\author{Chaofan Li$^{*}$ \quad Qihan Ding$^{*}$ \quad Shihan Wang \quad Zhehao Shen \quad Boyu Shen\\
Quanyu Chen \quad Haoyu Wu \quad Lan Xu$^{\dagger}$ \quad Xin Lou$^{\dagger}$\\[0.4em]
ShanghaiTech University\\
{\tt\small \{lichf2025, dingqh2025, wangshh2023, shenzhh2025, shenby2024,}\\
{\tt\small chenqy2026, wuhy2025, xulan1, louxin\}@shanghaitech.edu.cn}}

\newcommand{\FFMinBytesRatio}{14.7}

\newcommand{\FFBestNewName}{WorldMirror~2.0}
\newcommand{\FFBestNewPSNR}{20.47}
\newcommand{\FFLiteOverBestNewDB}{6.37}
\newcommand{\FFBestNewBytesOverPro}{22.7}
\newcommand{\FFPointPSNR}{26.17}
\newcommand{\FFNewMinPSNR}{18.38}
\newcommand{\FFClipLabels}{PointSplat512 5{,}949, 28.18 / 4{,}462, 28.31; CodecSplat 46.5, 23.74 / 46.7, 24.68; DA3 32{,}908, 19.65 / 24{,}100, 21.03; WorldMirror~2.0 5{,}868, 19.82 / 4{,}591, 21.92; GPS-Gaussian 33{,}018, 16.79 / 23{,}592, 20.52}
\newcommand{\FFCodecSplatFourGPUFPS}{8.4}
\newcommand{\FFCodecSplatFPSBound}{9}
\newcommand{\RingProPSNR}{27.51}
\newcommand{\RingProOverPointDB}{0.93}
\newcommand{\RingProOverGPSDB}{10.99}
\newcommand{\RingPointBytesOverPro}{26.5}

\usepackage{placeins}
\begin{document}
\maketitle
{\renewcommand{\thefootnote}{}\footnotetext{$^{*}$Equal contribution. \quad $^{\dagger}$Corresponding authors.}}
\input{fig/teaser}
\begin{abstract}
Live free-viewpoint video streaming of humans enables immersive telepresence and interactive XR applications. However, this requires reconstructing each frame in real time, transmitting the data within strict bandwidth budgets, and decoding it on lightweight client GPUs. Existing approaches, whether relying on per-frame optimization or feed-forward inference, generate massive 3D Gaussian payloads that exceed practical streaming limits, while aggressive compression severely degrades rendering quality.
In this work, we present \sys, a real-time streaming framework for human-centric volumetric video. \sys splits a feed-forward reconstructor before its Gaussian head, introducing a multi-rate streaming codec driven by server-side token compression and client-side token restoration. On the server side, we apply opacity-guided token pruning, PCA-based dimensionality reduction, and lossless geometry coding for token anchors. On the client side, a feature reconstruction hypernetwork is fused into the Gaussian head to recover compression-induced quality loss. Furthermore, we introduce Neural Session Calibration (NSC), which rapidly fine-tunes the Gaussian head online using incoming camera views during a warm-up phase. This resolves quality degradation from novel camera parameters and varying capture conditions with zero per-frame overhead.
In our evaluations, \sys averages \ThirtyFPSMean\,FPS on four RTX~5090 GPUs at \BudgetProMbps\,Mbit/s. It decodes each frame into 3D Gaussians in \LDLaptopClientPro\,ms on an RTX~3050~Ti laptop GPU, transmitting \CompressionRatio$\times$ less data than uncompressed float32 Gaussians, while calibrated decoding improves PSNR by \ProDelta\,dB.
\end{abstract}

\section{Introduction}
\label{sec:intro}
Live free-viewpoint video streaming of humans enables immersive telepresence and interactive XR applications, allowing viewers to render arbitrary viewpoints in real time. At 30\,FPS, three stages in series must meet strict per-frame budgets (\cref{fig:positioning}). Specifically, the server must \emph{reconstruct} each frame in real time (every 33\,ms with bounded delay), the network must \emph{transmit} each frame within practical bandwidth budgets such as a 100\,Mbit/s 5G downlink~\cite{iturm2410}, and a lightweight \emph{client} must decode each frame into renderable 3D Gaussians~\cite{kerbl3dgs} in 33\,ms.

\input{fig/positioning}
However, whether reconstructed via per-frame optimisation or feed-forward inference, each frame generates heavy 3D Gaussian payloads that exceed practical streaming bandwidth, while aggressive compression degrades rendering quality (\cref{fig:positioning}). Per-frame optimisation methods~\cite{3dgstream,hicom,queen,instantgs,morgs} render in real time on the client~\cite{queen,streamstgs}, but fitting frame $t$ from frame $t{-}1$ turns server reconstruction into a sequential optimisation taking \CmpDynLaterS\,s per frame that cannot be parallelised across GPUs, at ${\ge}$\BudgetDynMinMbps\,Mbit/s~\cite{3dgstream,hicom,queen}. Feed-forward inference~\cite{gpsgaussian,pointsplat,tele360} reconstructs each frame in one pass, yet still emits dense 3D Gaussians (${\ge}$\BudgetFormatMinMbps\,Mbit/s with 3D Gaussian splatting (3DGS) compression methods~\cite{spz,fcgs}) or loses quality under aggressive video coding~\cite{tele360}. In alternative paradigms, the bottleneck merely shifts (\cref{sec:related}). Static-scene feature codecs~\cite{codecsplat} cause client overload (\CmpCodecSplatRecvMs\,ms), multi-view video streams duplicate reconstruction on every client (\CmpReconForwardMs\,ms), server-rendered video ties viewpoint changes to a network round trip, and person-specific avatars~\cite{deepappearance,animatablegaussians} need \HumanAvatarDays\ days of training.

We present \sys, a real-time streaming framework for human-centric volumetric video that splits a frame-parallel feed-forward reconstructor before its Gaussian head, streaming compressed tokens from the server while the client restores tokens and runs the head to generate 3D Gaussians. Existing 3DGS streaming methods compress explicit Gaussians after the head, where non-linear activations---particularly exponential scaling---map compact, correlated tokens into an expanded attribute space in which small quantisation errors amplify rendering distortion by four orders of magnitude (\cref{sec:exp_cut}). Compressing pre-head tokens instead of post-head attributes achieves high compression ratios with minimal quality loss; although each 1024-dimensional token in PointSplat~\cite{pointsplat} has more dimensions than the 224 attributes of its 16 Gaussians, the same codec outperforms post-head attribute compression by \CutTOverAttrMin--\CutTOverAttrMax\,dB.

Building on this split, we design a multi-rate streaming codec consisting of server-side token compression and client-side token restoration. On the server, we perform opacity-guided token pruning, PCA-based token dimensionality reduction, and lossless geometry coding of token anchor coordinates. On the client, we design a FiLM-conditioned \emph{feature reconstruction hypernetwork} to recover the quality lost to compression, and fuse both inverse PCA and the hypernetwork into the client Gaussian head. Leveraging kept PCA coefficients, local anchor geometry, and spatial neighbours, the hypernetwork and its one-time update recover \LDRecoveredPro\% of the compression loss at zero per-frame bitrate overhead, without explicit 1024-dimensional token expansion.

Furthermore, we introduce \emph{neural session calibration} (NSC), which rapidly fine-tunes the Gaussian head online during the warm-up stage using incoming camera views, resolving the quality loss of the Gaussian head caused by new camera parameters and varying capture conditions at zero per-frame overhead. On HiFi4G~\cite{hifi4g} and ActorsHQ~\cite{humanrf}, before calibration, removing token compression changes PSNR by at most \NSCHyperCodecCostMax\,dB (\cref{fig:nsc_gap,tab:nsc_gap}), confirming that this quality loss occurs even without compression rather than from codec distortion; NSC fits a compact \NSCKB\,kB low-rank update on the first second of the eight input views and fuses it into the client Gaussian head before streaming.

\sys meets all three budgets (\cref{fig:positioning}, bottom). It averages \ThirtyFPSMean\,FPS on \ThirtyFPSGPUsWord\ RTX~5090 GPUs at \BudgetProMbps\,Mbit/s for the Pro tier and decodes each frame into 3D Gaussians in \LDLaptopClientPro\,ms on an RTX~3050~Ti laptop GPU, with five GPUs streaming live at 30\,FPS over Wi-Fi at \GoLiveRenderMinMs--\GoLiveRenderMaxMs\,ms end-to-end latency. Pro transmits \CompressionRatio$\times$ less data than float32 Gaussians while calibrated decoding improves PSNR by \ProDelta\,dB, and saves \ReviewNSCSavingMin--\ReviewNSCSavingMax\% in bitrate over 3DGS compression methods at equal quality. From the same eight views, all three tiers outperform HiCoM, QUEEN, and 3DGStream in quality and bitrate on every sequence at \CmpDynSpeedupMean$\times$ the speed---with Pro outperforming offline 4C4D~\cite{fourcfourd} by \CmpDynFourCProGain\,dB at \CmpDynFourCOverPro$\times$ lower payload size---while NSC adds \NSCDNADeltaLearnedPro\,dB on DNA-Rendering~\cite{dnarendering} and \NSCHiFiDeployedIncrementPro\ and \NSCAHQDeployedIncrementPro\,dB on unseen rigs.
\paragraph{Contributions.}
\begin{enumerate}
\item \textbf{Multi-rate token streaming codec.} We design a multi-rate streaming codec for human-centric volumetric video consisting of server-side token compression and client-side token restoration. On the server, we perform opacity-guided token pruning, PCA-based token dimensionality reduction, and lossless geometry coding of token anchor coordinates, transmitting \CompressionRatio$\times$ less data than float32 Gaussians and, with our client decoder, saving \ReviewNSCSavingMin--\ReviewNSCSavingMax\% in bitrate over 3DGS compression methods at equal quality.
\item \textbf{Feature reconstruction hypernetwork and fused decoding.} On the client, we design a feature reconstruction hypernetwork that leverages kept PCA coefficients, local anchor geometry, and spatial neighbours---recovering \LDRecoveredPro\% of the quality lost to compression with its one-time update---and fuse inverse PCA and the hypernetwork into the Gaussian head to decode in \LDLaptopClientPro\,ms on a laptop GPU, enabling the full system to beat per-frame optimisation methods in quality, bitrate, and speed.
\item \textbf{Neural session calibration.} During the warm-up stage, we rapidly fine-tune the Gaussian head online using incoming camera views (\NSCKB\,kB low-rank update fused into the client head at zero per-frame overhead), resolving the quality loss of the Gaussian head caused by new camera parameters and varying capture conditions.
\end{enumerate}

\section{Related Work}
\label{sec:related}

\paragraph{Per-frame optimisation.}
3DGStream~\cite{3dgstream} trains a per-frame transformation cache, while HiCoM~\cite{hicom} and QUEEN~\cite{queen} accelerate per-frame optimisation with hierarchical motion and quantised residuals; ReCon-GS~\cite{recongs} adapts a motion hierarchy, ComGS~\cite{comgs} updates only motion-sensitive keypoints, and iFVC~\cite{ifvc} entropy-codes a binary transformation cache.
Instant Gaussian Stream~\cite{instantgs} predicts motion with a generalisable network but still refines key frames, MoRGS~\cite{morgs} regularises per-Gaussian motion with optical flow on key views, 4DGC~\cite{fourdgc} optimises motion grids for rate and distortion, and StreamSTGS~\cite{streamstgs} encodes canonical attributes as images and temporal features as video.
V$^3$~\cite{vcubed}, VideoRF~\cite{videorf}, and TeTriRF~\cite{tetrirf} pack attributes or features into 2D maps for video codecs, and ReRF~\cite{rerf} codes residual radiance fields.
Human-centric systems such as HiFi4G~\cite{hifi4g} and DualGS~\cite{dualgs} track a performer and compress the resulting 4D Gaussians, SharpTimeGS~\cite{sharptimegs} modulates Gaussian lifespans, and NeuS2~\cite{neus2} fine-tunes a hash-encoded surface from frame to frame.
Because each frame is optimised sequentially from the previous one for seconds to minutes (\cref{tab:cmp_lit,tab:cmp_human_speed}), per-frame optimisation cannot reach real-time live streaming.
Sparse-camera 4C4D~\cite{fourcfourd} and SparseCam4D~\cite{sparsecam4d} instead optimise a whole sequence offline using future frames.

\paragraph{Feed-forward reconstruction and 3DGS compression.}
PointSplat~\cite{pointsplat} regresses 3D Gaussians around a coarse point proxy with a Point-Image Transformer in a single pass, which we retain frozen as our reconstructor.
GPS-Gaussian~\cite{gpsgaussian,gpsgaussianplus,fastvolcap} regresses pixel-wise Gaussian maps from adjacent view pairs in real time, and HiReFF~\cite{hireff} regresses Gaussians from four uncalibrated views.
Concurrent Tele360~\cite{tele360} reconstructs Gaussian maps from four to six unposed cameras at over 25\,FPS on one GPU and streams the output Gaussians as 2D maps in an H.265 video~\cite{vcubed,lgscv}, losing roughly 2\,dB to video compression and flickering under rapid motion. LiveGS~\cite{livegs} brings live free-viewpoint video to mobile devices.
General 3DGS compression methods such as SPZ~\cite{spz}, pruning and entropy-coding pipelines~\cite{lightgaussian,compgs,hac,mesongs}, 2D video-codec grids~\cite{sog,lgscv}, and training-free or feed-forward compressors~\cite{tinysplat,fcgs,dfcgs} all operate after the Gaussian head has expanded tokens into non-linear attributes.

\paragraph{Latent feature coding and split computing.}
Split computing executes early network layers on one device and transmits intermediate activations to another~\cite{neurosurgeon,choibaf,cohenfeature,matsubara}.
GIFStream~\cite{gifstream} compresses temporal feature streams of a 4D representation with learned networks.
CodecSplat~\cite{codecsplat} and FeCoSplat~\cite{fecosplat} train entropy codecs for intermediate features of a static-scene feed-forward model, and concurrent MFGS~\cite{mfgs} codes multi-view static-scene features before Gaussian prediction.
GS-SCNet~\cite{gsscnet} jointly trains a stereo image codec with a two-view Gaussian predictor at 19.2\,FPS on an A100, and Feat2GS~\cite{feat2gs} projects foundation-model features with PCA before readout.
Rather than training a heavy entropy codec for static scenes, \sys splits a frozen multi-view reconstructor before its Gaussian head, pairing server PCA token compression with a fused client feature reconstruction hypernetwork for frame-parallel multi-GPU streaming.
Where FeCoSplat codes sender-side reconstruction feedback as a second bitstream for every scene, \sys amortises session adaptation into one-time client updates at zero per-frame overhead.

\paragraph{Live 3D telepresence and human avatars.}
Holoportation~\cite{holoportation}, Fusion4D~\cite{fusion4d}, DoubleFusion~\cite{doublefusion}, and Function4D~\cite{function4d} capture performers live by fusing depth sensors.
From RGB cameras, Live4D~\cite{live4d} streams server-fused meshes, Tele-Aloha~\cite{telealoha} renders telepresence views from four cameras, LiveStre4m~\cite{livestre4m} renders novel-view video from two unposed streams with frame interpolation, and StreamSplat~\cite{streamsplat} updates a static feed-forward reconstruction causally.
Person-specific avatars transmit a compact driving code per frame but require dedicated per-subject capture and days of training~\cite{deepappearance,animatablegaussians,holochar}.
\sys instead streams pre-head tokens of arbitrary subjects, enabling clients to decode 3D Gaussians and render free-viewpoint video locally.

\paragraph{Test-time and session adaptation.}
Instance-adaptive neural compression fine-tunes a codec per sequence or static scene and transmits weight updates~\cite{vanrozendaal,ohpeft,codecnerf,lora}, while test-time training adapts reconstruction backbones to distribution shift~\cite{tttsun,videottt,lora3d,test3r,tttlrm,tco}.
Neural session calibration instead keeps the reconstructor and token stream fixed, rapidly fine-tuning the Gaussian head online on warm-up input views and fusing the update at zero per-frame overhead.

\section{Method}
\label{sec:method}
\input{fig/method}
\Cref{fig:method} shows the system. The server receives eight synchronised, calibrated $1024\times1024$ RGB views per frame, area-averages them to $512\times512$, predicts soft alpha with BiRefNet Lite~\cite{birefnet} fine-tuned on mask-annotated DNA-Rendering data, and passes RGB and alpha to frozen PointSplat~\cite{pointsplat}, which fuses all eight calibrated views jointly rather than a stereo pair per target view~\cite{gpsgaussian}.

We split the network immediately before its Gaussian head. Each token $f_i\in\mathbb{R}^{1024}$ generates a group of 16 Gaussian slots around anchors $X_i\in\mathbb{R}^{16\times3}$ from which the head predicts bounded offsets. A frame thus consists of tokens $F=(f_1,\ldots,f_N)$ and anchors $X=(X_1,\ldots,X_N)$. The server (\cref{sec:encoder}) performs opacity-guided token pruning, PCA-based token dimensionality reduction, and lossless anchor coding; the client (\cref{sec:decoder}) fuses inverse PCA and a feature reconstruction hypernetwork into the client head $h_\psi$ to decode 3D Gaussians from $(\hat F,X)$, while neural session calibration (\cref{sec:nsc}) rapidly fine-tunes $h_\psi$ online during warm-up. A session sends its PCA basis, hypernetwork update, and NSC update once at warm-up; each packet then carries only coefficients, ranges, and anchors for stateless frame-parallel streaming (\cref{sec:production}) and independent decoding.

\subsection{Server-Side Token Compression}
\label{sec:encoder}
\paragraph{PCA-based token dimensionality reduction.}
Let $\mu\in\mathbb{R}^{1024}$ and $U_r\in\mathbb{R}^{1024\times r}$ be the mean and first $r$ principal components of the session's first-frame tokens. Each token is coded and decoded as
\begin{equation}
 c_i=U_r^\top(f_i-\mu),\qquad \hat f_i=\mu+U_r\hat c_i,
 \label{eq:pca}
\end{equation}
where $\hat c_i$ are the quantised coefficients. Keeping $(\mu,U_r)$ fixed for the whole session makes encoding strictly causal. Every frame sends one range per component, quantises coefficients uniformly to $b$ bits within it, and compresses them in component-major order with Zstd. The tiers Lite $(r,b)=(32,4)$, default Pro $(48,6)$, and Ultra $(80,6)$ share this representation, where a quantised Pro token occupies $48\times6=288$ bits instead of the $1024\times16$ bits of a BF16 token. As the Karhunen--Lo\`eve transform~\cite{goyal}, PCA minimises squared linear projection error at rank $r$ on the fitting frame without training an encoder against the frozen reconstructor, and the Pro basis takes only \StateProKB\,kB, sent once.

\paragraph{Lossless anchor coding.}
Because an anchor error shifts a whole group, anchor coordinates are coded losslessly. On the voxel grid, each packet stores coordinate tables, each group's occupied $3\times3\times3$ neighbourhood as a 32-bit mask, and one index per slot, Zstd-coded into \GeometryKB\,kB/frame.

\paragraph{Opacity-guided token pruning.}
Many token groups generate only near-transparent Gaussians, and the server identifies them without running the head. A lightweight predictor $g_\phi$ standardises the raw token per dimension and regresses the 16 slot-opacity logits of the uncompressed head via a bias-free $1024\to16$ projection followed by $16\to32\to16$ SiLU layers. Group $i$ is kept if and only if
\begin{equation}
 q_i=\max_{k}\,\sigma\bigl(g_\phi(f_i)_k\bigr)\ \ge\ \delta,\qquad \delta=0.075.
 \label{eq:prune}
\end{equation}
The maximum retains every group with at least one visible slot. Trained against the uncompressed head's logits, one predictor and threshold serve all tiers, sequences, and rigs (\cref{sec:supp_controls}).

\subsection{Client-Side Token Restoration}
\label{sec:decoder}
\paragraph{Fusing inverse PCA into the Gaussian head.}
The head begins with a linear map $y=Wf+b_0$ before its first activation. Substituting \cref{eq:pca} gives
\begin{equation}
 y=(WU_r)\,\hat c+(W\mu+b_0).
 \label{eq:fused}
\end{equation}
Precomputing both terms once fuses inverse PCA into the Gaussian head, avoiding explicit 1024-dimensional token expansion; fused and unfused decoding differ by only \FuseDelta\,dB.

\paragraph{Feature reconstruction hypernetwork.}
Although PCA projects each token linearly and independently, the kept coefficients, local anchor geometry, and neighbouring groups retain non-linear and spatial cues about the discarded components. The client restores them via a residual hypernetwork $H_\theta$,
\begin{equation}
 \tilde f_i=\hat f_i+H_\theta\Bigl(\hat f_i,\;X_i-\bar X_i,\;\tfrac18\textstyle\sum_{j\in\mathcal N_i}\hat f_j;\;r,b\Bigr),
 \label{eq:hypernetwork}
\end{equation}
where $X_i-\bar X_i$ is the anchor layout about centroid $\bar X_i$ and $\mathcal N_i$ holds the eight nearest groups in the same frame. An embedding of the tier $(r,b)$ generates FiLM~\cite{film} parameters for a shared four-layer width-1024 MLP, so one network serves every tier. Because the head's first layer projects tokens via $W$, $H_\theta$ is trained by minimising $\|W(f_i-\tilde f_i)\|_2^2$ directly against the uncompressed token without rendering. Since $\hat f_i$ and the neighbour mean are linear in $\hat c$, the first layer of $H_\theta$ fuses with $U_r$ (averaging neighbours in $r$ dimensions) and its output layer fuses into $W$, so restored tokens are never explicitly expanded either. $H_\theta$ is trained once on \HyperTrainScenes\ DNA-Rendering training scenes, selected on \HyperValActors\ held-out actors, and shipped with the client.

\paragraph{Hypernetwork update and client pruning.}
During warm-up, the server fits, for each streamed tier, a rank-32 subspace update~\cite{lora} of the four layers of $H_\theta$ on frames 0--29 against the uncompressed tokens (\HyperUpdateKB\,kB in FP16), which the client fuses into $H_\theta$ at zero per-frame cost; with the PCA basis and NSC update, the Pro session state totals \SessionStateProKB\,kB. The client then prunes the lowest-opacity 20\% of Gaussians.

\subsection{Neural Session Calibration}
\label{sec:nsc}
\begin{figure}[t]
\centering
\begin{tikzpicture}
\begin{axis}[hcplot, width=0.66\columnwidth, height=4.2cm, scale only axis=false, y dir=reverse, xmin=-0.90, xmax=2.07, ymin=-0.45, ymax=3.55, ytick={0,...,3}, yticklabels={{Pro, plain decoding},{Pro + hypernetwork},{Pro + hypernetwork + NSC},{Uncompressed + NSC}}, ytick style={draw=none}, y axis line style={draw=none}, ymajorgrids=false, xlabel={PSNR gain over the uncompressed reconstruction (dB)}, legend style={font=\sffamily\hcsmall, draw=none, fill=none, at={(0.5,1.02)}, anchor=south, legend columns=3, /tikz/every even column/.append style={column sep=6pt}}, legend cell align=left, legend image code/.code={\fill[#1] (0cm,-0.06cm) rectangle (0.22cm,0.08cm);}]
\addplot[xbar, bar width=3.6pt, fill=cA, draw=none] coordinates {(-0.601,-0.30) (-0.118,0.70) (1.036,1.70) (1.240,2.70)};
\addlegendentry{DNA-Rendering}
\node[anchor=west, font=\sffamily\fontsize{5}{5.5}\selectfont, text=ink!80, inner sep=0.6pt] at (axis cs:0.000,-0.30) {-0.60};
\node[anchor=west, font=\sffamily\fontsize{5}{5.5}\selectfont, text=ink!80, inner sep=0.6pt] at (axis cs:0.000,0.70) {-0.12};
\node[anchor=west, font=\sffamily\fontsize{5}{5.5}\selectfont, text=ink!80, inner sep=0.6pt] at (axis cs:1.036,1.70) {+1.04};
\node[anchor=west, font=\sffamily\fontsize{5}{5.5}\selectfont, text=ink!80, inner sep=0.6pt] at (axis cs:1.240,2.70) {+1.24};
\addplot[xbar, bar width=3.6pt, fill=cB, draw=none] coordinates {(-0.284,0.00) (0.042,1.00) (1.462,2.00) (1.462,3.00)};
\addlegendentry{HiFi4G}
\node[anchor=west, font=\sffamily\fontsize{5}{5.5}\selectfont, text=ink!80, inner sep=0.6pt] at (axis cs:0.000,0.00) {-0.28};
\node[anchor=west, font=\sffamily\fontsize{5}{5.5}\selectfont, text=ink!80, inner sep=0.6pt] at (axis cs:0.042,1.00) {+0.04};
\node[anchor=west, font=\sffamily\fontsize{5}{5.5}\selectfont, text=ink!80, inner sep=0.6pt] at (axis cs:1.462,2.00) {+1.46};
\node[anchor=west, font=\sffamily\fontsize{5}{5.5}\selectfont, text=ink!80, inner sep=0.6pt] at (axis cs:1.462,3.00) {+1.46};
\addplot[xbar, bar width=3.6pt, fill=cC, draw=none] coordinates {(-0.541,0.30) (-0.069,1.30) (1.367,2.30) (1.465,3.30)};
\addlegendentry{ActorsHQ}
\node[anchor=west, font=\sffamily\fontsize{5}{5.5}\selectfont, text=ink!80, inner sep=0.6pt] at (axis cs:0.000,0.30) {-0.54};
\node[anchor=west, font=\sffamily\fontsize{5}{5.5}\selectfont, text=ink!80, inner sep=0.6pt] at (axis cs:0.000,1.30) {-0.07};
\node[anchor=west, font=\sffamily\fontsize{5}{5.5}\selectfont, text=ink!80, inner sep=0.6pt] at (axis cs:1.367,2.30) {+1.37};
\node[anchor=west, font=\sffamily\fontsize{5}{5.5}\selectfont, text=ink!80, inner sep=0.6pt] at (axis cs:1.465,3.30) {+1.47};
\draw[ink!55, line width=0.5pt] (axis cs:0,-0.45) -- (axis cs:0,3.55);
\end{axis}
\end{tikzpicture}
\caption{\textbf{Neural session calibration improves quality across datasets while token coding costs little.} Decoded by inverse PCA alone, the Pro stream loses up to \NSCGapProPlainMax\,dB to our uncompressed reconstruction with the same pruning; the hypernetwork keeps it within \NSCGapProStreamMax\,dB, and NSC adds \NSCDNADeltaLearnedPro, \NSCHiFiDeployedIncrementPro, and \NSCAHQDeployedIncrementPro\,dB on top. Evaluated on the 16 DNA-Rendering test scenes and on held-out cameras over all frames after the one-second warm-up on HiFi4G and ActorsHQ. NSC is fitted on the input views, without the codec for Uncompressed + NSC.}
\label{fig:nsc_gap}
\end{figure}
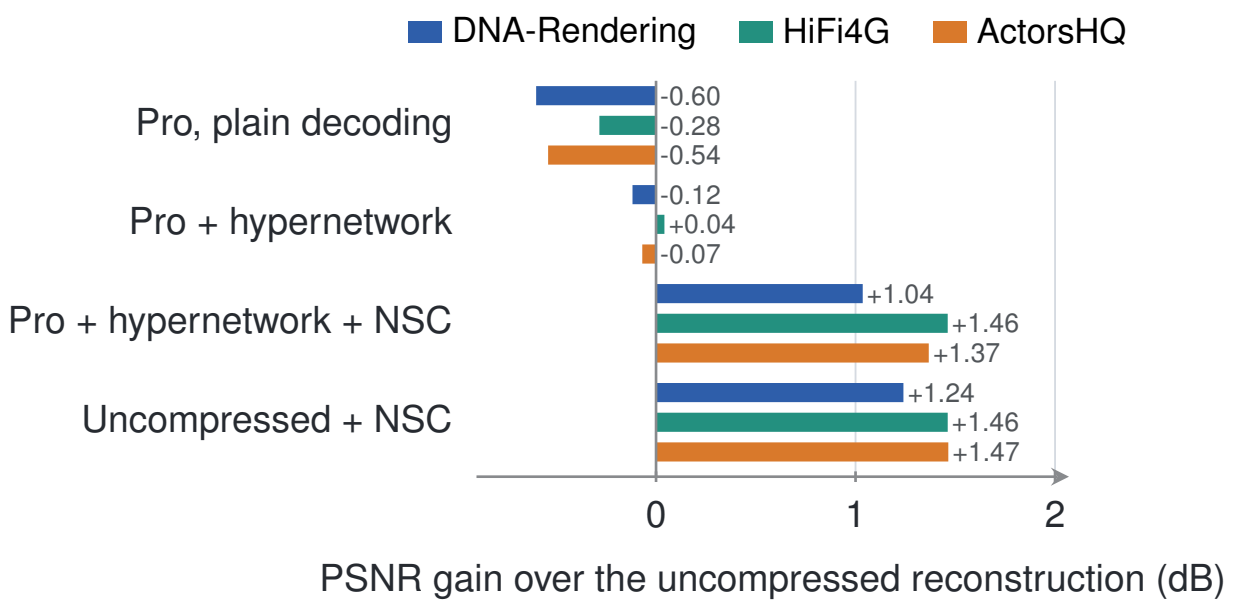
A pre-trained Gaussian head suffers quality loss under new camera parameters and varying capture conditions, even without compression (\cref{fig:nsc_gap}). During warm-up, we rapidly fine-tune the Gaussian head online using incoming camera views under four streaming constraints: (i)~\emph{the token stream remains unchanged} for every viewer; (ii)~\emph{self-supervision is causal}, using the session's input views prior to evaluated frames; (iii)~\emph{zero per-frame overhead} is added; and (iv)~\emph{the one-time update stays compact}.

\paragraph{Self-supervised objective.}
Let $\mathcal{W}$ be the first 30 frames of the session (one second of capture), $\tilde F_t$ the restored tokens of frame $t$ at the streamed tier, and $\mathcal{R}_v$ rendering after pruning into input view $v$ with image $I_{v,t}$. The server solves
\begin{equation}
 \min_{\Delta}\;\sum_{t\in\mathcal{W}}\sum_{v=1}^{8}\mathcal{L}\Bigl(\mathcal{R}_v\bigl(h_{\psi+\Delta}(\tilde F_t,X_t)\bigr),\,I_{v,t}\Bigr)
 \label{eq:nsc}
\end{equation}
for each streamed tier with $\mathcal{L}=0.8\,\ell_1+0.2\,(1-\mathrm{SSIM})+0.1\,\mathrm{LPIPS}$, keeping the encoder, pruning predictor, codec, and hypernetwork frozen to satisfy (i) and (ii).

\paragraph{Low-rank Gaussian head update.}
To adjust Gaussian positions, scales, opacities, and colours jointly, we parameterise $\Delta$ as a rank-5 update $\Delta W_\ell = A_\ell B_\ell$ across the three per-slot MLP layers $\ell$ of the head~\cite{lora}, requiring only \NSCParams\ FP16 parameters (\NSCPayloadKB\,kB; \NSCKB\,kB with headers and checksum, \NSCShareOfPro\% of an average Pro packet), satisfying (iv); a full head update would be \NSCFullHeadRatio$\times$ larger. Fusing each $A_\ell B_\ell$ into its layer once before inverse PCA fusion (\cref{eq:fused}) preserves the head shape and per-frame cost to satisfy (iii).

\subsection{Real-Time Frame-Parallel Streaming}
\label{sec:production}
Because frames are coded independently, the server parallelises across frames. Each GPU runs three persistent worker processes under CUDA MPS~\cite{cudamps}; incoming frames are dispatched to the GPU with the fewest outstanding frames, overlapping matting, inference, token pruning, and PCA projection without a global barrier, while one central encoder quantises and Zstd-codes packets. At most 30 in-flight frames bound memory, and a reorder buffer releases packets strictly in capture order.

\section{Experiments}
\label{sec:experiments}

\subsection{Experimental Setup}
\label{sec:setup}
\paragraph{Datasets and metrics.}
We follow the official PointSplat evaluation protocol on DNA-Rendering~\cite{dnarendering} (16 test scenes, eight input and 48 evaluated cameras at $512\times512$, frame 30, after the warm-up), plus three unseen rigs scored on held-out views over all frames after a one-second warm-up: HiFi4G~\cite{hifi4g}, ActorsHQ~\cite{humanrf}, and synthetic X-Humans~\cite{xavatar} (\cref{sec:supp_xhumans}). On DNA-Rendering, soft masks from BiRefNet Lite~\cite{birefnet} fine-tuned on DNA-Rendering masks composite captured inputs and ground truth. We report PSNR, SSIM, and LPIPS (VGG) on a black background within the ground-truth mask bounding box, weighting scenes equally.

\paragraph{Baselines and accounting.}
Standard PointSplat512 is the released model with native opacity pruning and 14 float32 attributes per Gaussian. All tiers use opacity-guided token pruning, the fused feature reconstruction hypernetwork with its one-time update, NSC, and 20\% client pruning, and packet sizes include all side information of each coded frame.
Servers run the full streaming pipeline on one to six RTX~5090 GPUs; we report steady-state throughput in frames/s (\cref{sec:supp_handoff}).

\subsection{Token Streaming Quality and Rate}
\label{sec:exp_streaming}
\begin{table*}[t]
\centering
\small
\setlength{\tabcolsep}{3.0pt}
\resizebox{\textwidth}{!}{%
\begin{tabular}{lrrrrrrrrrrrrrrrr}
\toprule
Method & \multicolumn{4}{c}{DNA-Rendering (16 scenes)} & \multicolumn{4}{c}{HiFi4G} & \multicolumn{4}{c}{ActorsHQ} & \multicolumn{4}{c}{X-Humans (synthetic)} \\ 
 & PSNR\,$\uparrow$ & SSIM\,$\uparrow$ & LPIPS\,$\downarrow$ & kB/frame\,$\downarrow$ & PSNR\,$\uparrow$ & SSIM\,$\uparrow$ & LPIPS\,$\downarrow$ & kB/frame\,$\downarrow$ & PSNR\,$\uparrow$ & SSIM\,$\uparrow$ & LPIPS\,$\downarrow$ & kB/frame\,$\downarrow$ & PSNR\,$\uparrow$ & SSIM\,$\uparrow$ & LPIPS\,$\downarrow$ & kB/frame\,$\downarrow$ \\
\midrule
PointSplat512 & 26.169 & 0.8650 & 0.1034 & 4753.1 & 22.810 & 0.7017 & 0.2137 & 2564.6 & 24.438 & 0.8131 & 0.1749 & 10631.7 & 26.189 & 0.8692 & 0.1057 & 2552.0 \\
Uncompressed + NSC & 27.423 & 0.8857 & 0.0884 & -- & 24.271 & 0.7553 & 0.1895 & -- & 25.904 & 0.8355 & 0.1642 & -- & 27.445 & 0.9014 & 0.0878 & -- \\
Ours Lite (32/4) & 26.840 & 0.8700 & 0.1043 & \textbf{94.8} & 24.148 & 0.7414 & 0.2069 & \textbf{65.0} & 25.721 & 0.8274 & 0.1809 & \textbf{177.9} & 26.733 & 0.8877 & 0.1073 & \textbf{65.0} \\
Ours Pro (48/6) & 27.219 & 0.8802 & 0.0941 & 208.2 & 24.270 & 0.7511 & 0.1953 & 141.6 & 25.806 & 0.8320 & 0.1703 & 404.4 & 27.188 & 0.8965 & 0.0947 & 141.1 \\
Ours Ultra (80/6) & \textbf{27.273} & \textbf{0.8818} & \textbf{0.0923} & 321.9 & \textbf{24.306} & \textbf{0.7519} & \textbf{0.1931} & 219.2 & \textbf{25.854} & \textbf{0.8333} & \textbf{0.1680} & 628.3 & \textbf{27.267} & \textbf{0.8982} & \textbf{0.0922} & 217.5 \\
\bottomrule
\end{tabular}}
\caption{\textbf{\sys sends a fraction of PointSplat's Gaussian bytes and exceeds its PSNR and SSIM at every tier across four camera rigs.} Quality on a black background within the ground-truth mask bounding box and mean per-frame kB; held-out cameras on HiFi4G, ActorsHQ, and synthetic X-Humans. All tiers use the fused hypernetwork decoder and NSC; the second row (uncompressed tokens with NSC) is a reference, and bold marks the best non-reference result.}
\label{tab:official_quality}
\end{table*}

On DNA-Rendering (\cref{tab:official_quality}), our uncompressed path matches PointSplat512 (\UncDelta\,dB; \cref{tab:official_quality_full}). Pre-head token compression shrinks float32 Gaussians \CompressionRatio$\times$ to \ProKB\,kB/frame at Pro, while the feature reconstruction hypernetwork and NSC improve Pro by \ProDelta\,dB over it; over frames 30--89 of all 16 sequences, Lite/Pro/Ultra sustain \CmpDynOursLitePSNR/\CmpDynOursProPSNR/\CmpDynOursUltraPSNR\,dB (\cref{tab:cmp_dyn}). Across unseen HiFi4G, ActorsHQ, and X-Humans, every tier outperforms PointSplat512 in PSNR and SSIM—with Pro and Ultra also improving LPIPS on all three—gaining \HiFiGainMin--\HiFiGainMax, \AHQGainMin--\AHQGainMax, and \XHGainMin--\XHGainMax\,dB at \HiFiRatioMin--\HiFiRatioMax$\times$, \AHQRatioMin--\AHQRatioMax$\times$, and \XHRatioMin--\XHRatioMax$\times$ smaller payloads.

\paragraph{Opacity-guided token pruning.}
Token pruning removes \ServerDrop\% of token groups without costing PSNR and reduces an Ultra packet by \ByteDrop\%. Over six codec settings (\cref{tab:two_layer_full}), the pruned stream gains \PruneBDPSNR\,dB BD-PSNR and saves \PruneBDRate\% BD-rate~\cite{bjontegaard}.

\paragraph{Feature reconstruction hypernetwork and fused decoding.}
From identical packets, the hypernetwork (with its one-time update) improves Lite, Pro, and Ultra by \LDDeltaLite, \LDDeltaPro, and \LDDeltaUltra\,dB over plain PCA decoding, recovering \LDRecoveredLite, \LDRecoveredPro, and \LDRecoveredUltra\% of the compression loss (\cref{tab:learned_decoding}); with NSC, gains reach \NSCDNADeltaPlainLite, \NSCDNADeltaPlainPro, and \NSCDNADeltaPlainUltra\,dB (positive on all 16 scenes), outperforming our uncompressed tokens without NSC at every tier (\cref{fig:nsc_gap,tab:nsc_dna}).

\subsection{Pre-Head Token vs.\ Post-Head Attribute Compression}
\label{sec:exp_cut}
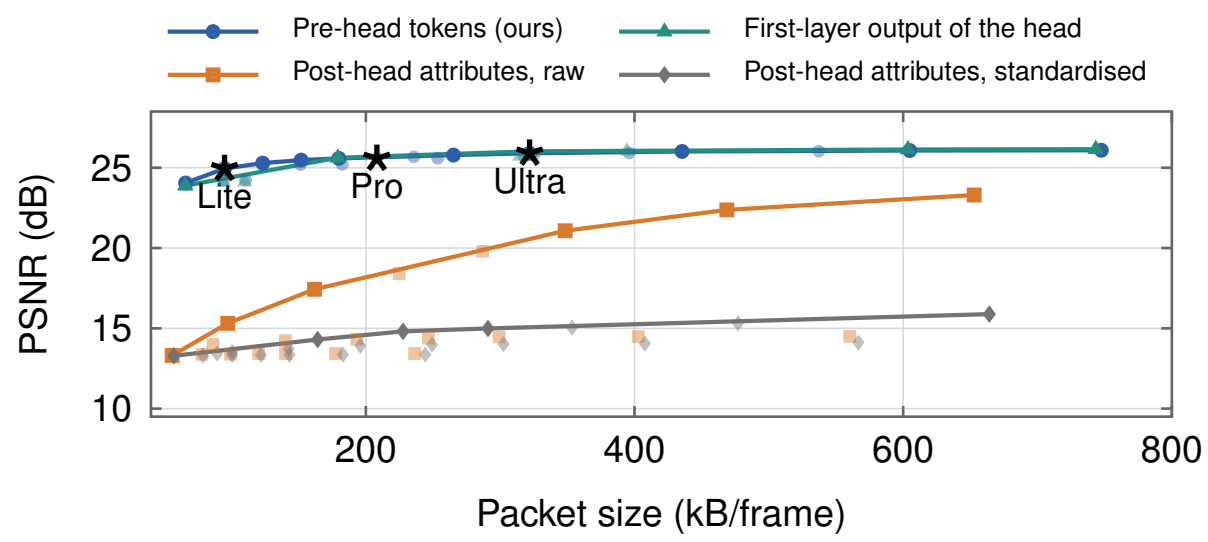
\begin{figure}[t]
\centering
\begin{tikzpicture}
\begin{axis}[paper, width=\columnwidth, height=3.6cm, xmin=40, xmax=800, ymin=9.5, ymax=28.5, xlabel={Packet size (kB/frame)}, ylabel={PSNR (dB)}, legend columns=2, legend style={at={(0.5,1.02)}, anchor=south, column sep=4pt}]
\addplot[cA, mark=*, mark size=1pt, line width=0.8pt] coordinates {(65.8,24.052) (94.8,24.948) (123.2,25.289) (151.8,25.475) (179.9,25.580) (265.2,25.789) (321.9,25.902) (435.4,26.013) (604.9,26.072) (747.5,26.088)};\addlegendentry{Pre-head tokens (ours)}
\addplot[only marks, cA, mark=*, mark size=0.7pt, opacity=0.45, forget plot] coordinates {(65.8,24.052) (94.0,24.227) (94.8,24.948) (110.4,24.235) (123.2,25.289) (151.3,25.204) (151.8,25.475) (179.9,25.580) (182.4,25.215) (208.2,25.580) (235.7,25.678) (253.5,25.592) (265.2,25.789) (317.8,25.729) (321.9,25.902) (325.0,25.802) (395.9,25.918) (435.4,26.013) (537.2,26.029) (604.9,26.072) (747.5,26.088)};
\addplot[cB, mark=triangle*, mark size=1pt, line width=0.8pt] coordinates {(65.6,23.880) (179.1,25.621) (321.6,26.021) (603.6,26.135) (743.3,26.154)};\addlegendentry{First-layer output of the head}
\addplot[only marks, cB, mark=triangle*, mark size=0.7pt, opacity=0.45, forget plot] coordinates {(65.6,23.880) (93.9,24.109) (109.8,24.121) (179.1,25.621) (314.8,25.723) (321.6,26.021) (394.3,26.041) (603.6,26.135) (743.3,26.154)};
\addplot[cC, mark=square*, mark size=1pt, line width=0.8pt] coordinates {(55.8,13.313) (97.2,15.312) (161.8,17.432) (348.4,21.076) (468.7,22.375) (652.9,23.301)};\addlegendentry{Post-head attributes, raw}
\addplot[only marks, cC, mark=square*, mark size=0.7pt, opacity=0.45, forget plot] coordinates {(55.8,13.313) (78.0,13.381) (86.2,13.989) (97.2,15.312) (99.3,13.391) (120.1,13.423) (140.1,14.232) (140.2,13.417) (161.8,17.432) (177.4,13.433) (193.3,14.323) (224.8,18.406) (236.3,13.430) (246.5,14.394) (287.0,19.793) (299.2,14.462) (348.4,21.076) (402.8,14.483) (468.7,22.375) (560.3,14.502) (652.9,23.301)};
\addplot[black!55, mark=diamond*, mark size=1pt, line width=0.8pt] coordinates {(57.1,13.292) (164.1,14.304) (227.7,14.812) (290.9,14.993) (664.1,15.885)};\addlegendentry{Post-head attributes, standardised}
\addplot[only marks, black!55, mark=diamond*, mark size=0.7pt, opacity=0.45, forget plot] coordinates {(57.1,13.292) (78.7,13.315) (89.2,13.424) (100.4,13.525) (100.4,13.336) (121.9,13.345) (142.5,13.732) (143.1,13.351) (164.1,14.304) (183.0,13.364) (195.9,13.913) (227.7,14.812) (244.2,13.366) (249.3,13.986) (290.9,14.993) (302.5,14.014) (353.5,15.054) (407.7,14.049) (477.1,15.313) (566.7,14.122) (664.1,15.885)};
\addplot[only marks, mark=star, mark size=2.6pt, black, forget plot] coordinates {(94.8,24.948) (208.2,25.580) (321.9,25.902)};
\node[font=\sffamily\hcsmall, anchor=north, inner sep=2.5pt] at (axis cs:94.8,24.948) {Lite};
\node[font=\sffamily\hcsmall, anchor=north, inner sep=2.5pt] at (axis cs:208.2,25.580) {Pro};
\node[font=\sffamily\hcsmall, anchor=north, inner sep=2.5pt] at (axis cs:321.9,25.902) {Ultra};
\end{axis}
\end{tikzpicture}
\caption{\textbf{Coding tokens before the Gaussian head beats post-head attribute coding by \CutTOverAttrMin--\CutTOverAttrMax\,dB across rates.} The same session PCA, quantisation, and Zstd encoder at three split points of the frozen reconstructor, with opacity-guided token pruning and no hypernetwork, on the 16 DNA-Rendering test scenes. Curves join each split point's best settings; stars mark the Lite, Pro, and Ultra settings.}
\label{fig:cut}
\end{figure}
We apply the encoder of \cref{sec:encoder} without the hypernetwork at three split points of the reconstructor (\cref{fig:cut}). Raw or standardised, a group's 224 pre-activation attributes keep \CutVarAttrMin--\CutVarAttrMax\% of their variance at rank 80 vs.\ \CutVarTokens\% for tokens, yet coding them fails because a 1\% relative attribute perturbation costs $14{,}446\times$ the rendered MSE of an equal-energy token perturbation, driven by exponential scales (\cref{sec:supp_tiers}). At \CutAttrBestKB\,kB they reach \CutAttrBestPSNR\,dB, \CutLiteOverAttrBest\,dB below Lite at \CutAttrBestOverLiteBytes$\times$ its payload size. Coding the head's 16{,}384-dimensional first-layer output gains only \CutHGainMin--\CutHGainMax\,dB at ranks 80 and 160; 3DGS codecs quantising log-scales require \CmpMinCodecOverPro$\times$ Pro's payload size (\cref{sec:exp_compare}), and pre-head token coding likewise outperforms post-head attribute coding on DepthSplat and GPS-Gaussian (\cref{sec:supp_dyn}).

\subsection{Comparison with State-of-the-Art Methods}
\label{sec:exp_compare}
\begin{table}[t]
\centering
\small
\setlength{\tabcolsep}{4pt}
\resizebox{\columnwidth}{!}{%
\begin{tabular}{lrrrr}
\toprule
Format & kB/frame\,$\downarrow$ & PSNR\,$\uparrow$ & SSIM\,$\uparrow$ & Decode (ms)\,$\downarrow$ \\
\midrule
PointSplat512, float32 & 4753.1 & 26.169 & 0.8650 & -- \\
Uncompressed + NSC, float32 & 3630.7 & 27.423 & 0.8857 & -- \\
\midrule
SPZ v4 & 852.7 & 27.357 & 0.8844 & 27.9 \\
PlayCanvas compressed PLY & 1056.3 & 27.364 & 0.8846 & 209.4 \\
FCGS~\cite{fcgs} & 866.5 & 27.207 & 0.8777 & 4562.7 \\
ksplat & 1564.2 & 27.197 & 0.8824 & 154.6 \\
Draco & 723.0 & 27.415 & 0.8854 & 15.3 \\
PlayCanvas SOG & 841.2 & 27.226 & 0.8827 & 221.2 \\
\midrule
CodecSplat~\cite{codecsplat} & 45.4 & 23.143 & 0.7434 & 199.1 \\
\midrule
Ours Lite & 94.8 & 26.840 & 0.8700 & 21.7 \\
Ours Pro & 208.2 & 27.219 & 0.8802 & 22.1 \\
Ours Ultra & 321.9 & 27.273 & 0.8818 & 22.8 \\
\bottomrule
\end{tabular}}
\caption{\textbf{With the same calibration, the smallest Gaussian format needs at least \CmpSameCalMinBytesOverUltra$\times$ the bytes of Ultra and \CmpSameCalMinBytesOverPro$\times$ those of Pro.} Formats encode our uncompressed pruned Gaussians with NSC at their default or recommended setting; CodecSplat codes its own features from our eight views at its top rate, excluding input masks. Decode: per-frame decoding to Gaussians on an RTX~3050~Ti laptop (ours), an RTX~5090 (CodecSplat), or CPU reference decoders (formats; FCGS on GPU); \cref{tab:cmp_formats_nonsc} drops NSC.}
\label{tab:cmp_formats}
\end{table}

\begin{table}[t]
\centering
\small
\setlength{\tabcolsep}{3pt}
\resizebox{\columnwidth}{!}{%
\begin{tabular}{lrrrrr}
\toprule
Method & Views & kB/frame\,$\downarrow$ & s/frame\,$\downarrow$ & PSNR\,$\uparrow$ & LPIPS\,$\downarrow$ \\
\midrule
Ours Lite & 8 & 82--139 & 0.13--0.17 & 25.95 & 0.118 \\
Ours Pro & 8 & 166--296 & 0.13--0.18 & 26.30 & 0.107 \\
Ours Ultra & 8 & 252--455 & 0.13--0.18 & 26.38 & 0.104 \\
\midrule
HiCoM~\cite{hicom} & 8 & 2969--3994 & 2.0--2.8 & 17.68 & 0.389 \\
QUEEN-l~\cite{queen} & 8 & 50038--57680 & 1.2--1.4 & 15.91 & 0.399 \\
3DGStream~\cite{3dgstream} & 8 & 8019--8133 & 2.2--5.2 & 23.57 & 0.268 \\
4C4D~\cite{fourcfourd} (offline) & 8 & 3873--39878 & 30--124 & 24.68 & 0.184 \\
HiCoM~\cite{hicom} & 40 & 644--3309 & 2.1--2.4 & 19.60 & 0.346 \\
QUEEN-l~\cite{queen} & 40 & 1707--199177 & 4.3--23.4 & 20.67 & 0.313 \\
3DGStream~\cite{3dgstream} & 40 & 7916--8105 & 2.1--2.6 & 25.58 & 0.210 \\
\bottomrule
\end{tabular}}
\caption{\textbf{From the same eight views, every tier of \sys beats every per-frame optimisation method, and Pro beats offline 4C4D, in PSNR, LPIPS, and bytes on all \CmpDynSparseAttempted\ test sequences.} Official settings on DNA-Rendering frames 30--89 (\cref{sec:supp_dyn}).}
\label{tab:cmp_dyn}
\end{table}

\input{fig/qual}
\input{fig/production}
\paragraph{Per-frame optimisation.}
HiCoM~\cite{hicom}, QUEEN~\cite{queen}, and 3DGStream~\cite{3dgstream} spend \CmpDynFirstS\,s on frame zero and \CmpDynLaterS\,s per later frame on shared GPUs, against our \CmpDynOursLaterS\,s on one GPU, \CmpDynSpeedupMean$\times$ faster on average at 8 and 40 views. On an idle GPU the fastest requires \CmpIdleFitMinS\,s per frame, so even ideal scaling requires at least \CmpThirtyGPUsMin\ GPUs for 30\,FPS (\cref{tab:cmp_speed_idle}); newer methods~\cite{instantgs,morgs} take \CmpLitFastestS\,s or more (\cref{tab:cmp_lit}).

From the same eight views, every tier of \sys outperforms all three in PSNR, LPIPS, and bitrate on all \CmpDynSparseAttempted\ sequences (\cref{tab:cmp_dyn,fig:qual}; \cref{fig:teaser}, right) and gains 2.55--4.59\,dB in temporal-difference PSNR (\cref{sec:supp_temporal}); even with 40 views, all three trail eight-view Lite in mean PSNR. Offline 4C4D~\cite{fourcfourd}, fitted to all 60 frames at once, is outperformed by Pro on every sequence---by \CmpDynFourCProGain\,dB on average at a median \CmpDynFourCOverPro$\times$ Pro's payload size (\cref{tab:cmp_dyn_4c4d}).

\paragraph{3DGS compression methods.}
We apply SPZ v4~\cite{spz}, PlayCanvas compressed PLY and SOG, ksplat, Draco, and FCGS~\cite{fcgs} at default or recommended settings to our uncompressed Gaussians after the same NSC (\CmpUncKB\,kB/frame). The smallest, Draco, needs \CmpMinCodecOverPro$\times$ Pro's payload size, \BudgetFormatMinMbps\,Mbit/s at 30\,FPS (\cref{tab:cmp_formats}); our tiers save \ReviewNSCSavingMin--\ReviewNSCSavingMax\% in BD-rate over Draco, FCGS, and uniform quantisation (\cref{fig:review_rd}), and V$^3$-style x265~\cite{vcubed} peaks at \VidBestPSNR\,dB (\cref{tab:video_gaussians}).

\paragraph{Latent feature coding.}
On our eight views, the released static-scene model of CodecSplat~\cite{codecsplat} plateaus at \CmpCodecSplatTopPSNR\,dB at its top rate, \CmpCodecSplatLiteGap\,dB below Lite, and Pro outperforms it on every sequence, by \CmpCodecSplatProGap\,dB on average (\cref{tab:cmp_formats,tab:cmp_feature}). Its decoder takes \CmpCodecSplatRecvMs\,ms per frame on an RTX~5090, \CmpCodecSplatRecvOverBudget$\times$ the per-frame client budget, whereas our fused client decodes Pro in \CmpOursClientSameGPUMs\,ms on the same GPU, \CmpCodecSplatRecvRatio$\times$ faster (\cref{tab:cmp_same_gpu}).

\paragraph{Digital-human systems.}
Per-sequence human reconstruction methods fit \HumanSeqFitMin--\HumanSeqFitMax\ per frame (\cref{tab:cmp_human_speed}); HiReFF's \CmpHiReFFSameGPUFPS\,FPS on an RTX~5090 trails our single-GPU throughput (\cref{tab:cmp_same_gpu}), and H.265 Gaussian-map streams lose about 2\,dB to video coding~\cite{tele360}.

\subsection{Real-Time Frame-Parallel Streaming}
\label{sec:exp_production}
On one RTX~5090, a frame spends \StageMattingMs\,ms in matting and \StageReconMs\,ms in reconstruction, then \StageHandMs\,ms to prune, project, and packetise tokens and \StageEncodeMs\,ms to quantise and compress them (\cref{fig:production}a). On \GanttGPUsWord\ GPUs, all \GanttWorkersWord\ workers are active \GanttFullPct\% of a one-second window, releasing packets in capture order (\cref{fig:production}b). Across all \ProdSceneCount\ sequences, throughput scales from \SingleFPSMean\ frames/s on one GPU to \FourFPSMean\ on four (above 30\,FPS) and \SixFPSMean\ on six, \ScalingEffMin--\ScalingEffMax\% of ideal linear scaling (\cref{fig:production}c), with all 16 above 30\,FPS from five GPUs, where live 30\,FPS Wi-Fi streaming to the RTX~3050~Ti laptop averages \GoLiveRenderMinMs--\GoLiveRenderMaxMs\,ms end-to-end latency without drift (\cref{sec:supp_handoff}). Stateless packets also support random access at every frame, bit-exact tier switching, and single-frame loss recovery.

\subsection{Neural Session Calibration}
\label{sec:exp_nsc}
On HiFi4G and ActorsHQ, scored on held-out cameras after calibrating on the first second of eight input views (\cref{sec:supp_nsc}), plain PCA decoding loses up to \NSCCodecCostMax\,dB to our uncompressed reconstruction, and the hypernetwork trained on DNA-Rendering narrows this gap to at most \NSCHyperCodecCostMax\,dB (\cref{fig:nsc_gap,tab:nsc_gap}). Whereas a 12-parameter colour-affine transform changes PSNR by only \NSCColorHiFiPro\ and \NSCColorAHQPro\,dB, NSC adds \NSCHiFiDeployedIncrementPro\ and \NSCAHQDeployedIncrementPro\,dB at Pro, so even Lite beats the uncalibrated, unpruned reconstruction by \NSCHiFiLiteOverUnc\ and \NSCAHQLiteOverUnc\,dB; without further updates, NSC still adds \LongNSCDeltaLastHiFi\ and \LongNSCDeltaLastAHQ\,dB on the final evaluated frames of each sequence.

\section{Conclusion}
\label{sec:conclusion}
\sys meets the three per-frame budgets of live free-viewpoint video by splitting a feed-forward reconstructor before its Gaussian head into a multi-rate token codec with opacity-guided pruning, PCA dimensionality reduction, and lossless anchor coding. Pro transmits \CompressionRatio$\times$ less data than float32 Gaussians while a fused feature reconstruction hypernetwork and its one-time update recover \LDRecoveredPro\% of the compression loss in \LDLaptopClientPro\,ms on a laptop GPU, NSC fine-tunes the head online during warm-up with a \NSCKB\,kB update, and frame-parallel encoding averages \ThirtyFPSMean\,FPS on \ThirtyFPSGPUsWord\ GPUs.

\paragraph{Limitations.}
Frames are coded without tracking Gaussian identities, trading temporal correspondence for barrier-free parallelism, random access at every frame, and zero error propagation. Session setup from the first second of capture takes \SessionReadySeconds\,s on four GPUs before going live, at zero per-frame stream or client overhead.

\section*{LLM Usage Statement}
AI assistants (Anthropic Claude and OpenAI GPT models) assisted the authors in discussing research ideas and claims, searching the literature and drafting the related work, designing, coding and running the experiments and evaluation, generating the synthetic evaluation data and processing the data, analysing the results, generating the figures, and drafting and editing the text of the paper and the supplementary material.

{
    \small
    \bibliographystyle{ieeenat_fullname}
    \bibliography{main,supp}
}

\maketitlesupplementary
\setcounter{section}{0}\setcounter{figure}{0}\setcounter{table}{0}
\renewcommand{\theHsection}{S\arabic{section}}
\renewcommand{\theHfigure}{S\arabic{figure}}
\renewcommand{\theHtable}{S\arabic{table}}
\makeatletter
\setlength{\@fptop}{0pt}\setlength{\@fpbot}{0pt plus 1fil}
\setlength{\@dblfptop}{0pt}\setlength{\@dblfpbot}{0pt plus 1fil}
\renewcommand{\floatpagefraction}{0.5}\renewcommand{\dblfloatpagefraction}{0.5}
\makeatother
\renewcommand{\thetable}{S\arabic{table}}
\renewcommand{\thefigure}{S\arabic{figure}}
\renewcommand{\thesection}{S\arabic{section}}
\newcommand{\tablepair}[2]{%
  \begin{table*}[t]%
  \renewenvironment{table}[1][]{}{}%
  \begin{minipage}[t]{0.485\textwidth}\input{#1}\end{minipage}\hfill
  \begin{minipage}[t]{0.485\textwidth}\input{#2}\end{minipage}%
  \end{table*}}
\section{Additional quality results}
\label{sec:supp_quality}
\Cref{fig:showcase} shows two performers streamed live with the frozen reconstructor of the main paper. Our uncompressed reconstruction matches standard PointSplat512 (\UncDelta\,dB, \cref{tab:official_quality_full}). All tables in this section use $512\times512$ full-image reconstruction, soft masks from BiRefNet Lite fine-tuned on mask-annotated DNA-Rendering data, and a black background. \Cref{tab:backgrounds} scores the full image, \cref{tab:leakage} measures alpha and colour outside the ground-truth (GT) mask, \cref{tab:per_scene} lists per-scene PSNR and bytes, and \cref{tab:official_quality_full} adds client Gaussian counts.
\input{fig/showcase}
\begin{table*}[t]
\centering
\small
\setlength{\tabcolsep}{3.5pt}
\begin{tabular}{llrrr}
\toprule
Method & Setting & PSNR & SSIM & LPIPS \\
\midrule
PointSplat512 & black, full image & 33.786 & 0.9774 & 0.0208 \\
Uncompressed + NSC & black, full image & 35.042 & 0.9809 & 0.0174 \\
Ours Lite (32/4) & black, full image & 34.459 & 0.9784 & 0.0207 \\
Ours Pro (48/6) & black, full image & 34.837 & 0.9800 & 0.0185 \\
Ours Ultra (80/6) & black, full image & 34.892 & 0.9803 & 0.0182 \\
\bottomrule
\end{tabular}
\caption{\textbf{Every tier improves full-image PSNR, SSIM, and LPIPS over standard PointSplat512.} The predictions of \cref{tab:official_quality} scored on the full black-background image; streamed tiers use the hypernetwork and NSC.}
\label{tab:backgrounds}
\end{table*}

\begin{table*}[t]
\centering
\small
\setlength{\tabcolsep}{3.5pt}
\begin{tabular}{lrrr}
\toprule
Method & Mean $\alpha$ & $\alpha$ share & MAE on black \\
\midrule
PointSplat512 & 0.000279 & 0.002894 & 0.000102 \\
Ours uncompressed & 0.000279 & 0.002898 & 0.000102 \\
Ours Lite (32/4) & 0.000431 & 0.004564 & 0.000120 \\
Ours Pro (48/6) & 0.000372 & 0.003916 & 0.000104 \\
Ours Ultra (80/6) & 0.000368 & 0.003871 & 0.000103 \\
\bottomrule
\end{tabular}
\caption{\textbf{Outside-mask colour error remains at or below $1.2\times 10^{-4}$ across all tiers.} Mean alpha outside the GT mask, share of predicted alpha mass outside it, and outside-mask MAE on black; streamed tiers use the hypernetwork and NSC.}
\label{tab:leakage}
\end{table*}

\begin{table}[t]
\centering
\small
\setlength{\tabcolsep}{3.5pt}
\begin{tabular}{lrrrrr}
\toprule
Scene & PointSplat512 & \shortstack[r]{Uncom-\\pressed} & Pro & \shortstack[r]{$\Delta$\\PSNR} & \shortstack[r]{Pro\\kB} \\
\midrule
0012\_09 & 21.566 & 21.569 & 21.810 & +0.244 & 212.3 \\
0022\_10 & 27.707 & 27.707 & 29.392 & +1.685 & 239.0 \\
0166\_04 & 22.579 & 22.580 & 23.078 & +0.498 & 255.1 \\
0173\_02 & 22.426 & 22.426 & 22.505 & +0.079 & 206.2 \\
0219\_07 & 26.030 & 26.030 & 26.688 & +0.659 & 198.1 \\
0235\_11 & 24.632 & 24.632 & 25.592 & +0.960 & 252.3 \\
0326\_07 & 25.040 & 25.040 & 25.225 & +0.185 & 189.2 \\
0345\_08 & 26.686 & 26.686 & 27.780 & +1.094 & 209.7 \\
0379\_09 & 29.088 & 29.089 & 30.393 & +1.305 & 151.1 \\
0410\_05 & 28.695 & 28.695 & 29.867 & +1.173 & 193.4 \\
0438\_02 & 28.309 & 28.309 & 29.834 & +1.525 & 187.1 \\
0536\_06 & 29.351 & 29.351 & 31.660 & +2.308 & 162.4 \\
0647\_02 & 28.149 & 28.149 & 30.010 & +1.861 & 187.8 \\
0657\_05 & 28.568 & 28.568 & 30.862 & +2.293 & 168.0 \\
0718\_06 & 21.696 & 21.696 & 21.925 & +0.230 & 249.9 \\
0811\_06 & 28.177 & 28.177 & 28.879 & +0.702 & 269.2 \\
\bottomrule
\end{tabular}
\caption{\textbf{Pro improves PSNR over standard PointSplat512 on every one of the 16 test scenes.} Per-scene PSNR; Pro uses the hypernetwork and NSC, and $\Delta$ is Pro minus standard PointSplat512.}
\label{tab:per_scene}
\end{table}

\begin{table*}[t]
\centering
\small
\setlength{\tabcolsep}{3.5pt}
\begin{tabular}{lrrrrr}
\toprule
Method & \multicolumn{3}{c}{16 scenes} & kB $\downarrow$ & Gaussians (k) $\downarrow$ \\ 
 & PSNR $\uparrow$ & SSIM $\uparrow$ & LPIPS $\downarrow$ & & \\
\midrule
PointSplat512 & 26.169 & 0.8650 & 0.1034 & 4753.1$^*$ & 84.9 \\
Ours uncompressed & 26.169 & 0.8650 & 0.1034 & 5941.4$^*$ & 106.1 \\
Ours Lite (32/4) & 26.840 & 0.8700 & 0.1043 & 94.8 & 64.8 \\
Ours Pro (48/6) & 27.219 & 0.8802 & 0.0941 & 208.2 & 64.8 \\
Ours Ultra (80/6) & 27.273 & 0.8818 & 0.0923 & 321.9 & 64.8 \\
\bottomrule
\end{tabular}
\caption{\textbf{Our tiers exceed PointSplat512 in PSNR and SSIM at a fraction of its Gaussian bytes.} DNA-Rendering, 16 test scenes; all tiers use the hypernetwork and NSC (\cref{tab:nsc_dna} gives the plain decoder). $^*$Float32 storage of 14 attributes per Gaussian.}
\label{tab:official_quality_full}
\end{table*}

\section{Codec and pruning controls}
\label{sec:supp_controls}
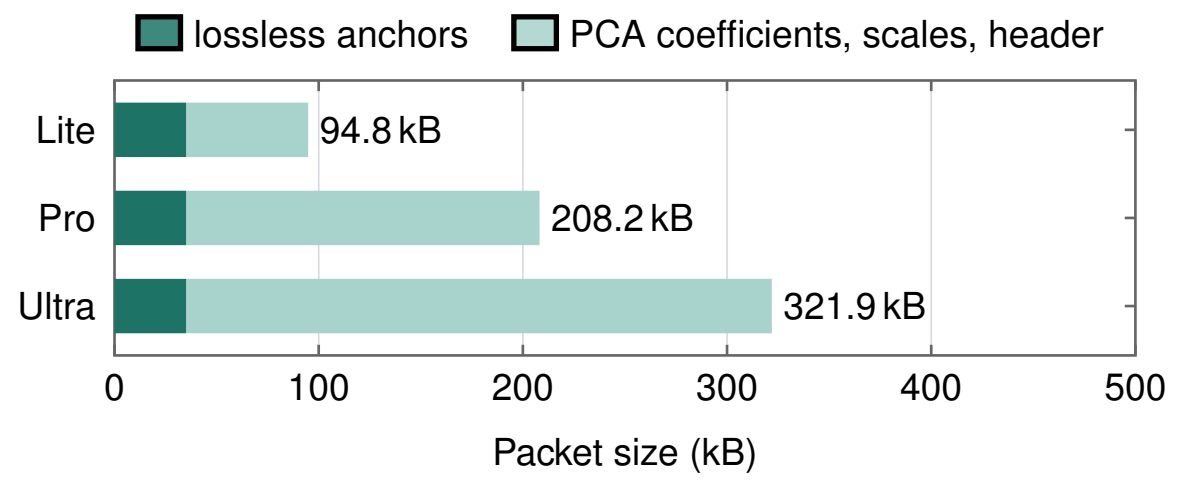
\begin{figure}[t]
\centering
\begin{tikzpicture}
\begin{axis}[paper,xbar stacked,width=\columnwidth,height=3.4cm,bar width=0.36cm,xmin=0,xmax=500,clip=false,symbolic y coords={Ultra,Pro,Lite},ytick=data,enlarge y limits=0.28,xlabel={Packet size (kB)},legend style={font=\sffamily\scriptsize,at={(0.5,1.03)},anchor=south,legend columns=2,/tikz/every even column/.append style={column sep=6pt}},legend cell align=left,ymajorgrids=false]
\addplot[fill=pkt!80!black,draw=none] coordinates {(35.048,Lite) (35.048,Pro) (35.048,Ultra)};\addlegendentry{lossless anchors}
\addplot[fill=pkt!40,draw=none] coordinates {(59.786,Lite) (173.126,Pro) (286.896,Ultra)};\addlegendentry{PCA coefficients, scales, header}
\node[anchor=west,inner sep=2pt,font=\sffamily\scriptsize] at (axis cs:94.834,Lite) {94.8\,kB};
\node[anchor=west,inner sep=2pt,font=\sffamily\scriptsize] at (axis cs:208.174,Pro) {208.2\,kB};
\node[anchor=west,inner sep=2pt,font=\sffamily\scriptsize] at (axis cs:321.944,Ultra) {321.9\,kB};
\end{axis}
\end{tikzpicture}
\caption{\textbf{All tiers spend the same 35.0\,kB on lossless anchors and differ in their PCA coefficients.} Average packet size over the 16 DNA-Rendering test scenes. The float32 PCA state is sent once per session (136\,kB for Lite, 201\,kB for Pro, 333\,kB for Ultra); float32 Gaussians of standard PointSplat512 take 4753\,kB per frame.}\label{fig:packet_bytes}
\end{figure}

\begin{figure}[t]
\centering
\begin{tikzpicture}
\begin{axis}[paper,width=\columnwidth,height=5cm,xlabel={Retained token fraction},ylabel={PSNR (dB)},legend pos=south east]
\addplot+[mark=*] coordinates {(0.8,26.14507) (0.85,26.16256) (0.9,26.16811) (0.95,26.16892)};\addlegendentry{Learned}
\addplot+[mark=*] coordinates {(0.8,26.11209) (0.85,26.16226) (0.9,26.18504) (0.95,26.18769)};\addlegendentry{Random}
\addplot+[mark=*] coordinates {(0.8,26.14541) (0.85,26.16315) (0.9,26.16741) (0.95,26.16879)};\addlegendentry{Exact opacity}
\end{axis}
\end{tikzpicture}
\caption{\textbf{At the deployed 76.4\% retention, learned token selection beats random selection by 0.201\,dB (\cref{tab:filter_controls}), and it stays within 0.001\,dB of exact-opacity ranking at every plotted fraction.} Uncompressed outputs without client pruning, keeping a fixed fraction of token groups in every frame rather than thresholding; \cref{tab:filter_controls} compares the three selections at the token count of the deployed threshold, followed by 20\% client pruning.}\label{fig:token_keep}
\end{figure}
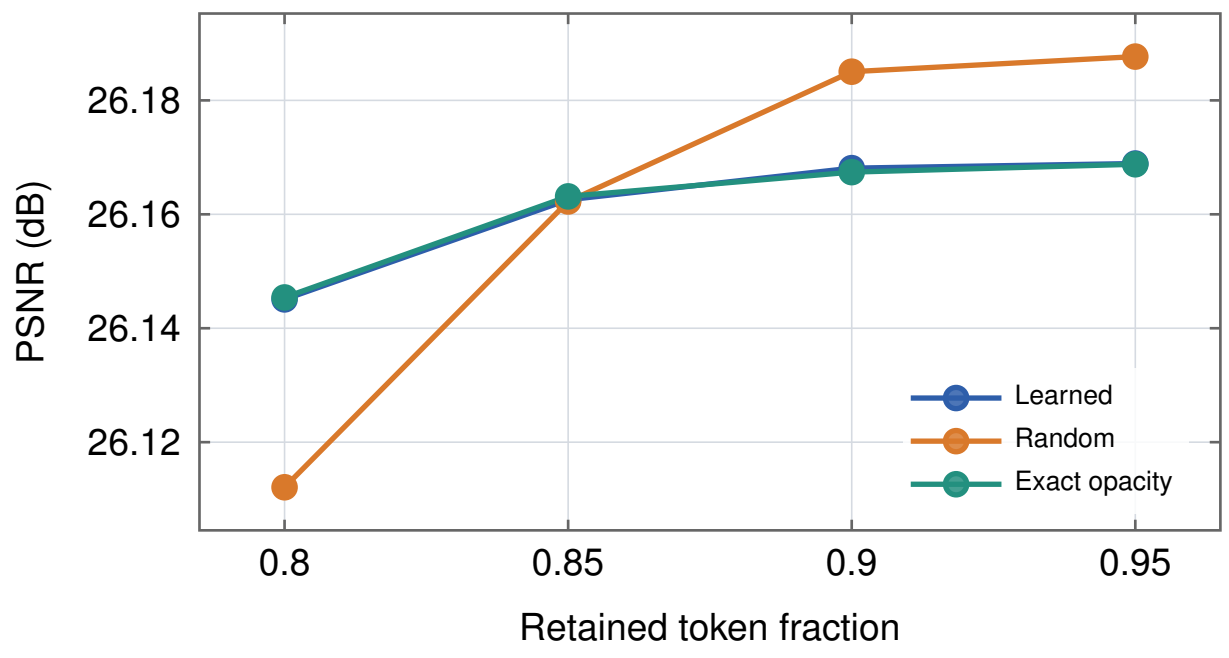

Token pruning removes token groups and their bytes, whereas client pruning removes Gaussians at an unchanged packet size (\cref{fig:filtering,tab:server_filter}). \Cref{tab:filter_controls} compares the learned group score with exact-opacity and random selection at matched token counts, and \cref{fig:token_keep} sweeps the retained fraction of tokens. \Cref{tab:codec_controls} compares PCA coding with SPZ and product quantisation. Fusing the PCA basis into the head (\cref{eq:fused}) changes PSNR by \FuseDelta\,dB at $(r,b){=}(80,6)$; \cref{tab:server_filter,tab:two_layer_full} and \cref{fig:cut} therefore decode plain packets with the fused head, and the other plain rows decode without it. All feature codecs share the lossless anchor stream, which \cref{fig:packet_bytes} separates from the coefficients of each tier.
\input{fig/filtering}
\begin{table}[t]
\centering
\small
\setlength{\tabcolsep}{3.5pt}
\begin{tabular}{llrrrr}
\toprule
Codec & Pruning & PSNR & kB & Tokens & \shortstack[r]{Gaussians\\(k)} \\
\midrule
Uncompressed & none & 26.169 & 5941.4 & 6631 & 106.1 \\
Uncompressed & client & 26.169 & 4753.1 & 6631 & 84.9 \\
Uncompressed & token & 26.181 & 4538.5 & 5065 & 81.0 \\
Uncompressed & both & 26.183 & 3630.7 & 5065 & 64.8 \\
PCA 80/6 & client & 25.894 & 420.7 & 6631 & 84.9 \\
PCA 80/6 & token & 25.900 & 321.9 & 5065 & 81.0 \\
PCA 80/6 & both & 25.902 & 321.9 & 5065 & 64.8 \\
\bottomrule
\end{tabular}
\caption{\textbf{Token pruning and client pruning reduce different costs.} Token pruning removes token groups and shrinks the packet; client pruning removes 20\% of the generated Gaussians at unchanged packet size. Uncompressed kB counts raw float32 Gaussians; plain decoder.}
\label{tab:server_filter}
\end{table}

\begin{table}[t]
\centering
\small
\setlength{\tabcolsep}{3.5pt}
\begin{tabular}{lrrrr}
\toprule
Method & PSNR & SSIM & LPIPS & kB \\
\midrule
Uncompressed & 26.169 & 0.8650 & 0.1034 & 4753.1 \\
SPZ v4 (SH0) & 26.165 & 0.8639 & 0.1087 & 1089.3 \\
Product quantisation & 25.712 & 0.8388 & 0.1463 & 356.9 \\
Lite, plain & 24.950 & 0.8198 & 0.1809 & 94.8 \\
Lite, hypernetwork & 25.772 & 0.8491 & 0.1239 & 94.8 \\
Lite, hypernetwork\,+\,NSC & 26.840 & 0.8700 & 0.1043 & 94.8 \\
Pro, plain & 25.582 & 0.8419 & 0.1459 & 208.2 \\
Pro, hypernetwork & 26.065 & 0.8598 & 0.1121 & 208.2 \\
Pro, hypernetwork\,+\,NSC & 27.219 & 0.8802 & 0.0941 & 208.2 \\
Ultra, plain & 25.904 & 0.8528 & 0.1256 & 321.9 \\
Ultra, hypernetwork & 26.095 & 0.8614 & 0.1097 & 321.9 \\
Ultra, hypernetwork\,+\,NSC & 27.273 & 0.8818 & 0.0923 & 321.9 \\
\bottomrule
\end{tabular}
\caption{\textbf{Pro with the hypernetwork improves all three metrics over product quantisation (PQ) with fewer bytes, while Ultra uses \ControlSPZOverUltra$\times$ fewer bytes than SPZ; adding NSC lifts all three tiers above both.} SPZ compresses our uncompressed Gaussians without token pruning after 20\% client pruning, whereas \cref{tab:cmp_formats_nonsc} applies it after token pruning; PQ uses 64 subvectors of dimension 16 and 256 centroids with Pro's token pruning. Plain and hypernetwork rows decode the same tokens.}
\label{tab:codec_controls}
\end{table}

\begin{table}[t]
\centering
\small
\setlength{\tabcolsep}{3.5pt}
\begin{tabular}{lrrr}
\toprule
Group selection & PSNR & Gaussians (k) & Tokens \\
\midrule
Client pruning only & 26.169 & 84.9 & 6631 \\
Random token groups & 25.982 & 64.8 & 5065 \\
Exact-opacity groups & 26.177 & 64.8 & 5065 \\
Learned token groups & 26.183 & 64.8 & 5065 \\
\bottomrule
\end{tabular}
\caption{\textbf{Learned group selection matches exact-opacity ranking and clearly beats random selection.} Matched token counts without feature compression, followed by 20\% client pruning. Exact-opacity ranking requires running the full head.}
\label{tab:filter_controls}
\end{table}

\paragraph{Entropy and inter-frame coding controls.}
Intra-frame coding produces smaller packets than inter-frame coding and keeps every packet stateless. On the 16 DNA-Rendering test scenes (frames 1--30, bit-exact round trips of the real Lite/Pro/Ultra packets), matching tokens across adjacent frames on a 1\,cm world voxel grid and delta-coding the matched integer PCA codes saves $1.82\%$/$0.59\%$/$-0.51\%$ of feature bytes at Lite/Pro/Ultra. Unordered surface groups, however, need an explicit inter-frame token permutation, which increases geometry bytes by $31.6\%$ and enlarges whole packets by $6.12\%$/$4.98\%$/$4.00\%$. Causal 8\,kB Zstd~\cite{zstd} dictionaries trained on frame~0 and rebuilt deterministically on the client at zero wire bytes save $0.19\%$/$0.14\%$/$0.09\%$ of whole-packet bytes across frames 1--30. For entropy coding (frames 10, 20 and 30), bit-exact 32-bit byte-oriented range asymmetric numeral system (rANS) coding with frame-zero probability tables saves $9.35\%$ of whole-packet bytes at Pro with static order-0 models ($5.00$ versus $5.65$\,bits/coefficient for Zstd; $-0.59\%$ at Lite and $10.42\%$ at Ultra), and intra-token prefix-energy conditioning adds at most $0.03$\,bits/coefficient. Single-core CPU rANS decoding takes $2.38$\,ms per Pro frame against \LDLaptopDecodePro\,ms for Zstd unpacking and upload on the laptop, so we keep Zstd and this stage stays below one millisecond.

\paragraph{Token pruning training and selection.}
The predictor normalises each input dimension with a standard-deviation floor of 0.05. Its labels come from the uncompressed head, without PCA or quantisation. We regress the 16 slot-opacity logits with a smooth-$\ell_1$ loss plus a loss on the largest logit weighted by 0.5, on all 384,000 tokens of 50 frames of each of 0008\_01, 0019\_06, 0123\_02, 0206\_04 and 0219\_07 (AdamW, learning rate $3\times10^{-3}$, weight decay $10^{-4}$, batch size 8192, 48 epochs), with checkpoint selection on 0165\_08 and 0235\_11. The deployed model and threshold are chosen from 53 candidates. Each candidate must lose at most 0.02\,dB against the uncompressed reconstruction with 20\% client pruning on 0219\_07 and 0235\_11, and among those within 1\% of the lowest token count we select the highest PSNR on these two scenes.

\section{Rate--quality and pruning study}
\label{sec:supp_tiers}
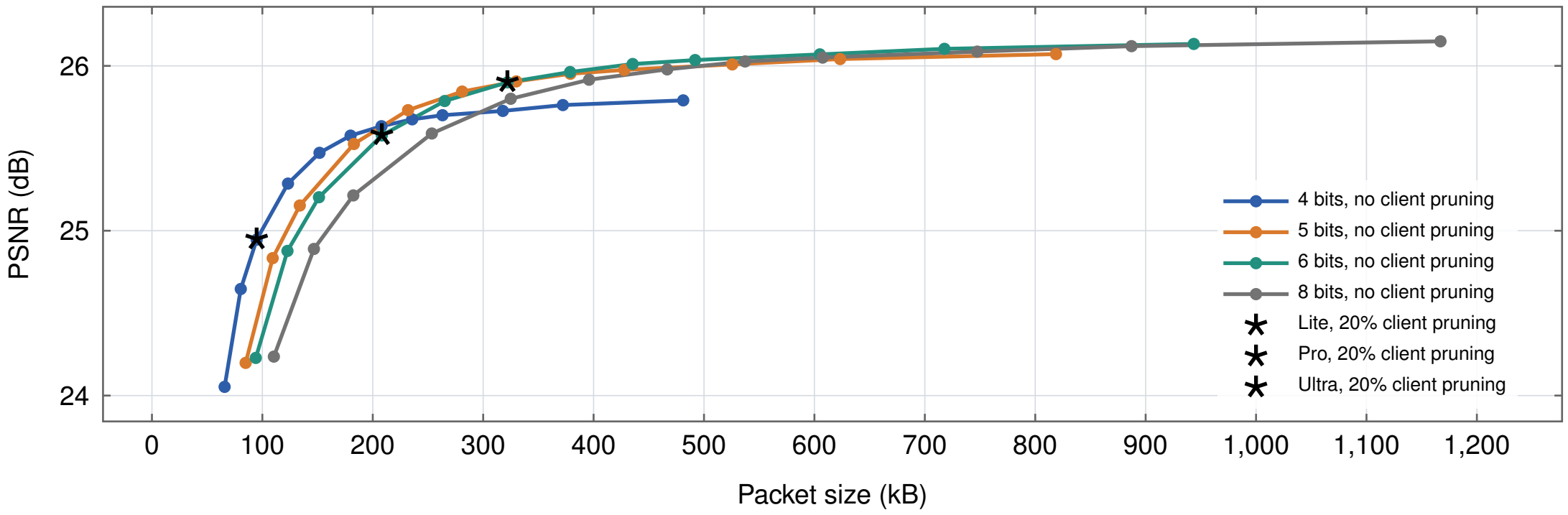
\begin{figure*}[t]
\centering
\begin{tikzpicture}
\begin{axis}[paper,width=0.86\textwidth,height=5.4cm,xlabel={Packet size (kB)},ylabel={PSNR (dB)},legend pos=south east]
\addplot+[mark=*,mark size=1.1pt] coordinates {(65.816,24.05277) (80.440,24.64648) (94.835,24.94579) (123.225,25.28595) (151.759,25.47259) (179.889,25.57693) (207.910,25.63411) (235.700,25.67568) (263.158,25.70063) (317.813,25.72676) (372.177,25.76223) (481.178,25.79072)};\addlegendentry{4 bits, no client pruning}
\addplot+[mark=*,mark size=1.1pt] coordinates {(84.856,24.19823) (109.363,24.83368) (133.909,25.15301) (182.907,25.52611) (231.956,25.73143) (280.954,25.84418) (329.964,25.90577) (378.943,25.95203) (427.851,25.97555) (525.605,26.00889) (623.298,26.04134) (818.797,26.07138)};\addlegendentry{5 bits, no client pruning}
\addplot+[mark=*,mark size=1.1pt] coordinates {(94.045,24.22841) (122.725,24.87734) (151.274,25.20314) (208.175,25.57796) (265.154,25.78676) (321.944,25.89982) (378.707,25.96265) (435.381,26.01141) (491.927,26.03528) (604.914,26.06981) (717.757,26.10345) (943.572,26.13307)};\addlegendentry{6 bits, no client pruning}
\addplot+[mark=*,mark size=1.1pt] coordinates {(110.415,24.23634) (146.696,24.88945) (182.382,25.21419) (253.490,25.58984) (324.960,25.80046) (395.899,25.91598) (466.672,25.97871) (537.227,26.02677) (607.439,26.05024) (747.505,26.08653) (887.212,26.11964) (1167.027,26.14887)};\addlegendentry{8 bits, no client pruning}
\addplot[only marks,mark=star,mark size=3pt,black] coordinates {(94.834,24.94981)};\addlegendentry{Lite, 20\% client pruning}
\addplot[only marks,mark=star,mark size=3pt,black] coordinates {(208.174,25.58225)};\addlegendentry{Pro, 20\% client pruning}
\addplot[only marks,mark=star,mark size=3pt,black] coordinates {(321.944,25.90378)};\addlegendentry{Ultra, 20\% client pruning}
\end{axis}
\end{tikzpicture}
\caption{\textbf{Six-bit quantisation nearly matches 8 bits while client pruning shifts the frontier upward at unchanged bytes.} 48 PCA settings without client pruning; stars mark the deployed tiers with 20\% client pruning. All points use the plain decoder.}\label{fig:tier_curve}
\end{figure*}
Pre-head tokens tolerate coding error that Gaussian attributes cannot. \Cref{fig:tier_curve} scans 48 rank and bit-width settings, from which Lite is 32/4, Pro 48/6 and Ultra 80/6, and \cref{tab:two_layer_full} crosses token pruning, codec and client pruning. To isolate why high attribute variance fails to preserve rendering quality (\cref{sec:exp_cut}), we perturb the kept groups of frame~30 at each split point on four scenes (0012\_09, 0345\_08, 0536\_06, 0811\_06; 48 cameras) before 20\% client pruning. To a split representation $x$, the tokens $T$, hidden features $H$ or the 224 pre-activation attributes $A$ of each kept group, we add $\sqrt{\rho}\,z$, where $z$ is a fixed Gaussian direction scaled so that $\lVert z\rVert^2=\sum_{i,c}(x_{ic}-\bar{x}_c)^2$, the energy of $x$ about its per-channel mean over the kept groups $i$. The perturbation thus carries a fraction $\rho\in\{0.1\%,1\%,3\%,10\%,30\%\}$ of each representation's own centred energy. At $\rho{=}1\%$, perturbing $T$ or $H$ changes PSNR by less than 0.01\,dB. At the same relative perturbation level $\rho$, perturbing $A$ drops PSNR by 16.38\,dB ($26.46\to 10.08$\,dB), and the MSE between perturbed and uncompressed renders, averaged over cameras and scenes, is $14{,}446\times$ that of perturbing $T$. Log-scales dominate this failure. Coding individual attribute types at $(r,b){=}(80,6)$, lossy scales alone drop PSNR to 11.25\,dB, against 21.81--24.68\,dB when only another attribute type is lossy; keeping only the scales exact while coding the other four types lossily restores 19.75\,dB, against 14.03\,dB with all types lossy. The exponential $\exp(\text{log-scale})$ turns small pre-activation errors into large covariance and screen-coverage shifts. Even at $(160,8)$, lossy scales alone drop PSNR from 26.46 to 22.55\,dB (25.69--26.41\,dB for any other single attribute type), so attribute sensitivity is structural rather than a low-bit artefact.
\begin{table*}[t]
\centering
\small
\setlength{\tabcolsep}{3.5pt}
\begin{tabular}{lllrrrrr}
\toprule
Token threshold & Rank/bits & Client pruning & PSNR & SSIM & LPIPS & kB & Gaussians (k) \\
\midrule
None & 64/6 & 20\% & 25.783 & 0.8492 & 0.1314 & 346.4 & 84.9 \\
None & 64/6 & 30\% & 25.786 & 0.8492 & 0.1315 & 346.4 & 74.3 \\
None & 64/6 & none & 25.784 & 0.8491 & 0.1314 & 346.4 & 106.1 \\
None & 80/4 & 20\% & 25.568 & 0.8429 & 0.1341 & 235.5 & 84.9 \\
None & 80/4 & 30\% & 25.573 & 0.8429 & 0.1343 & 235.5 & 74.3 \\
None & 80/4 & none & 25.569 & 0.8429 & 0.1341 & 235.5 & 106.1 \\
None & 80/6 & 20\% & 25.894 & 0.8532 & 0.1243 & 420.7 & 84.9 \\
None & 80/6 & 30\% & 25.897 & 0.8532 & 0.1245 & 420.7 & 74.3 \\
None & 80/6 & none & 25.894 & 0.8532 & 0.1243 & 420.7 & 106.1 \\
None & 96/6 & 20\% & 25.955 & 0.8557 & 0.1200 & 494.9 & 84.9 \\
None & 96/6 & 30\% & 25.959 & 0.8557 & 0.1202 & 494.9 & 74.3 \\
None & 96/6 & none & 25.956 & 0.8557 & 0.1200 & 494.9 & 106.1 \\
None & 128/6 & 20\% & 26.027 & 0.8584 & 0.1150 & 643.1 & 84.9 \\
None & 128/6 & 30\% & 26.031 & 0.8584 & 0.1152 & 643.1 & 74.3 \\
None & 128/6 & none & 26.028 & 0.8584 & 0.1150 & 643.1 & 106.1 \\
None & 160/6 & 20\% & 26.061 & 0.8600 & 0.1122 & 791.0 & 84.9 \\
None & 160/6 & 30\% & 26.064 & 0.8600 & 0.1124 & 791.0 & 74.3 \\
None & 160/6 & none & 26.061 & 0.8600 & 0.1122 & 791.0 & 106.1 \\
0.075 & 64/6 & 20\% & 25.789 & 0.8487 & 0.1329 & 265.2 & 64.8 \\
0.075 & 64/6 & 30\% & 25.790 & 0.8486 & 0.1331 & 265.2 & 56.7 \\
0.075 & 64/6 & none & 25.787 & 0.8486 & 0.1327 & 265.2 & 81.0 \\
0.075 & 80/4 & 20\% & 25.580 & 0.8426 & 0.1356 & 179.9 & 64.8 \\
0.075 & 80/4 & 30\% & 25.586 & 0.8426 & 0.1359 & 179.9 & 56.7 \\
0.075 & 80/4 & none & 25.577 & 0.8425 & 0.1355 & 179.9 & 81.0 \\
0.075 & 80/6 & 20\% & 25.902 & 0.8528 & 0.1258 & 321.9 & 64.8 \\
0.075 & 80/6 & 30\% & 25.905 & 0.8527 & 0.1260 & 321.9 & 56.7 \\
0.075 & 80/6 & none & 25.900 & 0.8527 & 0.1257 & 321.9 & 81.0 \\
0.075 & 96/6 & 20\% & 25.964 & 0.8553 & 0.1214 & 378.7 & 64.8 \\
0.075 & 96/6 & 30\% & 25.968 & 0.8552 & 0.1218 & 378.7 & 56.7 \\
0.075 & 96/6 & none & 25.963 & 0.8553 & 0.1213 & 378.7 & 81.0 \\
0.075 & 128/6 & 20\% & 26.037 & 0.8581 & 0.1164 & 491.9 & 64.8 \\
0.075 & 128/6 & 30\% & 26.042 & 0.8580 & 0.1167 & 491.9 & 56.7 \\
0.075 & 128/6 & none & 26.035 & 0.8580 & 0.1162 & 491.9 & 81.0 \\
0.075 & 160/6 & 20\% & 26.072 & 0.8597 & 0.1136 & 604.9 & 64.8 \\
0.075 & 160/6 & 30\% & 26.077 & 0.8597 & 0.1139 & 604.9 & 56.7 \\
0.075 & 160/6 & none & 26.070 & 0.8596 & 0.1134 & 604.9 & 81.0 \\
0.075 & 32/4 & 20\%, hypernetwork & 25.772 & 0.8491 & 0.1239 & 94.8 & 64.8 \\
0.075 & 32/4 & 20\%, hypernetwork + NSC & 26.840 & 0.8700 & 0.1043 & 94.8 & 64.8 \\
0.075 & 48/6 & 20\%, hypernetwork & 26.065 & 0.8598 & 0.1121 & 208.2 & 64.8 \\
0.075 & 48/6 & 20\%, hypernetwork + NSC & 27.219 & 0.8802 & 0.0941 & 208.2 & 64.8 \\
0.075 & 80/6 & 20\%, hypernetwork & 26.095 & 0.8614 & 0.1097 & 321.9 & 64.8 \\
0.075 & 80/6 & 20\%, hypernetwork + NSC & 27.273 & 0.8818 & 0.0923 & 321.9 & 64.8 \\
\bottomrule
\end{tabular}
\caption{\textbf{Token pruning saves \PruneBDRate\% of the bytes at equal PSNR.} Token-pruning threshold, codec and client pruning rules on the 16 test scenes; the hypernetwork and NSC are also shown for Lite, Pro and Ultra with 20\% client pruning. The saving is the Bj{\o}ntegaard delta rate (BD-rate)~\cite{bjontegaard} over six codecs with 20\% client pruning and the plain decoder.}
\label{tab:two_layer_full}
\end{table*}

\section{Resolution and temporal behaviour}
\label{sec:supp_temporal}
\Cref{tab:resolution} varies only the reconstruction input among 224, 256, 384 and 512, which changes the point and token population; rendering and scoring stay at $512\times512$. The temporal study replays three sequences over frames 30--149 with eight input and four held-out cameras. Each tier fits its PCA state on the complete tokens of frame zero and keeps it, and a second Pro variant without token pruning isolates its share. Over the last 30 frames, the full-image PSNR of Pro with the hypernetwork stays within \DriftLastWithin\,dB of the uncompressed reconstruction (\cref{fig:codec_drift,tab:drift_summary}).
\begin{table*}[t]
\centering
\small
\setlength{\tabcolsep}{4pt}
\begin{tabular}{r l rrrrr}
\toprule
Input (px) & Configuration & PSNR & SSIM & LPIPS & kB & Gaussians (k) \\
\midrule
224 & Uncompressed & 24.025 & 0.7831 & 0.1882 & 5622.4$^*$ & 100.4 \\
224 & Token + client pruning & 24.013 & 0.7824 & 0.1896 & 3210.0$^*$ & 57.3 \\
224 & Pro, plain & 23.456 & 0.7668 & 0.2315 & 187.9 & 57.3 \\
224 & Pro, hypernetwork & 23.781 & 0.7747 & 0.2086 & 187.9 & 57.3 \\
224 & Pro, hypernetwork + NSC & 24.083 & 0.7878 & 0.1921 & 187.9 & 57.3 \\
256 & Uncompressed & 24.627 & 0.8040 & 0.1642 & 5720.7$^*$ & 102.2 \\
256 & Token + client pruning & 24.617 & 0.8033 & 0.1655 & 3298.2$^*$ & 58.9 \\
256 & Pro, plain & 24.021 & 0.7848 & 0.2112 & 193.7 & 58.9 \\
256 & Pro, hypernetwork & 24.416 & 0.7955 & 0.1839 & 193.7 & 58.9 \\
256 & Pro, hypernetwork + NSC & 24.880 & 0.8115 & 0.1663 & 193.7 & 58.9 \\
384 & Uncompressed & 25.961 & 0.8527 & 0.1170 & 5920.0$^*$ & 105.7 \\
384 & Token + client pruning & 25.969 & 0.8523 & 0.1186 & 3514.6$^*$ & 62.8 \\
384 & Pro, plain & 25.236 & 0.8276 & 0.1654 & 204.0 & 62.8 \\
384 & Pro, hypernetwork & 25.798 & 0.8453 & 0.1302 & 204.0 & 62.8 \\
384 & Pro, hypernetwork + NSC & 26.696 & 0.8652 & 0.1107 & 204.0 & 62.8 \\
512 & Uncompressed & 26.169 & 0.8650 & 0.1034 & 5941.4$^*$ & 106.1 \\
512 & Token + client pruning & 26.183 & 0.8649 & 0.1048 & 3630.7$^*$ & 64.8 \\
512 & Pro, plain & 25.582 & 0.8419 & 0.1459 & 208.2 & 64.8 \\
512 & Pro, hypernetwork & 26.065 & 0.8598 & 0.1121 & 208.2 & 64.8 \\
512 & Pro, hypernetwork + NSC & 27.219 & 0.8802 & 0.0941 & 208.2 & 64.8 \\
\bottomrule
\end{tabular}
\caption{\textbf{Quality rises monotonically with input resolution to 512 at minimal added bytes.} Only the reconstruction input size changes; rendering and scoring stay at 512. Pro fits its PCA on frame 0 at each input size and reuses the hypernetwork update and NSC of the 512-input session. $^*$Raw float32 Gaussians.}
\label{tab:resolution}
\end{table*}

\input{fig/codec_drift}
\begin{table*}[t]
\centering
\small
\setlength{\tabcolsep}{4pt}
\begin{tabular}{llrrrrrr}
\toprule
Scene & Configuration & PSNR & SSIM & LPIPS & kB & $\Delta$, last 30 frames & Slope (dB/100 frames) \\
\midrule
0012\_09 & Uncompressed & 21.779 & 0.8099 & 0.1053 & 5236.8$^*$ & -- & -- \\
0012\_09 & Lite, plain & 20.176 & 0.7426 & 0.1806 & 93.6 & $-$1.246 & +0.584 \\
0012\_09 & Pro, plain & 21.030 & 0.7757 & 0.1404 & 207.7 & $-$0.516 & +0.380 \\
0012\_09 & Ultra, plain & 21.340 & 0.7901 & 0.1248 & 321.6 & $-$0.334 & +0.114 \\
0012\_09 & Pro, plain, no token pruning & 21.058 & 0.7763 & 0.1399 & 239.6 & $-$0.472 & +0.424 \\
0012\_09 & Pro, hypernetwork & 21.608 & 0.8030 & 0.1145 & 207.7 & $-$0.164 & $-$0.073 \\
0022\_10 & Uncompressed & 27.471 & 0.8974 & 0.1145 & 7148.7$^*$ & -- & -- \\
0022\_10 & Lite, plain & 27.216 & 0.8789 & 0.1846 & 112.1 & $-$0.395 & $-$0.176 \\
0022\_10 & Pro, plain & 27.441 & 0.8898 & 0.1593 & 246.2 & $-$0.128 & $-$0.176 \\
0022\_10 & Ultra, plain & 27.602 & 0.8955 & 0.1401 & 380.1 & +0.054 & $-$0.194 \\
0022\_10 & Pro, plain, no token pruning & 27.435 & 0.8898 & 0.1586 & 327.6 & $-$0.134 & $-$0.172 \\
0022\_10 & Pro, hypernetwork & 27.500 & 0.8967 & 0.1249 & 246.2 & $-$0.033 & $-$0.079 \\
0166\_04 & Uncompressed & 24.449 & 0.8376 & 0.0911 & 6741.8$^*$ & -- & -- \\
0166\_04 & Lite, plain & 23.096 & 0.7782 & 0.1568 & 117.2 & $-$1.364 & +0.039 \\
0166\_04 & Pro, plain & 23.787 & 0.8074 & 0.1256 & 258.3 & $-$0.641 & +0.010 \\
0166\_04 & Ultra, plain & 24.134 & 0.8208 & 0.1103 & 400.0 & $-$0.295 & $-$0.017 \\
0166\_04 & Pro, plain, no token pruning & 23.776 & 0.8074 & 0.1248 & 307.6 & $-$0.653 & +0.018 \\
0166\_04 & Pro, hypernetwork & 24.234 & 0.8258 & 0.1021 & 258.3 & $-$0.251 & $-$0.102 \\
\bottomrule
\end{tabular}
\caption{\textbf{A frame-zero basis holds over 120 frames.} Quality in the ground-truth mask bounding box over frames 30--149 with the plain decoder and the hypernetwork; the last two columns give the full-image PSNR difference to the uncompressed reconstruction over frames 120--149 and the slope of that difference over frames 30--149, per 100 frames. $^*$Raw Gaussian storage.}
\label{tab:drift_summary}
\end{table*}

\paragraph{Long sequences and refreshing the session state.}
The deployed system keeps its frame-zero state for the whole session. We extend five DNA-Rendering test sequences to frame 224 and follow the HiFi4G and ActorsHQ sequences to frame 149, with the conversion steps of the main protocol. With the frame-zero PCA basis and the hypernetwork update (its one-time fine-tuning on frames 0--29, \cref{sec:supp_handoff}) frozen, Pro changes by \LongNoRefreshLastFirstNine\,dB relative to the uncompressed reconstruction between the first and the last 30 evaluated frames, averaged over the \LongSeqCount\ sequences, and its per-sequence trend lies between \LongNoRefreshSlopeMin\ and \LongNoRefreshSlopeMax\,dB per 100 frames (\cref{fig:long_drift}). NSC, fitted once on frames 0--29 and never updated, keeps its gain over the hypernetwork alone to the end of the unseen-rig sequences, adding \LongNSCDeltaLastHiFi\ and \LongNSCDeltaLastAHQ\,dB over the last 30 evaluated frames on HiFi4G and ActorsHQ. Refreshing the basis every 120 frames adds \LongRefreshPCAOneTwentyAddedBytesNine\,kB per frame and changes PSNR by \LongRefreshPCAOneTwentyDeltaNine\,dB; refitting the hypernetwork update at each refresh as well leaves a net change of \LongLoRAOneTwentyDeltaFive\,dB (\cref{tab:long_refresh}).

\paragraph{Frame-to-frame stability.}
Deployed Pro follows the motion between frames more closely than every per-frame optimisation baseline. We replay frames 40--70 of 0012\_09, 0022\_10 and 0166\_04 on eight held-out cameras (30 adjacent frame pairs per camera, 720 pairs per variant). On signed inter-frame RGB differences within the union of adjacent GT masks, deployed Pro reaches 24.27\,dB temporal-difference PSNR (23.98\,dB for plain Pro and 24.48\,dB for uncompressed tokens), 2.55, 3.68 and 4.59\,dB above 3DGStream (21.71\,dB), QUEEN-l (20.59\,dB) and HiCoM (19.67\,dB). On static pixels (both frames' $\alpha{>}0.5$ and GT $|\Delta\text{RGB}|{<}2/255$; GT floor $0.855/255$), deployed Pro averages $3.44/255$ absolute inter-frame change, reducing static flicker by 46.5\% against HiCoM ($6.44/255$) and 57.4\% against 3DGStream ($8.07/255$).
\input{fig/long_drift}
\begin{table}[t]
\centering
\small
\begin{tabular}{lrr}
\toprule
Refresh policy & $\Delta$PSNR (dB) & $\Delta$bytes/frame \\
\midrule
\multicolumn{3}{l}{(a) PCA basis} \\
No refresh & +0.0000 & 0.0 \\
Every 60 frames & $-$0.1750 & 3,198.0 \\
Every 120 frames & $-$0.0740 & 1,359.4 \\
Variance $<$ 0.90 & $-$0.1282 & 4,160.8 \\
Variance $<$ 0.95 & $-$0.1853 & 23,603.0 \\
Variance $<$ 0.98 & $-$0.1830 & 84,009.9 \\
\midrule
\multicolumn{3}{l}{(b) PCA basis + hypernetwork update} \\
No refresh (PCA) & +0.0000 & 0.0 \\
Every 120 frames (PCA) & $-$0.0571 & 1,413.0 \\
Every 60 frames + update & +0.0002 & 12,869.8 \\
Every 120 frames + update & +0.0038 & 5,551.6 \\
Variance $<$ 0.90 + update & +0.0043 & 8,785.9 \\
Variance $<$ 0.95 + update & $-$0.0011 & 63,605.9 \\
Variance $<$ 0.98 + update & +0.0136 & 278,359.6 \\
\bottomrule
\end{tabular}
\caption{\textbf{A frame-zero session state needs no refresh.} Change in PSNR of deployed Pro against keeping the frame-zero state, sequences weighted equally, and the net bytes per frame that refreshing adds, including the change in packet size. (a) Refitting the PCA basis every 60 or 120 frames, or once the explained variance of the current frame falls below the given fraction of that of frame zero; a refresh detected at frame $t$ takes effect from $t{+}1$. (b) Also refitting the hypernetwork update on the latest 30 frames at each refresh. Each block averages the sequences that carry all of its rows.}\label{tab:long_refresh}
\end{table}

\section{Comparison details}
\label{sec:supp_dyn}
From the same eight input cameras, every tier has higher PSNR, lower LPIPS and fewer bytes than each per-frame optimisation method on all \CmpDynSparseAttempted\ sequences, and Pro and Ultra beat the offline 4C4D on every sequence in PSNR, SSIM, LPIPS and bytes. The paragraphs below give the settings behind each comparison of the main paper and extend it to every sequence.

\begin{figure*}[t]
\centering
\begin{tikzpicture}
\begin{groupplot}[paper,group style={group size=2 by 1,horizontal sep=1.5cm},width=.46\textwidth,height=7.5cm,xmode=log,xlabel={Packet size (kB, log scale)},ylabel={PSNR (dB)},legend style={font=\tiny,legend columns=4,/tikz/every even column/.append style={column sep=6pt}},legend cell align=left]
\nextgroupplot[title={without NSC},legend to name=reviewrdlegend]
\addplot[mark=*,mark size=1pt,color=cA] coordinates {(3630.746000000,26.183096743)};
\addlegendentry{Uncompressed float32}
\addplot[mark=*,mark size=1pt,color=cB] coordinates {(852.177812500,26.171617498)};
\addlegendentry{SPZ v4}
\addplot[mark=*,mark size=1pt,color=cC] coordinates {(1595.280937500,26.182661445) (1607.024562500,26.182661445) (1673.679125000,26.182661445)};
\addlegendentry{FP16 + Zstd}
\addplot[mark=*,mark size=1pt,color=black!60] coordinates {(708.584500000,21.403178316) (916.996000000,25.795272703) (1076.744000000,26.182169420)};
\addlegendentry{Uniform quantisation + Zstd}
\addplot[mark=*,mark size=1pt,color=violet] coordinates {(651.551687500,25.920102037) (723.883750000,26.180068048) (1101.312687500,26.181601343) (1551.190000000,26.182998650)};
\addlegendentry{Draco}
\addplot[mark=*,mark size=1pt,color=cyan!60!black] coordinates {(1056.284000000,26.203626758)};
\addlegendentry{PlayCanvas compressed PLY}
\addplot[mark=*,mark size=1pt,color=orange!85!black] coordinates {(840.895562500,26.258158642) (856.312062500,26.258158642) (857.898812500,26.258158642) (933.435437500,26.258158642)};
\addlegendentry{PlayCanvas SOG}
\addplot[mark=*,mark size=1pt,color=green!50!black] coordinates {(1564.202750000,26.326988336) (1564.202750000,26.326988336) (2857.849000000,26.327092222)};
\addlegendentry{ksplat}
\addplot[mark=*,mark size=1pt,color=red!70!black] coordinates {(696.095187500,26.010892276) (727.850562500,26.069554850) (781.536750000,25.895536758) (826.595750000,26.142343943) (872.244437500,26.099673825)};
\addlegendentry{FCGS}
\addplot[mark=star,mark size=2.4pt,line width=1.3pt,color=black] coordinates {(94.834062500,25.772292038) (208.174062500,26.065117449) (321.944375000,26.095082837)};
\addlegendentry{Ours (hypernetwork)}
\addplot[mark=*,mark size=1pt,color=cA] coordinates {(8.178750000,21.376234858) (23.180812500,22.930482049) (45.448937500,23.142571547)};
\addlegendentry{CodecSplat}
\addplot[densely dotted,mark=none,line width=.6pt,color=cB] coordinates {(65.815750000,24.052772830) (80.439625000,24.646475347) (94.835062500,24.945793614) (123.224812500,25.285947714) (151.759437500,25.472590340) (179.889000000,25.576931340) (207.909750000,25.634114183) (235.700250000,25.675675082) (263.157625000,25.700634920) (317.812875000,25.726757066) (372.177375000,25.762227486) (481.178000000,25.790721282)};
\addlegendentry{Ours, plain decoding, 4 bits}
\addplot[densely dotted,mark=none,line width=.6pt,color=cC] coordinates {(84.855687500,24.198234491) (109.363312500,24.833677446) (133.908812500,25.153013378) (182.907125000,25.526108420) (231.955750000,25.731428387) (280.954250000,25.844182093) (329.964000000,25.905772095) (378.943062500,25.952032223) (427.851375000,25.975551425) (525.604750000,26.008894593) (623.297687500,26.041343937) (818.797062500,26.071381668)};
\addlegendentry{Ours, plain decoding, 5 bits}
\addplot[densely dotted,mark=none,line width=.6pt,color=black!60] coordinates {(94.044812500,24.228408009) (122.725437500,24.877344560) (151.273500000,25.203142118) (208.175062500,25.577956685) (265.154187500,25.786756629) (321.944375000,25.899815635) (378.707500000,25.962653622) (435.381062500,26.011409625) (491.927437500,26.035279228) (604.914250000,26.069807778) (717.757125000,26.103446074) (943.571500000,26.133069078)};
\addlegendentry{Ours, plain decoding, 6 bits}
\addplot[densely dotted,mark=none,line width=.6pt,color=violet] coordinates {(110.415375000,24.236344663) (146.695625000,24.889449677) (182.382125000,25.214192595) (253.489937500,25.589842976) (324.960437500,25.800463494) (395.899062500,25.915982501) (466.672000000,25.978714825) (537.226562500,26.026767710) (607.439375000,26.050237042) (747.505500000,26.086533317) (887.211687500,26.119636600) (1167.027437500,26.148866925)};
\addlegendentry{Ours, plain decoding, 8 bits}
\nextgroupplot[title={with NSC}]
\addplot[mark=*,mark size=1pt,forget plot,color=cA] coordinates {(3630.746000000,27.423307060)};
\addplot[mark=*,mark size=1pt,forget plot,color=cB] coordinates {(852.696687500,27.356862444)};
\addplot[mark=*,mark size=1pt,forget plot,color=cC] coordinates {(1599.397875000,27.422507486) (1611.174937500,27.422507486) (1674.889312500,27.422507486)};
\addplot[mark=*,mark size=1pt,forget plot,color=black!60] coordinates {(708.694562500,21.457541591) (916.802375000,27.308335229) (1078.821062500,27.418762967)};
\addplot[mark=*,mark size=1pt,forget plot,color=violet] coordinates {(650.604250000,27.154409252) (722.991687500,27.415457639) (1100.365125000,27.422530151) (1550.132437500,27.423337432)};
\addplot[mark=*,mark size=1pt,forget plot,color=cyan!60!black] coordinates {(1056.284000000,27.364383016)};
\addplot[mark=*,mark size=1pt,forget plot,color=orange!85!black] coordinates {(841.153875000,27.225777892) (854.449625000,27.225777892) (856.201750000,27.225777892) (933.751125000,27.225777892)};
\addplot[mark=*,mark size=1pt,forget plot,color=green!50!black] coordinates {(1564.202750000,27.197483766) (1564.202750000,27.197483766) (2857.849000000,27.198371059)};
\addplot[mark=*,mark size=1pt,forget plot,color=red!70!black] coordinates {(724.723312500,27.042542481) (779.821187500,27.022503657) (823.514312500,27.083118455) (866.470812500,27.206839630)};
\addplot[mark=star,mark size=2.4pt,line width=1.3pt,forget plot,color=black] coordinates {(94.834062500,26.839858158) (208.174062500,27.218734105) (321.944375000,27.273150475)};
\addplot[mark=*,mark size=1pt,forget plot,color=cA] coordinates {(8.178750000,21.376234858) (23.180812500,22.930482049) (45.448937500,23.142571547)};
\end{groupplot}
\end{tikzpicture}
\par\smallskip\pgfplotslegendfromname{reviewrdlegend}
\caption{\textbf{At equal PSNR, our tiers need on average \ReviewNSCSavingMin--\ReviewNSCSavingMax\% fewer bytes than uniform quantisation, Draco and FCGS with NSC (\ReviewNoNSCSavingMin--\ReviewNoNSCSavingMax\% without).} DNA-Rendering, 16 test scenes, 48 evaluated cameras. Format curves sweep each encoder's settings on our uncompressed pruned Gaussians; dotted curves sweep our codec with plain decoding and no client pruning at 4, 5, 6 and 8 bits, as in \cref{fig:tier_curve}. On the right, the Gaussian formats and our tiers use the same NSC, while CodecSplat codes its own features as on the left. BD-rate savings of our tiers are \ReviewNSCUniformZstdSaving, \ReviewNSCDracoSaving\ and \ReviewNSCFCGSSaving\% against uniform quantisation, Draco and FCGS with NSC, and \ReviewNoNSCUniformZstdSaving, \ReviewNoNSCDracoSaving\ and \ReviewNoNSCFCGSSaving\% without, over the PSNR range each shares with our tiers. The other formats have fewer than three Pareto-optimal points, too few for a delta, and CodecSplat stays below our PSNR range.}\label{fig:review_rd}
\end{figure*}
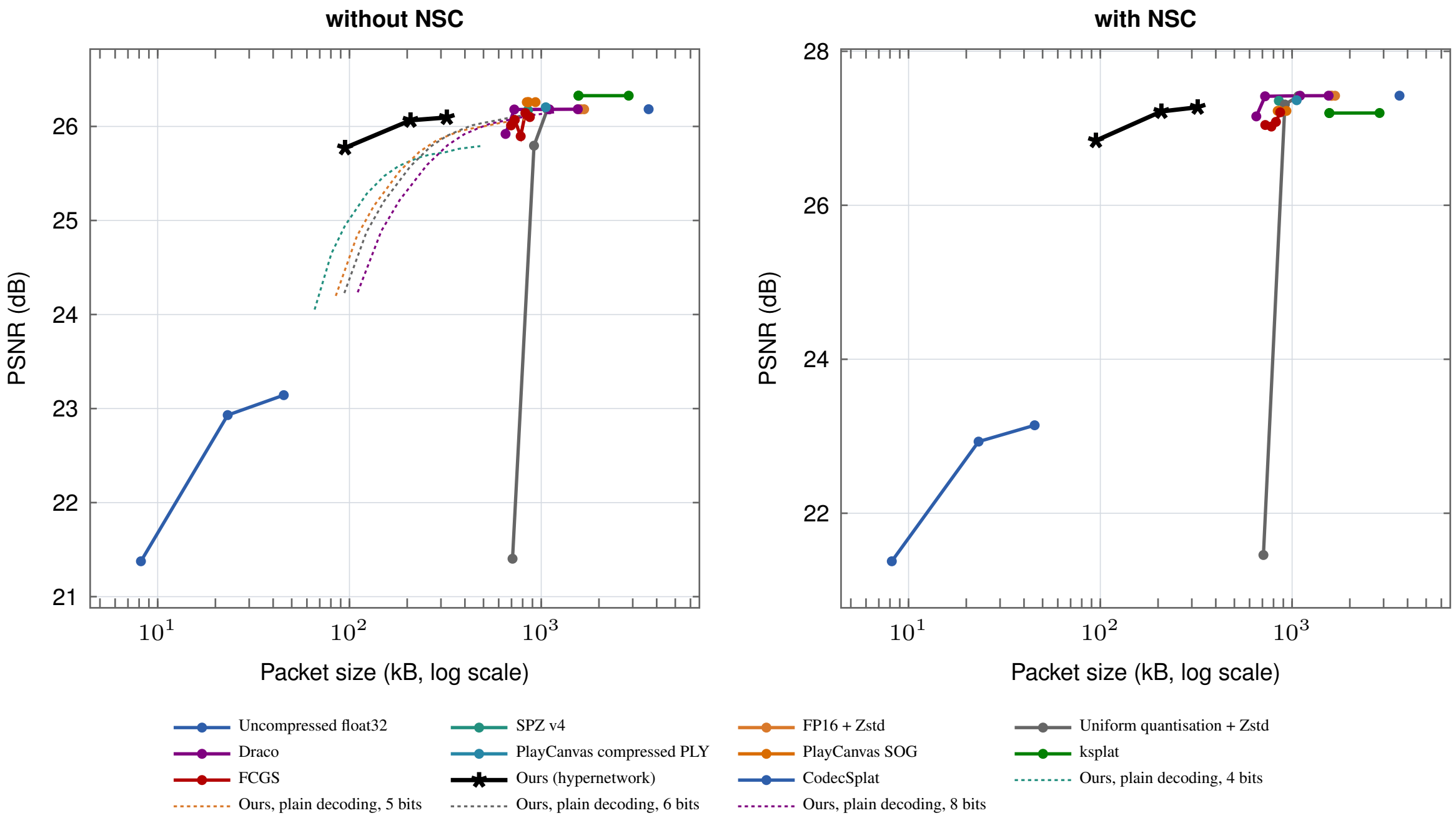

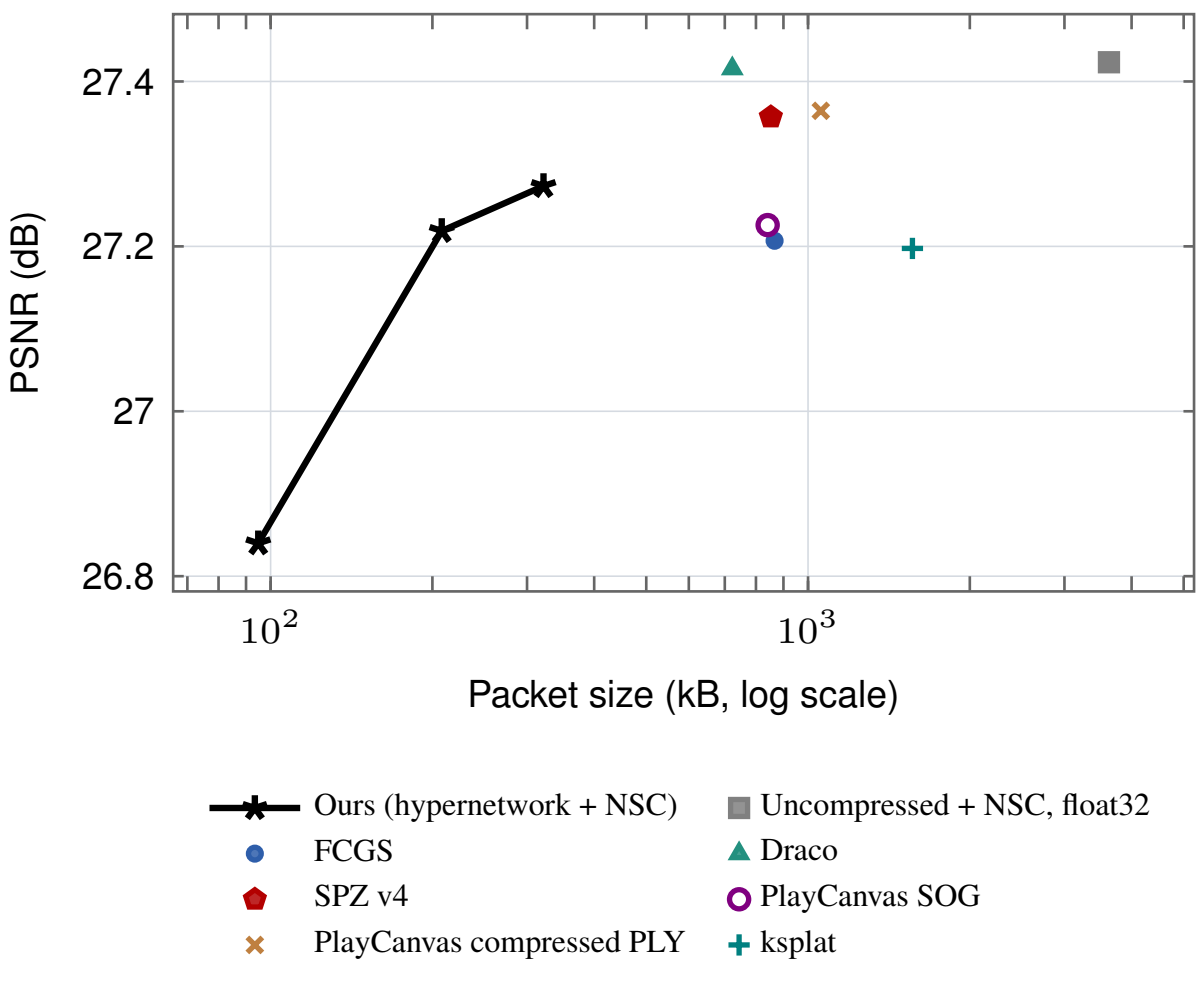
\begin{figure}[t]
\centering
\begin{tikzpicture}
\begin{axis}[paper,width=\columnwidth,height=5.4cm,xmode=log,xlabel={Packet size (kB, log scale)},ylabel={PSNR (dB)},legend columns=2,legend cell align=left,legend style={font=\fontsize{6}{7}\selectfont,at={(0.5,-0.32)},anchor=north,/tikz/every even column/.append style={column sep=6pt}}]
\addplot[black,very thick,mark=star,mark size=2.4pt] coordinates {(94.834,26.8399) (208.174,27.2187) (321.944,27.2732)};
\addlegendentry{Ours (hypernetwork + NSC)}
\addplot[only marks,mark=square*,mark size=1.6pt,gray] coordinates {(3630.746,27.4233)};
\addlegendentry{Uncompressed + NSC, float32}
\addplot[only marks,cA,mark=*,mark size=1.2pt] coordinates {(866.471,27.2068)};
\addlegendentry{FCGS}
\addplot[only marks,cB,mark=triangle*,mark size=1.5pt] coordinates {(722.992,27.4155)};
\addlegendentry{Draco}
\addplot[only marks,mark=pentagon*,mark size=1.8pt,red!70!black] coordinates {(852.697,27.3569)};
\addlegendentry{SPZ v4}
\addplot[only marks,mark=o,mark size=1.8pt,violet] coordinates {(841.154,27.2258)};
\addlegendentry{PlayCanvas SOG}
\addplot[only marks,mark=x,mark size=2pt,brown] coordinates {(1056.284,27.3644)};
\addlegendentry{PlayCanvas compressed PLY}
\addplot[only marks,mark=+,mark size=2pt,teal] coordinates {(1564.203,27.1975)};
\addlegendentry{ksplat}
\end{axis}
\end{tikzpicture}
\caption{\textbf{Ultra comes within \CmpSameCalMaxGainRounded\,dB of the best of six Gaussian compression formats with at least 2.24$\times$ fewer bytes.} Rate--PSNR points of \cref{tab:cmp_codecs} on a logarithmic byte axis.}
\label{fig:cmp_rate}
\end{figure}

\paragraph{Existing formats.}
Each format runs at the official default of its encoder or the setting recommended in its paper: SPZ v4~\cite{spz} with its default packing, the PlayCanvas compressed-PLY writer~\cite{pcply}, SOG~\cite{pcsog} with libwebp's default lossless settings, ksplat~\cite{ksplat} compression level 1, the Google Draco~\cite{draco} encoder default (position 11 bits, generic attributes 8 bits, compression level 7) and FCGS with its default $\lambda{=}10^{-4}$. As in \cref{tab:cmp_formats}, every Gaussian set receives the NSC that our tiers use, with scales clamped to the smallest normal float32 before encoding. \Cref{tab:cmp_codecs} adds each format's setting, LPIPS and coding times, \cref{fig:cmp_rate} plots it against bytes, and \cref{tab:cmp_thresholds} marks which format defaults reach each tier. \Cref{tab:cmp_formats_nonsc} repeats the comparison without NSC on either side, and \cref{fig:review_rd} sweeps the settings of every format with and without it.
\begin{table}[t]
\centering
\small
\setlength{\tabcolsep}{4pt}
\resizebox{\columnwidth}{!}{%
\begin{tabular}{lrrrr}
\toprule
Format & kB\,$\downarrow$ & PSNR\,$\uparrow$ & SSIM\,$\uparrow$ & Decode (ms)\,$\downarrow$ \\
\midrule
PointSplat512, float32 & 4753.1 & 26.169 & 0.8650 & -- \\
Uncompressed, float32 & 3630.7 & 26.183 & 0.8649 & -- \\
\midrule
SPZ v4 & 852.2 & 26.172 & 0.8638 & 27.9 \\
PlayCanvas compressed PLY & 1056.3 & 26.204 & 0.8642 & 209.4 \\
FCGS~\cite{fcgs} & 872.2 & 26.100 & 0.8626 & 4562.7 \\
ksplat & 1564.2 & 26.327 & 0.8648 & 154.6 \\
Draco & 723.9 & 26.180 & 0.8647 & 15.3 \\
PlayCanvas SOG & 840.9 & 26.258 & 0.8638 & 221.2 \\
\midrule
CodecSplat~\cite{codecsplat} & 45.4 & 23.143 & 0.7434 & 199.1 \\
\midrule
Ours Lite & 94.8 & 25.772 & 0.8491 & 21.7 \\
Ours Pro & 208.2 & 26.065 & 0.8598 & 22.1 \\
Ours Ultra & 321.9 & 26.095 & 0.8614 & 22.8 \\
\bottomrule
\end{tabular}}
\caption{\textbf{Without NSC on either side, Gaussian compression formats need at least \CmpMinCodecOverPro$\times$ Pro's bytes at comparable PSNR.} The comparison of \cref{tab:cmp_formats} with uncalibrated Gaussians and our tiers decoded by the hypernetwork alone; NSC leaves our bytes and client times unchanged.}
\label{tab:cmp_formats_nonsc}
\end{table}

\paragraph{Gaussians in video.}
Following the packing of V$^3$~\cite{vcubed} and 2D Gaussian-video telepresence streams~\cite{tele360}, where quantisation plus H.265 together cost Tele360 about 2\,dB (32.65 to 30.57\,dB in its Table~11), we quantise our uncompressed Gaussians into 17 grey planes and code them with x265~\cite{x265} at five constant rate factors (CRFs), in Morton order or on the token grid (\cref{tab:video_gaussians}). The best point reaches \VidBestPSNR\,dB at \VidBestMbps\,Mbit/s. Eight-bit scales cannot hold the dynamic range of \VidExcludedScenes\ scenes; on the other scenes the best point reaches \VidDiagnosticBestPSNR\,dB, still below every tier at several times its bytes. Decoding the planes on the CPU takes \VidCPUMinMs--\VidCPUMaxMs\,ms per frame.
\begin{table*}[t]
\centering
\small
\begin{tabular}{llrrrrrrr}
\toprule
Layout & Calibration & CRF & kB/frame & Mbit/s & PSNR & SSIM & LPIPS & CPU ms/frame \\
\midrule
L1 & no NSC & 18 & 531.9 & 127.66 & 14.17 & 0.4681 & 0.3959 & 131.02 \\
L1 & no NSC & 23 & 439.0 & 105.36 & 13.77 & 0.4521 & 0.4118 & 126.50 \\
L1 & no NSC & 28 & 405.8 & 97.40 & 12.21 & 0.4238 & 0.4199 & 121.12 \\
L1 & no NSC & 33 & 371.0 & 89.04 & 10.56 & 0.3906 & 0.4334 & 120.18 \\
L1 & no NSC & 38 & 337.7 & 81.05 & 9.32 & 0.3615 & 0.4524 & 112.79 \\
L1 & NSC & 18 & 530.9 & 127.42 & 15.51 & 0.4895 & 0.3808 & -- \\
L1 & NSC & 23 & 437.9 & 105.10 & 15.00 & 0.4724 & 0.3962 & -- \\
L1 & NSC & 28 & 404.8 & 97.14 & 13.25 & 0.4406 & 0.4065 & -- \\
L1 & NSC & 33 & 370.0 & 88.80 & 11.28 & 0.4042 & 0.4240 & -- \\
L1 & NSC & 38 & 337.0 & 80.88 & 9.86 & 0.3748 & 0.4441 & -- \\
L2 & no NSC & 18 & 775.8 & 186.18 & 15.15 & 0.4790 & 0.3892 & 143.86 \\
L2 & no NSC & 23 & 660.1 & 158.43 & 14.82 & 0.4642 & 0.4042 & 137.86 \\
L2 & no NSC & 28 & 619.1 & 148.59 & 13.10 & 0.4343 & 0.4128 & 134.24 \\
L2 & no NSC & 33 & 576.8 & 138.44 & 11.38 & 0.4011 & 0.4264 & 129.41 \\
L2 & no NSC & 38 & 537.8 & 129.08 & 10.71 & 0.3849 & 0.4356 & 121.59 \\
L2 & NSC & 18 & 773.7 & 185.69 & 16.36 & 0.5004 & 0.3739 & -- \\
L2 & NSC & 23 & 658.1 & 157.94 & 15.96 & 0.4832 & 0.3913 & -- \\
L2 & NSC & 28 & 617.1 & 148.11 & 14.25 & 0.4504 & 0.4025 & -- \\
L2 & NSC & 33 & 575.1 & 138.02 & 12.39 & 0.4153 & 0.4182 & -- \\
L2 & NSC & 38 & 536.5 & 128.75 & 11.59 & 0.3981 & 0.4278 & -- \\
\bottomrule
\end{tabular}
\caption{\textbf{Packing Gaussians into video costs quality at several times our bytes.} V$^3$-style packing of our uncompressed Gaussians over DNA-Rendering frames 30--89 of the 16 test scenes, scored at frames 40, 50, \dots, 80 on eight held-out cameras, using 16-bit positions with lossless high bytes and 8-bit other attributes in 17 grey planes coded with x265 at a group-of-pictures (GOP) length of 30 in Morton order (L1) or on the token grid (L2), without and with NSC. Bytes count every stream and all side information. Means cover all 16 scenes, including the \VidExcludedScenes\ whose scales exceed the 8-bit range (\cref{sec:supp_dyn}); CPU times measure decoding all planes with ffmpeg on three scenes, process start-up included.}\label{tab:video_gaussians}
\end{table*}

\begin{table*}[t]
\centering
\small
\setlength{\tabcolsep}{3pt}
\begin{tabular}{llrrrrrr}
\toprule
Format & Setting & kB & PSNR & SSIM & LPIPS & Encode (s) & Decode (s) \\
\midrule
Ours Lite & -- & 94.8 & 26.840 & 0.8700 & 0.1043 & -- & -- \\
Ours Pro & -- & 208.2 & 27.219 & 0.8802 & 0.0941 & -- & -- \\
Ours Ultra & -- & 321.9 & 27.273 & 0.8818 & 0.0923 & -- & -- \\
\midrule
Uncompressed + NSC, float32 & -- & 3630.7 & 27.423 & 0.8857 & 0.0884 & -- & -- \\
SPZ v4 & -- & 852.7 & 27.357 & 0.8844 & 0.0914 & 0.06 & 0.03 \\
PlayCanvas compressed PLY & -- & 1056.3 & 27.364 & 0.8846 & 0.0907 & 0.29 & 0.21 \\
FCGS~\cite{fcgs} & $\lambda{=}10^{-4}$ & 866.5 & 27.207 & 0.8777 & 0.0991 & 4.20 & 4.56 \\
ksplat & level 1 & 1564.2 & 27.197 & 0.8824 & 0.0925 & 0.28 & 0.15 \\
Draco & position 11 / attributes 8 bits & 723.0 & 27.415 & 0.8854 & 0.0889 & 0.07 & 0.02 \\
PlayCanvas SOG & default & 841.2 & 27.226 & 0.8827 & 0.0925 & 1.88 & 0.22 \\
\bottomrule
\end{tabular}
\caption{\textbf{With the same NSC, the best format gains \CmpSameCalMaxGainRounded\,dB over Ultra at \CmpSameCalMinBytesOverUltra$\times$ its bytes or more.} Each format encodes our final uncompressed Gaussians of the 16 DNA-Rendering test scenes with NSC at its official default or featured setting (\cref{sec:supp_dyn}); our tiers use the hypernetwork and NSC. Encode and Decode give seconds per frame recorded with the reference implementations on the Gaussians without NSC, on CPU except FCGS; the command-line PlayCanvas, ksplat and FCGS tools include process start-up and file I/O.}
\label{tab:cmp_codecs}
\end{table*}

\begin{table*}[t]
\centering
\small
\setlength{\tabcolsep}{6pt}
\begin{tabular}{lrrrrrr}
\toprule
& \multicolumn{3}{c}{Reach Pro quality (\ProKBCmp\,kB)} & \multicolumn{3}{c}{Reach Ultra quality (\UltraKBCmp\,kB)} \\
Format & PSNR & SSIM & LPIPS & PSNR & SSIM & LPIPS \\
\midrule
SPZ v4 & 852.7 & 852.7 & 852.7 & 852.7 & 852.7 & 852.7 \\
PlayCanvas compressed PLY & 1056.3 & 1056.3 & 1056.3 & 1056.3 & 1056.3 & 1056.3 \\
FCGS~\cite{fcgs} & -- & -- & -- & -- & -- & -- \\
ksplat & -- & 1564.2 & 1564.2 & -- & 1564.2 & -- \\
Draco & 723.0 & 723.0 & 723.0 & 723.0 & 723.0 & 723.0 \\
PlayCanvas SOG & 841.2 & 841.2 & 841.2 & -- & 841.2 & -- \\
\bottomrule
\end{tabular}
\caption{\textbf{With the same NSC, every format that reaches a tier sends at least \CmpSameCalMinBytesOverUltra$\times$ its bytes.} Each cell gives the kB/frame of a format's default if it reaches the tier in that metric alone, and `--' otherwise.}
\label{tab:cmp_thresholds}
\end{table*}

\paragraph{Per-sequence protocol.}
\Cref{tab:cmp_dyn} covers all 16 test sequences, frames 30--89. Its ranges span the sequences, kB/frame includes the first frame or our session state, and s/frame is the time per later frame on one RTX~5090. HiCoM, QUEEN-l and 3DGStream are each fitted from our eight input cameras and from 40 cameras, \CmpDynBaselineRuns\ runs in total, with their official settings: HiCoM's schedule of 15,000 steps for the first frame and 200 per later frame, the QUEEN-l configuration with MiDaS depth initialisation, and 3DGStream's SH degree 1 with 150 + 100 update steps per frame after a 30,000-step first-frame fit. Times are from runs with the same settings, sequences and inputs on frames 0--59. RGB is reduced to 512 and composited on black with the same soft masks as ours, and poses are converted to OpenCV with halved intrinsics. All three methods start from COLMAP points of frame 30 triangulated with the known calibration. Where this yields fewer than 1000 points, for all \CmpDynHullSparse\ eight-camera and \CmpDynHullDense\ of the 16 forty-camera settings, we draw 100,000 points inside the visual hull of the input masks (\cref{tab:cmp_dyn_availability}). Bytes are those of a decodable stream with the first frame amortised over 60 frames: QUEEN's compressed packets, 3DGStream's initial points with per-frame caches and added Gaussians, and a float32 serialisation of HiCoM's hierarchy and added Gaussians, since HiCoM's code has no bitstream writer. Our rows use the hypernetwork, which never saw the test scenes, with the hypernetwork update and NSC, both fitted on frames 0--29.

We fixed bugs in the baselines' public code without changing their optimisation. QUEEN's bundled rasteriser culls every point closer than 4\,m to the camera, which removes much of the body at DNA-Rendering distances; we restored the original 3DGS rule (0.2\,m) and retrained HiCoM, which uses the same rasteriser. We also fixed two binding errors and an empty-suffix slice (\texttt{[-0:]}) that selects everything instead of nothing after frames that add no points.

\paragraph{Results.}
From the same eight input cameras, every tier has higher PSNR, lower LPIPS and fewer bytes than each method on every one of the \CmpDynSparseAttempted\ sequences. Lite leads HiCoM by \CmpDynHiCoMSparseBelowLite, 3DGStream by \CmpDynGSStreamSparseBelowLite\ and QUEEN-l by \CmpDynQUEENSparseBelowLite\,dB. With all 40 cameras, QUEEN-l and HiCoM remain below Lite on every sequence, by \CmpDynQUEENDenseBelowLite\ and \CmpDynHiCoMDenseLiteGainMin--\CmpDynHiCoMDenseLiteGainMax\,dB, and 3DGStream, with five times our cameras, stays \CmpDynGSStreamDenseMeanBelowLite\,dB below Lite on average (\cref{tab:cmp_dyn}) while spending \CmpDynGSStreamDenseBytesOverUltra$\times$ Ultra's bytes. Across all \CmpDynBaselineRuns\ runs, every tier has lower LPIPS than the run on the same sequence, by \CmpDynPairedLPIPSGain\ on average. Every baseline stream is larger than each tier's, with a median of \CmpDynMedianOverLite$\times$ Lite's, \CmpDynMedianOverPro$\times$ Pro's and \CmpDynMedianOverUltra$\times$ Ultra's size. \Cref{tab:cmp_dyn_full} lists 0012\_09, 0022\_10 and 0166\_04, and \crefrange{tab:cmp_dyn_more_full}{tab:cmp_dyn_more_full_4} the other \CmpSparseMoreTotal\ sequences, where all \CmpSparseMorePairs\ eight-camera runs favour every tier in PSNR, LPIPS and bytes, with Lite ahead by \CmpSparseLiteGainMin\ to \CmpSparseLiteGainMax\,dB.

\paragraph{Offline sparse-view baseline.}
4C4D~\cite{fourcfourd}, like SparseCam4D~\cite{sparsecam4d}, fits one 4D Gaussian model to a whole captured sequence offline. We run it on all 60 frames from the same eight input cameras with its official DyNeRF~\cite{n3dv} configuration of 30,000 steps, starting from the points that its code reconstructs from the input views with MASt3R-based structure from motion~\cite{mast3r} through the MAtCha codebase~\cite{matcha}, and render and score its final checkpoint like the streaming runs. Its kB/frame counts the float32 parameters of its \CmpDynFourCGaussiansM\,M Gaussians over 60 frames, and its s/frame divides the whole fit, without initialisation, by 60. Pro and Ultra beat it on every sequence in PSNR, SSIM, LPIPS and bytes (\cref{tab:cmp_dyn_4c4d}), Ultra by \CmpDynFourCUltraGain\,dB on average. Lite beats it in LPIPS on all \CmpDynSparseAttempted\ sequences and by \CmpDynFourCLiteGain\,dB PSNR on average. \Cref{fig:qual_dna_all} shows every eight-view method on the view of \cref{fig:qual}.
\begin{table}[t]
\centering
\small
\setlength{\tabcolsep}{3pt}
\resizebox{\columnwidth}{!}{%
\begin{tabular}{lrrrrrrrr}
\toprule
& \multicolumn{4}{c}{4C4D (offline)} & \multicolumn{4}{c}{Ours Pro} \\
\cmidrule(lr){2-5}\cmidrule(lr){6-9}
Sequence & PSNR & SSIM & LPIPS & kB/frame & PSNR & SSIM & LPIPS & kB/frame \\
\midrule
0012\_09 & 19.21 & 0.712 & 0.222 & 15111 & 22.51 & 0.824 & 0.105 & 224 \\
0022\_10 & 27.06 & 0.879 & 0.167 & 12927 & 29.41 & 0.922 & 0.095 & 257 \\
0166\_04 & 22.18 & 0.760 & 0.177 & 25542 & 22.41 & 0.787 & 0.115 & 270 \\
0173\_02 & 20.47 & 0.728 & 0.211 & 13585 & 21.77 & 0.781 & 0.128 & 224 \\
0219\_07 & 23.81 & 0.823 & 0.174 & 7128 & 26.49 & 0.886 & 0.082 & 211 \\
0235\_11 & 22.03 & 0.793 & 0.187 & 11833 & 24.16 & 0.846 & 0.098 & 281 \\
0326\_07 & 23.04 & 0.765 & 0.252 & 5304 & 24.42 & 0.856 & 0.123 & 205 \\
0345\_08 & 25.42 & 0.874 & 0.144 & 10464 & 26.83 & 0.908 & 0.088 & 235 \\
0379\_09 & 27.24 & 0.822 & 0.200 & 3873 & 29.14 & 0.894 & 0.100 & 166 \\
0410\_05 & 25.60 & 0.804 & 0.228 & 16125 & 27.63 & 0.882 & 0.126 & 208 \\
0438\_02 & 26.50 & 0.829 & 0.175 & 15217 & 28.36 & 0.883 & 0.107 & 211 \\
0536\_06 & 28.47 & 0.872 & 0.164 & 10208 & 30.15 & 0.915 & 0.100 & 183 \\
0647\_02 & 28.49 & 0.906 & 0.146 & 11292 & 28.78 & 0.924 & 0.111 & 227 \\
0657\_05 & 27.61 & 0.857 & 0.191 & 12928 & 28.79 & 0.903 & 0.126 & 202 \\
0718\_06 & 20.80 & 0.749 & 0.175 & 39878 & 21.63 & 0.778 & 0.128 & 267 \\
0811\_06 & 26.96 & 0.900 & 0.124 & 20266 & 28.28 & 0.924 & 0.080 & 296 \\
\midrule
Mean & 24.68 & 0.817 & 0.184 & 14480 & 26.30 & 0.870 & 0.107 & 229 \\
\bottomrule
\end{tabular}}
\caption{\textbf{Pro beats offline 4C4D on every sequence in PSNR, SSIM, LPIPS and bytes.} Eight input views and the protocol of \cref{tab:cmp_dyn}.}
\label{tab:cmp_dyn_4c4d}
\end{table}

\input{fig/qual_dna_all}

\paragraph{Decoding.}
Our rows use the hypernetwork and NSC on every DNA sequence and tier. \Cref{tab:cmp_dyn_more_ours,tab:cmp_dyn_more_ours_b,tab:cmp_dyn_original_decoders} compare them with plain decoding of identical packets on the 48 evaluated cameras of \cref{tab:official_quality}; across all \CmpSparseTotal\ sequences, the hypernetwork and NSC raise Lite by \CmpSparseLiteDecoderGainMin\ to \CmpSparseLiteDecoderGainMax\,dB.
\begin{table*}[t]
\centering
\small
\setlength{\tabcolsep}{6pt}
\begin{tabular}{lrlrl}
\toprule
Sequence & \multicolumn{2}{c}{8 input cameras} & \multicolumn{2}{c}{40 input cameras} \\
& COLMAP points & Initialisation & COLMAP points & Initialisation \\
\midrule
0012\_09 & 399 & visual hull & 6044 & COLMAP \\
0022\_10 & 63 & visual hull & 1218 & COLMAP \\
0166\_04 & 613 & visual hull & 4159 & COLMAP \\
0173\_02 & 183 & visual hull & 5313 & COLMAP \\
0219\_07 & 91 & visual hull & 3488 & COLMAP \\
0235\_11 & 135 & visual hull & 2197 & COLMAP \\
0326\_07 & 93 & visual hull & 1360 & COLMAP \\
0345\_08 & 35 & visual hull & 1640 & COLMAP \\
0379\_09 & 0 & visual hull & 384 & visual hull \\
0410\_05 & 33 & visual hull & 699 & visual hull \\
0438\_02 & 43 & visual hull & 1072 & COLMAP \\
0536\_06 & 25 & visual hull & 470 & visual hull \\
0647\_02 & 30 & visual hull & 948 & visual hull \\
0657\_05 & 253 & visual hull & 1060 & COLMAP \\
0718\_06 & 175 & visual hull & 5148 & COLMAP \\
0811\_06 & 100 & visual hull & 1588 & COLMAP \\
\bottomrule
\end{tabular}
\caption{\textbf{Every per-sequence baseline run receives calibrated COLMAP points or a 100,000-point visual-hull fallback.} Initialisation per sequence at 8 and 40 input cameras; below 1000 COLMAP points, all three methods start from 100,000 random points in the visual hull of the input masks.}
\label{tab:cmp_dyn_availability}
\end{table*}

\begin{table*}[t]
\centering
\scriptsize
\setlength{\tabcolsep}{3pt}
\resizebox{\textwidth}{!}{%
\begin{tabular}{lrlrrrrrrrrrr}
\toprule
Method & Cameras & Sequence & \shortstack{Later-frame\\kB} & \shortstack{Amortised\\kB} & PSNR & SSIM & LPIPS & \shortstack{First\\frame (s)} & s/frame & \shortstack{Serialise\\(ms)} & \shortstack{Decode\\(ms)} & \shortstack{Render\\FPS} \\
\midrule
Ours Lite & 8 & 0173\_02 & 96.051 & 108.953 & 21.449 & 0.7622 & 0.1404 & -- & -- & -- & -- & -- \\
Ours Pro & 8 & 0173\_02 & 209.782 & 223.779 & 21.771 & 0.7810 & 0.1282 & -- & -- & -- & -- & -- \\
Ours Ultra & 8 & 0173\_02 & 324.539 & 340.722 & 21.835 & 0.7845 & 0.1250 & -- & -- & -- & -- & -- \\
Ours Lite & 8 & 0219\_07 & 89.008 & 101.911 & 25.954 & 0.8701 & 0.0956 & -- & -- & -- & -- & -- \\
Ours Pro & 8 & 0219\_07 & 196.755 & 210.752 & 26.494 & 0.8864 & 0.0818 & -- & -- & -- & -- & -- \\
Ours Ultra & 8 & 0219\_07 & 304.415 & 320.598 & 26.568 & 0.8888 & 0.0796 & -- & -- & -- & -- & -- \\
Ours Lite & 8 & 0235\_11 & 119.386 & 132.288 & 23.924 & 0.8367 & 0.1083 & -- & -- & -- & -- & -- \\
Ours Pro & 8 & 0235\_11 & 266.591 & 280.587 & 24.159 & 0.8459 & 0.0980 & -- & -- & -- & -- & -- \\
Ours Ultra & 8 & 0235\_11 & 412.975 & 429.158 & 24.362 & 0.8500 & 0.0939 & -- & -- & -- & -- & -- \\
Ours Lite & 8 & 0326\_07 & 87.120 & 100.022 & 23.959 & 0.8429 & 0.1394 & -- & -- & -- & -- & -- \\
Ours Pro & 8 & 0326\_07 & 190.984 & 204.980 & 24.419 & 0.8564 & 0.1227 & -- & -- & -- & -- & -- \\
Ours Ultra & 8 & 0326\_07 & 295.063 & 311.247 & 24.527 & 0.8595 & 0.1184 & -- & -- & -- & -- & -- \\
Ours Lite & 8 & 0345\_08 & 100.487 & 113.389 & 26.397 & 0.8961 & 0.1000 & -- & -- & -- & -- & -- \\
Ours Pro & 8 & 0345\_08 & 220.564 & 234.561 & 26.832 & 0.9080 & 0.0881 & -- & -- & -- & -- & -- \\
Ours Ultra & 8 & 0345\_08 & 340.767 & 356.950 & 27.021 & 0.9116 & 0.0839 & -- & -- & -- & -- & -- \\
Ours Lite & 8 & 0379\_09 & 69.210 & 82.113 & 28.803 & 0.8844 & 0.1110 & -- & -- & -- & -- & -- \\
Ours Pro & 8 & 0379\_09 & 152.256 & 166.253 & 29.137 & 0.8936 & 0.0997 & -- & -- & -- & -- & -- \\
Ours Ultra & 8 & 0379\_09 & 235.781 & 251.965 & 29.181 & 0.8959 & 0.0969 & -- & -- & -- & -- & -- \\
Ours Lite & 8 & 0410\_05 & 87.917 & 100.819 & 27.278 & 0.8720 & 0.1407 & -- & -- & -- & -- & -- \\
Ours Pro & 8 & 0410\_05 & 193.878 & 207.875 & 27.626 & 0.8819 & 0.1265 & -- & -- & -- & -- & -- \\
Ours Ultra & 8 & 0410\_05 & 299.393 & 315.577 & 27.679 & 0.8839 & 0.1238 & -- & -- & -- & -- & -- \\
Ours Lite & 8 & 0438\_02 & 89.485 & 102.387 & 27.992 & 0.8709 & 0.1183 & -- & -- & -- & -- & -- \\
Ours Pro & 8 & 0438\_02 & 196.924 & 210.920 & 28.358 & 0.8834 & 0.1067 & -- & -- & -- & -- & -- \\
Ours Ultra & 8 & 0438\_02 & 303.919 & 320.103 & 28.428 & 0.8866 & 0.1033 & -- & -- & -- & -- & -- \\
Ours Lite & 8 & 0536\_06 & 77.619 & 90.521 & 29.947 & 0.9101 & 0.1109 & -- & -- & -- & -- & -- \\
Ours Pro & 8 & 0536\_06 & 169.456 & 183.452 & 30.151 & 0.9151 & 0.0998 & -- & -- & -- & -- & -- \\
Ours Ultra & 8 & 0536\_06 & 261.757 & 277.941 & 30.227 & 0.9170 & 0.0962 & -- & -- & -- & -- & -- \\
Ours Lite & 8 & 0647\_02 & 97.229 & 110.132 & 28.076 & 0.9162 & 0.1226 & -- & -- & -- & -- & -- \\
Ours Pro & 8 & 0647\_02 & 213.266 & 227.262 & 28.782 & 0.9238 & 0.1110 & -- & -- & -- & -- & -- \\
Ours Ultra & 8 & 0647\_02 & 329.755 & 345.938 & 28.820 & 0.9249 & 0.1079 & -- & -- & -- & -- & -- \\
\bottomrule
\end{tabular}}
\caption{\textbf{Every tier leads eight-view per-frame optimisation in PSNR, LPIPS, and bytes across the other \CmpSparseMoreTotal\ test sequences.} \Cref{tab:cmp_dyn_full} lists 0012\_09, 0022\_10 and 0166\_04. Later-frame kB excludes the first frame of each method, or our session state, and amortised kB averages frame 0 and the 59 later frames. First frame (s) and s/frame give fitting time of the first frame and mean time per later frame on one RTX~5090, from the runs on frames 0--59 on shared GPUs; Serialise is the serialisation time per frame.}
\label{tab:cmp_dyn_more_full}
\end{table*}

\begin{table*}[t]
\centering
\scriptsize
\setlength{\tabcolsep}{3pt}
\resizebox{\textwidth}{!}{%
\begin{tabular}{lrlrrrrrrrrrr}
\toprule
Method & Cameras & Sequence & \shortstack{Later-frame\\kB} & \shortstack{Amortised\\kB} & PSNR & SSIM & LPIPS & \shortstack{First\\frame (s)} & s/frame & \shortstack{Serialise\\(ms)} & \shortstack{Decode\\(ms)} & \shortstack{Render\\FPS} \\
\midrule
Ours Lite & 8 & 0657\_05 & 85.851 & 98.753 & 28.385 & 0.8952 & 0.1382 & -- & -- & -- & -- & -- \\
Ours Pro & 8 & 0657\_05 & 187.788 & 201.785 & 28.792 & 0.9031 & 0.1255 & -- & -- & -- & -- & -- \\
Ours Ultra & 8 & 0657\_05 & 289.721 & 305.905 & 28.932 & 0.9056 & 0.1213 & -- & -- & -- & -- & -- \\
Ours Lite & 8 & 0718\_06 & 114.496 & 127.398 & 21.431 & 0.7648 & 0.1392 & -- & -- & -- & -- & -- \\
Ours Pro & 8 & 0718\_06 & 253.118 & 267.114 & 21.629 & 0.7778 & 0.1276 & -- & -- & -- & -- & -- \\
Ours Ultra & 8 & 0718\_06 & 392.232 & 408.416 & 21.678 & 0.7811 & 0.1239 & -- & -- & -- & -- & -- \\
Ours Lite & 8 & 0811\_06 & 125.744 & 138.646 & 27.963 & 0.9181 & 0.0897 & -- & -- & -- & -- & -- \\
Ours Pro & 8 & 0811\_06 & 282.199 & 296.195 & 28.280 & 0.9240 & 0.0802 & -- & -- & -- & -- & -- \\
Ours Ultra & 8 & 0811\_06 & 438.361 & 454.545 & 28.328 & 0.9255 & 0.0777 & -- & -- & -- & -- & -- \\
\midrule
HiCoM & 8 & 0173\_02 & 2687.218 & 2968.770 & 13.313 & 0.3011 & 0.5255 & 100.122 & 2.050 & 6.713 & 5.642 & 779.8 \\
HiCoM & 8 & 0219\_07 & 2867.800 & 3173.828 & 17.354 & 0.5853 & 0.3534 & 100.082 & 1.964 & 6.481 & 5.441 & 878.0 \\
HiCoM & 8 & 0235\_11 & 3352.303 & 3707.152 & 13.955 & 0.5019 & 0.4481 & 103.358 & 2.056 & 7.446 & 5.650 & 773.3 \\
HiCoM & 8 & 0326\_07 & 2873.498 & 3181.108 & 16.863 & 0.5197 & 0.4096 & 104.805 & 1.984 & 6.989 & 5.482 & 802.4 \\
HiCoM & 8 & 0345\_08 & 3603.752 & 3994.407 & 16.506 & 0.6446 & 0.3170 & 97.292 & 2.036 & 7.175 & 5.467 & 769.1 \\
HiCoM & 8 & 0379\_09 & 2922.214 & 3243.183 & 22.247 & 0.6658 & 0.3152 & 103.069 & 2.110 & 6.954 & 5.743 & 703.6 \\
HiCoM & 8 & 0410\_05 & 3029.779 & 3362.997 & 18.003 & 0.5044 & 0.4621 & 107.607 & 2.055 & 6.723 & 5.200 & 677.7 \\
HiCoM & 8 & 0438\_02 & 2854.410 & 3160.745 & 18.434 & 0.5632 & 0.4133 & 156.574 & 2.771 & 16.202 & 5.758 & 776.6 \\
HiCoM & 8 & 0536\_06 & 2883.696 & 3198.811 & 23.788 & 0.7069 & 0.3303 & 108.969 & 2.073 & 6.524 & 5.451 & 765.2 \\
HiCoM & 8 & 0647\_02 & 2977.052 & 3301.044 & 18.858 & 0.7455 & 0.2960 & 103.125 & 2.016 & 6.861 & 5.531 & 778.3 \\
HiCoM & 8 & 0657\_05 & 2963.212 & 3289.517 & 21.756 & 0.6828 & 0.3642 & 95.675 & 2.089 & 7.617 & 5.619 & 782.7 \\
HiCoM & 8 & 0718\_06 & 2953.084 & 3260.782 & 16.121 & 0.4902 & 0.4333 & 86.496 & 1.959 & 6.966 & 5.373 & 804.8 \\
HiCoM & 8 & 0811\_06 & 3545.691 & 3920.234 & 17.820 & 0.6657 & 0.3477 & 73.698 & 2.158 & 7.905 & 6.628 & 618.9 \\
QUEEN-l & 8 & 0173\_02 & 54031.257 & 55316.626 & 11.601 & 0.3320 & 0.5296 & 49.681 & 1.284 & 6859.330 & 6106.814 & 420.7 \\
QUEEN-l & 8 & 0219\_07 & 51697.802 & 52924.233 & 16.116 & 0.6412 & 0.3305 & 51.187 & 1.253 & 7193.362 & 6433.909 & 412.0 \\
QUEEN-l & 8 & 0235\_11 & 48879.730 & 50037.696 & 12.431 & 0.5185 & 0.4384 & 54.972 & 1.277 & 6850.869 & 6143.515 & 390.1 \\
QUEEN-l & 8 & 0326\_07 & 54497.476 & 55794.250 & 15.658 & 0.5287 & 0.4257 & 52.516 & 1.314 & 6890.727 & 6170.218 & 394.0 \\
QUEEN-l & 8 & 0345\_08 & 50866.091 & 52073.056 & 14.108 & 0.6399 & 0.3529 & 52.392 & 1.301 & 7384.708 & 6647.156 & 384.5 \\
QUEEN-l & 8 & 0379\_09 & 51038.252 & 52247.990 & 21.455 & 0.6586 & 0.3317 & 53.540 & 1.376 & 7526.054 & 6366.469 & 359.4 \\
QUEEN-l & 8 & 0410\_05 & 51038.960 & 52246.742 & 17.335 & 0.5476 & 0.4246 & 54.389 & 1.290 & 7614.331 & 6431.493 & 394.6 \\
\bottomrule
\end{tabular}}
\caption{\textbf{\Cref{tab:cmp_dyn_more_full}, continued.}}
\label{tab:cmp_dyn_more_full_2}
\end{table*}

\begin{table*}[t]
\centering
\scriptsize
\setlength{\tabcolsep}{3pt}
\resizebox{\textwidth}{!}{%
\begin{tabular}{lrlrrrrrrrrrr}
\toprule
Method & Cameras & Sequence & \shortstack{Later-frame\\kB} & \shortstack{Amortised\\kB} & PSNR & SSIM & LPIPS & \shortstack{First\\frame (s)} & s/frame & \shortstack{Serialise\\(ms)} & \shortstack{Decode\\(ms)} & \shortstack{Render\\FPS} \\
\midrule
QUEEN-l & 8 & 0438\_02 & 53817.035 & 55093.921 & 16.551 & 0.5103 & 0.4171 & 52.980 & 1.317 & 7736.324 & 6896.579 & 405.4 \\
QUEEN-l & 8 & 0536\_06 & 54902.643 & 56209.136 & 22.478 & 0.6794 & 0.3413 & 50.013 & 1.351 & 8127.081 & 6875.982 & 398.3 \\
QUEEN-l & 8 & 0647\_02 & 52490.131 & 53734.412 & 17.656 & 0.6727 & 0.3344 & 52.147 & 1.301 & 7610.390 & 6791.931 & 398.3 \\
QUEEN-l & 8 & 0657\_05 & 56340.206 & 57680.325 & 18.966 & 0.5077 & 0.3890 & 53.331 & 1.318 & 8008.575 & 7065.417 & 415.6 \\
QUEEN-l & 8 & 0718\_06 & 52408.231 & 53650.186 & 13.436 & 0.4668 & 0.4470 & 50.259 & 1.229 & 7381.049 & 6520.242 & 431.2 \\
QUEEN-l & 8 & 0811\_06 & 52970.933 & 54224.639 & 15.399 & 0.6440 & 0.3415 & 51.753 & 1.283 & 5959.987 & 5369.567 & 379.5 \\
3DGStream & 8 & 0173\_02 & 7995.881 & 8066.231 & 20.202 & 0.6901 & 0.2738 & 206.558 & 2.217 & 9.493 & 15.875 & 729.8 \\
3DGStream & 8 & 0219\_07 & 7983.656 & 8049.417 & 23.488 & 0.7952 & 0.2260 & 217.192 & 2.259 & 8.590 & 15.879 & 649.3 \\
3DGStream & 8 & 0235\_11 & 7990.582 & 8052.323 & 22.328 & 0.7701 & 0.2429 & 215.996 & 2.598 & 11.176 & 18.332 & 576.5 \\
3DGStream & 8 & 0326\_07 & 7980.226 & 8039.286 & 23.000 & 0.7546 & 0.2914 & 202.970 & 2.415 & 8.624 & 40.734 & 723.3 \\
3DGStream & 8 & 0345\_08 & 7982.142 & 8051.220 & 24.167 & 0.8391 & 0.2140 & 216.018 & 5.206 & 10.889 & 58.089 & 296.3 \\
3DGStream & 8 & 0379\_09 & 7973.656 & 8024.221 & 27.223 & 0.8122 & 0.2372 & 171.835 & 2.322 & 8.405 & 37.788 & 685.9 \\
3DGStream & 8 & 0410\_05 & 7989.588 & 8039.952 & 23.973 & 0.7652 & 0.3347 & 228.208 & 2.377 & 8.500 & 39.808 & 662.9 \\
3DGStream & 8 & 0438\_02 & 7985.086 & 8029.767 & 25.574 & 0.7757 & 0.2989 & 187.813 & 2.348 & 8.404 & 41.496 & 695.1 \\
3DGStream & 8 & 0536\_06 & 7980.374 & 8024.118 & 27.297 & 0.8527 & 0.2421 & 209.227 & 2.340 & 9.664 & 38.671 & 713.5 \\
3DGStream & 8 & 0647\_02 & 7985.491 & 8029.215 & 27.458 & 0.8904 & 0.2108 & 198.333 & 2.270 & 9.412 & 40.506 & 747.6 \\
3DGStream & 8 & 0657\_05 & 7982.456 & 8019.030 & 25.538 & 0.8265 & 0.2881 & 204.992 & 2.283 & 8.102 & 38.465 & 703.1 \\
3DGStream & 8 & 0718\_06 & 8042.251 & 8132.599 & 18.915 & 0.6230 & 0.3379 & 199.549 & 2.419 & 12.143 & 41.078 & 767.4 \\
3DGStream & 8 & 0811\_06 & 8005.638 & 8069.395 & 23.169 & 0.8209 & 0.2548 & 202.777 & 2.671 & 11.701 & 38.760 & 593.6 \\
HiCoM & 40 & 0173\_02 & 1248.996 & 1365.100 & 15.701 & 0.4357 & 0.4574 & 95.109 & 2.179 & 4.473 & 4.890 & 957.9 \\
HiCoM & 40 & 0219\_07 & 1168.829 & 1281.629 & 18.758 & 0.6164 & 0.3423 & 95.509 & 2.167 & 3.726 & 4.779 & 1234.9 \\
HiCoM & 40 & 0235\_11 & 1051.011 & 1150.329 & 16.954 & 0.6239 & 0.3806 & 98.834 & 2.178 & 4.383 & 5.003 & 1067.5 \\
HiCoM & 40 & 0326\_07 & 993.080 & 1089.033 & 20.145 & 0.6696 & 0.3322 & 103.674 & 2.334 & 4.684 & 5.399 & 1080.9 \\
HiCoM & 40 & 0345\_08 & 1255.431 & 1381.497 & 18.372 & 0.6970 & 0.2996 & 99.830 & 2.253 & 4.118 & 4.683 & 1014.8 \\
HiCoM & 40 & 0379\_09 & 2901.087 & 3219.697 & 23.584 & 0.7252 & 0.2856 & 83.136 & 2.332 & 6.397 & 5.179 & 767.6 \\
HiCoM & 40 & 0410\_05 & 2982.209 & 3308.884 & 21.242 & 0.6969 & 0.3755 & 97.311 & 2.336 & 6.921 & 5.233 & 745.5 \\
HiCoM & 40 & 0438\_02 & 644.502 & 705.035 & 21.658 & 0.6743 & 0.3581 & 98.307 & 2.283 & 3.300 & 4.795 & 1142.5 \\
HiCoM & 40 & 0536\_06 & 2943.791 & 3267.711 & 24.720 & 0.7811 & 0.2891 & 88.022 & 2.392 & 6.584 & 5.272 & 792.5 \\
HiCoM & 40 & 0647\_02 & 2954.072 & 3274.551 & 21.014 & 0.7904 & 0.2799 & 105.265 & 2.397 & 6.356 & 5.097 & 812.7 \\
HiCoM & 40 & 0657\_05 & 586.328 & 643.746 & 21.660 & 0.7397 & 0.3486 & 100.336 & 2.325 & 2.956 & 4.624 & 1111.1 \\
\bottomrule
\end{tabular}}
\caption{\textbf{\Cref{tab:cmp_dyn_more_full}, continued.}}
\label{tab:cmp_dyn_more_full_3}
\end{table*}

\begin{table*}[t]
\centering
\scriptsize
\setlength{\tabcolsep}{3pt}
\resizebox{\textwidth}{!}{%
\begin{tabular}{lrlrrrrrrrrrr}
\toprule
Method & Cameras & Sequence & \shortstack{Later-frame\\kB} & \shortstack{Amortised\\kB} & PSNR & SSIM & LPIPS & \shortstack{First\\frame (s)} & s/frame & \shortstack{Serialise\\(ms)} & \shortstack{Decode\\(ms)} & \shortstack{Render\\FPS} \\
\midrule
HiCoM & 40 & 0718\_06 & 1015.610 & 1112.433 & 16.911 & 0.5399 & 0.4100 & 131.049 & 2.409 & 4.355 & 5.546 & 1006.4 \\
HiCoM & 40 & 0811\_06 & 1264.599 & 1391.049 & 19.515 & 0.7510 & 0.3017 & 97.912 & 2.072 & 4.353 & 4.560 & 890.5 \\
QUEEN-l & 40 & 0173\_02 & 4011.063 & 4070.350 & 17.982 & 0.6097 & 0.3387 & 135.636 & 4.547 & 494.910 & 360.807 & 794.5 \\
QUEEN-l & 40 & 0219\_07 & 2258.409 & 2294.353 & 22.778 & 0.8218 & 0.2016 & 134.288 & 4.571 & 287.350 & 165.868 & 841.5 \\
QUEEN-l & 40 & 0235\_11 & 1889.023 & 1908.310 & 20.688 & 0.7797 & 0.2265 & 124.676 & 4.488 & 364.870 & 245.978 & 834.2 \\
QUEEN-l & 40 & 0326\_07 & 1685.625 & 1707.861 & 22.003 & 0.7933 & 0.2419 & 142.899 & 4.608 & 282.581 & 166.742 & 845.9 \\
QUEEN-l & 40 & 0345\_08 & 2267.771 & 2292.600 & 20.844 & 0.7645 & 0.2676 & 149.622 & 4.732 & 287.185 & 184.402 & 835.1 \\
QUEEN-l & 40 & 0379\_09 & 183690.716 & 188133.781 & 20.983 & 0.6502 & 0.3581 & 545.174 & 20.586 & 7302.344 & 6380.865 & 190.1 \\
QUEEN-l & 40 & 0410\_05 & 158954.986 & 162794.101 & 20.363 & 0.6241 & 0.4001 & 610.550 & 18.400 & 6715.998 & 6058.561 & 208.4 \\
QUEEN-l & 40 & 0438\_02 & 1941.090 & 1966.346 & 21.793 & 0.7127 & 0.3256 & 155.171 & 4.746 & 284.267 & 157.668 & 849.5 \\
QUEEN-l & 40 & 0536\_06 & 194463.182 & 199176.677 & 22.693 & 0.6628 & 0.3548 & 602.551 & 23.431 & 8194.096 & 7355.756 & 167.2 \\
QUEEN-l & 40 & 0647\_02 & 183560.819 & 188001.519 & 19.660 & 0.6911 & 0.3482 & 598.649 & 21.293 & 7459.476 & 6599.103 & 186.7 \\
QUEEN-l & 40 & 0657\_05 & 2587.927 & 2629.904 & 23.192 & 0.7475 & 0.3237 & 152.203 & 4.280 & 182.587 & 113.020 & 891.5 \\
QUEEN-l & 40 & 0718\_06 & 4079.114 & 4131.700 & 18.013 & 0.5693 & 0.3725 & 141.980 & 4.675 & 634.924 & 426.831 & 772.7 \\
QUEEN-l & 40 & 0811\_06 & 2686.526 & 2712.549 & 21.532 & 0.7741 & 0.2911 & 149.128 & 4.571 & 511.741 & 343.377 & 815.3 \\
3DGStream & 40 & 0173\_02 & 8022.227 & 8034.606 & 22.014 & 0.7884 & 0.1908 & 200.729 & 2.155 & 9.704 & 38.941 & 754.8 \\
3DGStream & 40 & 0219\_07 & 8000.885 & 7985.765 & 25.360 & 0.8602 & 0.1781 & 204.988 & 2.142 & 8.979 & 42.788 & 862.3 \\
3DGStream & 40 & 0235\_11 & 8003.732 & 7974.120 & 24.523 & 0.8512 & 0.1607 & 197.319 & 2.175 & 9.184 & 40.798 & 804.1 \\
3DGStream & 40 & 0326\_07 & 7988.946 & 7945.213 & 27.474 & 0.8898 & 0.1579 & 188.441 & 2.145 & 25.565 & 40.802 & 851.9 \\
3DGStream & 40 & 0345\_08 & 7988.476 & 7982.580 & 25.530 & 0.8827 & 0.1666 & 193.588 & 2.238 & 17.306 & 37.018 & 748.3 \\
3DGStream & 40 & 0379\_09 & 7974.280 & 8024.764 & 30.156 & 0.9136 & 0.1300 & 210.423 & 2.257 & 7.892 & 47.291 & 720.8 \\
3DGStream & 40 & 0410\_05 & 7993.785 & 8042.653 & 25.941 & 0.8244 & 0.2914 & 204.280 & 2.298 & 8.162 & 40.940 & 710.3 \\
3DGStream & 40 & 0438\_02 & 7993.725 & 7932.900 & 26.573 & 0.8153 & 0.2663 & 188.871 & 2.128 & 8.274 & 40.139 & 912.8 \\
3DGStream & 40 & 0536\_06 & 7983.984 & 8028.786 & 29.645 & 0.8986 & 0.1907 & 208.672 & 2.404 & 10.053 & 40.567 & 615.6 \\
3DGStream & 40 & 0647\_02 & 7988.349 & 8056.197 & 29.018 & 0.9134 & 0.1930 & 216.001 & 2.593 & 12.343 & 41.981 & 678.3 \\
3DGStream & 40 & 0657\_05 & 7987.085 & 7915.963 & 26.528 & 0.8579 & 0.2669 & 180.661 & 2.171 & 8.774 & 62.026 & 853.8 \\
3DGStream & 40 & 0718\_06 & 8063.203 & 8095.365 & 19.741 & 0.6650 & 0.3122 & 203.321 & 2.408 & 13.518 & 44.273 & 734.6 \\
3DGStream & 40 & 0811\_06 & 8013.570 & 8002.455 & 24.208 & 0.8514 & 0.2300 & 202.440 & 2.378 & 8.502 & 38.454 & 637.0 \\
\bottomrule
\end{tabular}}
\caption{\textbf{\Cref{tab:cmp_dyn_more_full}, continued.}}
\label{tab:cmp_dyn_more_full_4}
\end{table*}

\begin{table*}[t]
\centering
\scriptsize
\setlength{\tabcolsep}{3pt}
\resizebox{\textwidth}{!}{%
\begin{tabular}{llrrrrrrrr}
\toprule
 & & & & \multicolumn{3}{c}{Plain} & \multicolumn{3}{c}{Hypernetwork\,+\,NSC} \\ \cmidrule(lr){5-7}\cmidrule(lr){8-10} Sequence & Tier & Packet kB & Amortised kB & PSNR & SSIM & LPIPS & PSNR & SSIM & LPIPS \\
\midrule
0173\_02 & Lite & 94.766 & 107.668 & 21.436 & 0.7215 & 0.2179 & 22.138 & 0.7770 & 0.1362 \\
0173\_02 & Pro & 206.238 & 220.235 & 22.018 & 0.7557 & 0.1652 & 22.505 & 0.7951 & 0.1229 \\
0173\_02 & Ultra & 319.001 & 335.185 & 22.246 & 0.7717 & 0.1397 & 22.581 & 0.7980 & 0.1209 \\
0219\_07 & Lite & 90.404 & 103.306 & 24.793 & 0.8250 & 0.1452 & 26.171 & 0.8677 & 0.0904 \\
0219\_07 & Pro & 198.138 & 212.135 & 25.507 & 0.8535 & 0.1122 & 26.688 & 0.8842 & 0.0780 \\
0219\_07 & Ultra & 306.373 & 322.557 & 25.776 & 0.8632 & 0.0981 & 26.753 & 0.8859 & 0.0764 \\
0235\_11 & Lite & 114.291 & 127.193 & 23.692 & 0.7918 & 0.1624 & 25.326 & 0.8523 & 0.0953 \\
0235\_11 & Pro & 252.315 & 266.312 & 24.312 & 0.8213 & 0.1268 & 25.592 & 0.8635 & 0.0855 \\
0235\_11 & Ultra & 390.828 & 407.012 & 24.482 & 0.8324 & 0.1080 & 25.627 & 0.8659 & 0.0837 \\
0326\_07 & Lite & 85.939 & 98.841 & 23.107 & 0.8183 & 0.2107 & 24.843 & 0.8721 & 0.1195 \\
0326\_07 & Pro & 189.172 & 203.169 & 23.715 & 0.8392 & 0.1712 & 25.225 & 0.8833 & 0.1046 \\
0326\_07 & Ultra & 292.412 & 308.596 & 24.161 & 0.8532 & 0.1442 & 25.319 & 0.8861 & 0.1011 \\
0345\_08 & Lite & 95.443 & 108.345 & 24.605 & 0.8533 & 0.1745 & 27.121 & 0.9071 & 0.0843 \\
0345\_08 & Pro & 209.692 & 223.689 & 25.766 & 0.8833 & 0.1252 & 27.780 & 0.9195 & 0.0730 \\
0345\_08 & Ultra & 324.073 & 340.257 & 26.493 & 0.9010 & 0.0987 & 27.871 & 0.9218 & 0.0710 \\
0379\_09 & Lite & 68.451 & 81.353 & 28.157 & 0.8593 & 0.1919 & 30.053 & 0.8981 & 0.1032 \\
0379\_09 & Pro & 151.100 & 165.097 & 28.679 & 0.8802 & 0.1474 & 30.393 & 0.9067 & 0.0921 \\
0379\_09 & Ultra & 234.121 & 250.305 & 28.849 & 0.8887 & 0.1254 & 30.405 & 0.9081 & 0.0899 \\
0410\_05 & Lite & 88.073 & 100.975 & 27.036 & 0.8472 & 0.2218 & 29.444 & 0.8949 & 0.1211 \\
0410\_05 & Pro & 193.390 & 207.387 & 27.655 & 0.8659 & 0.1831 & 29.867 & 0.9044 & 0.1104 \\
0410\_05 & Ultra & 298.339 & 314.523 & 28.102 & 0.8776 & 0.1582 & 29.902 & 0.9055 & 0.1082 \\
\bottomrule
\end{tabular}}
\caption{\textbf{The hypernetwork and NSC improve identical packets on every sequence and tier (the other 13 sequences, part one).} 48 evaluated cameras, as in \cref{tab:official_quality}; packet kB is the frame-30 payload, and amortised kB adds one sixtieth of the session state.}
\label{tab:cmp_dyn_more_ours}
\end{table*}

\begin{table*}[t]
\centering
\scriptsize
\setlength{\tabcolsep}{3pt}
\resizebox{\textwidth}{!}{%
\begin{tabular}{llrrrrrrrr}
\toprule
 & & & & \multicolumn{3}{c}{Plain} & \multicolumn{3}{c}{Hypernetwork\,+\,NSC} \\ \cmidrule(lr){5-7}\cmidrule(lr){8-10} Sequence & Tier & Packet kB & Amortised kB & PSNR & SSIM & LPIPS & PSNR & SSIM & LPIPS \\
\midrule
0438\_02 & Lite & 85.680 & 98.582 & 27.183 & 0.8342 & 0.1815 & 29.302 & 0.8905 & 0.1004 \\
0438\_02 & Pro & 187.099 & 201.096 & 27.788 & 0.8572 & 0.1491 & 29.834 & 0.9022 & 0.0906 \\
0438\_02 & Ultra & 288.853 & 305.037 & 28.159 & 0.8691 & 0.1269 & 29.892 & 0.9037 & 0.0886 \\
0536\_06 & Lite & 74.995 & 87.897 & 28.715 & 0.8803 & 0.1877 & 31.365 & 0.9212 & 0.0999 \\
0536\_06 & Pro & 162.365 & 176.362 & 29.044 & 0.8895 & 0.1618 & 31.660 & 0.9268 & 0.0898 \\
0536\_06 & Ultra & 250.635 & 266.819 & 29.141 & 0.8961 & 0.1439 & 31.704 & 0.9276 & 0.0881 \\
0647\_02 & Lite & 85.814 & 98.716 & 26.024 & 0.8768 & 0.1710 & 29.297 & 0.9324 & 0.0882 \\
0647\_02 & Pro & 187.779 & 201.776 & 27.105 & 0.8931 & 0.1408 & 30.010 & 0.9396 & 0.0795 \\
0647\_02 & Ultra & 289.971 & 306.155 & 27.736 & 0.9020 & 0.1198 & 30.099 & 0.9407 & 0.0777 \\
0657\_05 & Lite & 77.179 & 90.081 & 27.194 & 0.8498 & 0.2044 & 30.443 & 0.9153 & 0.1040 \\
0657\_05 & Pro & 167.997 & 181.994 & 27.827 & 0.8644 & 0.1786 & 30.862 & 0.9221 & 0.0957 \\
0657\_05 & Ultra & 259.313 & 275.497 & 28.351 & 0.8746 & 0.1552 & 30.921 & 0.9229 & 0.0946 \\
0718\_06 & Lite & 113.828 & 126.730 & 21.028 & 0.7175 & 0.1784 & 21.725 & 0.7712 & 0.1267 \\
0718\_06 & Pro & 249.904 & 263.901 & 21.386 & 0.7419 & 0.1472 & 21.925 & 0.7851 & 0.1166 \\
0718\_06 & Ultra & 386.825 & 403.009 & 21.546 & 0.7549 & 0.1303 & 21.969 & 0.7873 & 0.1148 \\
0811\_06 & Lite & 121.913 & 134.815 & 27.084 & 0.8906 & 0.1351 & 28.629 & 0.9200 & 0.0800 \\
0811\_06 & Pro & 269.215 & 283.212 & 27.787 & 0.9054 & 0.1114 & 28.879 & 0.9247 & 0.0723 \\
0811\_06 & Ultra & 416.944 & 433.128 & 28.013 & 0.9110 & 0.0976 & 28.915 & 0.9258 & 0.0709 \\
\bottomrule
\end{tabular}}
\caption{\textbf{The hypernetwork and NSC improve identical packets on every sequence and tier (the other 13 sequences, part two).} 48 evaluated cameras, as in \cref{tab:official_quality}; packet kB is the frame-30 payload, and amortised kB adds one sixtieth of the session state.}
\label{tab:cmp_dyn_more_ours_b}
\end{table*}

\begin{table*}[t]
\centering
\scriptsize
\setlength{\tabcolsep}{3pt}
\resizebox{\textwidth}{!}{%
\begin{tabular}{llrrrrrrrr}
\toprule
 & & & & \multicolumn{3}{c}{Plain} & \multicolumn{3}{c}{Hypernetwork\,+\,NSC} \\ \cmidrule(lr){5-7}\cmidrule(lr){8-10} Sequence & Tier & Packet kB & Amortised kB & PSNR & SSIM & LPIPS & PSNR & SSIM & LPIPS \\
\midrule
0012\_09 & Lite & 95.742 & 108.644 & 20.338 & 0.7602 & 0.1607 & 21.678 & 0.8081 & 0.1113 \\
0012\_09 & Pro & 212.258 & 226.255 & 21.023 & 0.7863 & 0.1270 & 21.810 & 0.8159 & 0.1029 \\
0012\_09 & Ultra & 329.051 & 345.235 & 21.239 & 0.7966 & 0.1130 & 21.830 & 0.8167 & 0.1022 \\
0022\_10 & Lite & 108.916 & 121.818 & 27.066 & 0.8694 & 0.1957 & 29.053 & 0.9107 & 0.0997 \\
0022\_10 & Pro & 238.999 & 252.996 & 27.425 & 0.8842 & 0.1620 & 29.392 & 0.9176 & 0.0910 \\
0022\_10 & Ultra & 368.989 & 385.173 & 27.698 & 0.8916 & 0.1409 & 29.439 & 0.9188 & 0.0896 \\
0166\_04 & Lite & 115.911 & 128.813 & 21.738 & 0.7220 & 0.1548 & 22.850 & 0.7813 & 0.1090 \\
0166\_04 & Pro & 255.124 & 269.121 & 22.280 & 0.7495 & 0.1246 & 23.078 & 0.7923 & 0.1005 \\
0166\_04 & Ultra & 395.382 & 411.566 & 22.467 & 0.7618 & 0.1094 & 23.146 & 0.7948 & 0.0989 \\
\bottomrule
\end{tabular}}
\caption{\textbf{The hypernetwork and NSC improve identical packets on every sequence and tier (0012\_09, 0022\_10 and 0166\_04).} 48 evaluated cameras, as in \cref{tab:official_quality}; packet kB is the frame-30 payload, and amortised kB adds one sixtieth of the session state.}
\label{tab:cmp_dyn_original_decoders}
\end{table*}

\begin{table*}[t]
\centering
\small
\setlength{\tabcolsep}{3pt}
\begin{tabular}{lrlrrrrrrrrr}
\toprule
Method & Cameras & Sequence & \shortstack{Later-frame\\kB} & \shortstack{Amortised\\kB} & PSNR & SSIM & LPIPS & \shortstack{First\\frame (s)} & s/frame & \shortstack{Decode\\(ms)} & \shortstack{Render\\FPS} \\
\midrule
Ours Lite & 8 & 0012\_09 & 94.7 & 107.6 & 22.367 & 0.8144 & 0.1125 & -- & 0.13 & -- & -- \\
Ours Lite & 8 & 0022\_10 & 110.5 & 123.4 & 29.168 & 0.9154 & 0.1024 & -- & 0.17 & -- & -- \\
Ours Lite & 8 & 0166\_04 & 116.6 & 129.5 & 22.142 & 0.7752 & 0.1245 & -- & 0.16 & -- & -- \\
Ours Pro & 8 & 0012\_09 & 209.9 & 223.9 & 22.512 & 0.8236 & 0.1045 & -- & 0.13 & -- & -- \\
Ours Pro & 8 & 0022\_10 & 242.8 & 256.8 & 29.415 & 0.9218 & 0.0946 & -- & 0.18 & -- & -- \\
Ours Pro & 8 & 0166\_04 & 256.2 & 270.2 & 22.408 & 0.7870 & 0.1148 & -- & 0.17 & -- & -- \\
Ours Ultra & 8 & 0012\_09 & 325.1 & 341.2 & 22.541 & 0.8249 & 0.1036 & -- & 0.13 & -- & -- \\
Ours Ultra & 8 & 0022\_10 & 374.9 & 391.1 & 29.471 & 0.9228 & 0.0921 & -- & 0.18 & -- & -- \\
Ours Ultra & 8 & 0166\_04 & 396.8 & 413.0 & 22.457 & 0.7899 & 0.1119 & -- & 0.17 & -- & -- \\
\midrule
HiCoM~\cite{hicom} & 8 & 0012\_09 & 3422.3 & 3792.2 & 14.013 & 0.5292 & 0.3217 & 112 & 2.16 & 6.5 & 732 \\
HiCoM~\cite{hicom} & 8 & 0022\_10 & 3308.5 & 3667.7 & 19.206 & 0.6361 & 0.3759 & 88 & 2.11 & 5.5 & 695 \\
HiCoM~\cite{hicom} & 8 & 0166\_04 & 2877.9 & 3177.2 & 14.589 & 0.3614 & 0.5144 & 72 & 2.06 & 5.6 & 823 \\
QUEEN-l~\cite{queen} & 8 & 0012\_09 & 52292.3 & 53534.2 & 11.054 & 0.5339 & 0.3891 & 52 & 1.31 & 4800.9 & 369 \\
QUEEN-l~\cite{queen} & 8 & 0022\_10 & 53149.8 & 54410.8 & 16.543 & 0.5408 & 0.4125 & 52 & 1.30 & 5905.5 & 367 \\
QUEEN-l~\cite{queen} & 8 & 0166\_04 & 52971.9 & 54234.1 & 13.762 & 0.4478 & 0.4792 & 51 & 1.25 & 5915.4 & 447 \\
3DGStream~\cite{3dgstream} & 8 & 0012\_09 & 8001.3 & 8095.9 & 18.848 & 0.6870 & 0.2815 & 162 & 2.50 & 42.6 & 642 \\
3DGStream~\cite{3dgstream} & 8 & 0022\_10 & 7985.1 & 8042.6 & 26.640 & 0.8554 & 0.2365 & 191 & 2.57 & 38.5 & 654 \\
3DGStream~\cite{3dgstream} & 8 & 0166\_04 & 8015.1 & 8106.2 & 19.300 & 0.6318 & 0.3228 & 211 & 2.47 & 37.7 & 589 \\
HiCoM~\cite{hicom} & 40 & 0012\_09 & 1644.2 & 1804.5 & 16.100 & 0.5770 & 0.3052 & 101 & 2.20 & 5.0 & 871 \\
HiCoM~\cite{hicom} & 40 & 0022\_10 & 894.2 & 982.2 & 20.754 & 0.7143 & 0.3456 & 103 & 2.21 & 4.9 & 975 \\
HiCoM~\cite{hicom} & 40 & 0166\_04 & 1158.5 & 1271.0 & 16.579 & 0.5070 & 0.4274 & 92 & 2.16 & 5.0 & 925 \\
QUEEN-l~\cite{queen} & 40 & 0012\_09 & 6373.7 & 6493.2 & 16.836 & 0.6844 & 0.2957 & 148 & 4.56 & 482.1 & 861 \\
QUEEN-l~\cite{queen} & 40 & 0022\_10 & 1685.9 & 1707.3 & 23.736 & 0.7951 & 0.2816 & 148 & 4.28 & 172.9 & 901 \\
QUEEN-l~\cite{queen} & 40 & 0166\_04 & 5287.4 & 5361.5 & 17.651 & 0.5555 & 0.3801 & 142 & 4.68 & 401.6 & 734 \\
3DGStream~\cite{3dgstream} & 40 & 0012\_09 & 8042.3 & 8084.8 & 23.214 & 0.8238 & 0.1883 & 199 & 2.36 & 38.5 & 691 \\
3DGStream~\cite{3dgstream} & 40 & 0022\_10 & 7992.8 & 7944.3 & 27.838 & 0.8844 & 0.2068 & 187 & 2.25 & 40.3 & 757 \\
3DGStream~\cite{3dgstream} & 40 & 0166\_04 & 8042.5 & 8104.7 & 21.556 & 0.7460 & 0.2373 & 181 & 2.51 & 39.2 & 702 \\
\bottomrule
\end{tabular}
\caption{\textbf{Every tier outperforms eight-view per-frame optimisation in quality, bytes, and speed on each of 0012\_09, 0022\_10 and 0166\_04.} Later-frame kB excludes the first frame of each method, or our session state, and amortised kB averages frame 0 and the 59 later frames. Times are on one RTX~5090, for the baselines from their runs on frames 0--59 on shared GPUs; our s/frame is the steady-state server time per frame with one worker. The other sequences are in \cref{tab:cmp_dyn_more_full}.}
\label{tab:cmp_dyn_full}
\end{table*}

\paragraph{Speed.}
\Cref{tab:cmp_speed_idle} times all three methods at their official settings on 0012\_09 with both camera counts on an idle RTX~5090. It excludes first-frame fitting, serialisation and evaluation, and grants them ideal linear scaling $G/t_{\min}$ although each of their frames depends on the previous one. With the 40 views they are designed for, \CmpIdleClosestMethod\ comes closest, and our average throughput exceeds it \CmpIdleRatio$\times$ at equal GPU count. \Cref{tab:cmp_speed} applies the same scaling to all \CmpDynBaselineRuns\ grid runs, recorded on shared GPUs, where the fastest reach \CmpSharedIdealOneMax\ and \CmpSharedIdealFourMax\,FPS on one and four GPUs against our average of \SingleFPSMean\ and \FourFPSMean\,FPS. \Cref{tab:cmp_format_speed} models feed-forward reconstruction followed by each existing format; such pipelines still send the per-frame bytes of \cref{tab:cmp_codecs}.
\begin{table*}[t]
\centering
\small
\setlength{\tabcolsep}{4pt}
\begin{tabular}{lrrrrrrrr}
\toprule
& & \multicolumn{3}{c}{Optimisation on an idle GPU (s/frame)} & \multicolumn{2}{c}{Ideal FPS $G/t_{\min}$} & \multicolumn{2}{c}{Ours / $(G/t_{\min})$} \\
Method & Cameras & Trial 1 & Trial 2 & Trial 3 & $G{=}1$ & $G{=}4$ & $G{=}1$ & $G{=}4$ \\
\midrule
HiCoM & 8 & 2.004 & 2.003 & 2.065 & 0.499 & 1.997 & 16.05 & 16.13 \\
HiCoM & 40 & 2.200 & 2.143 & 2.267 & 0.467 & 1.867 & 17.17 & 17.26 \\
QUEEN-l & 8 & 0.918 & 0.872 & 0.909 & 1.147 & 4.589 & 6.98 & 7.02 \\
QUEEN-l & 40 & 4.499 & 4.649 & 4.696 & 0.222 & 0.889 & 36.05 & 36.24 \\
3DGStream & 8 & 1.967 & 1.932 & 1.981 & 0.517 & 2.070 & 15.48 & 15.56 \\
3DGStream & 40 & 2.117 & 2.128 & 2.156 & 0.472 & 1.889 & 16.96 & 17.05 \\
\midrule
Ours & 8 & -- & -- & -- & 8.01 & 32.22 & -- & -- \\
\bottomrule
\end{tabular}
\caption{\textbf{With their designed 40-camera input and ideal scaling, the fitted baselines stay \CmpIdleRatio$\times$ below our streaming throughput at equal GPU count.} Sequence 0012\_09 on an idle RTX~5090; $t_{\min}$ is the fastest mean fitting time per later frame, and $G/t_{\min}$ grants ideal scaling across $G$ GPUs. Ours is the whole-server throughput over the 16 test sequences.}
\label{tab:cmp_speed_idle}
\end{table*}

\begin{table*}[t]
\centering
\small
\setlength{\tabcolsep}{5pt}
\begin{tabular}{llrrrrr}
\toprule
Method & Cameras & $t$ range (s/frame) & \multicolumn{2}{c}{Max.\ $G/t$ (FPS)} & \multicolumn{2}{c}{Ours / $(G/t)$} \\
 & & & $G{=}1$ & $G{=}4$ & $G{=}1$ & $G{=}4$ \\
\midrule
Ours & 8 & -- & 8.01 & 32.22 & -- & -- \\
\midrule
HiCoM & 8 & 1.96--2.77 & 0.51 & 2.04 & 15.7 & 15.8 \\
QUEEN-l & 8 & 1.23--1.38 & 0.81 & 3.25 & 9.9 & 9.9 \\
3DGStream & 8 & 2.22--5.21 & 0.45 & 1.80 & 17.8 & 17.9 \\
HiCoM & 40 & 2.07--2.41 & 0.48 & 1.93 & 16.6 & 16.7 \\
QUEEN-l & 40 & 4.28--23.43 & 0.23 & 0.93 & 34.3 & 34.5 \\
3DGStream & 40 & 2.13--2.59 & 0.47 & 1.88 & 17.1 & 17.1 \\
\bottomrule
\end{tabular}
\caption{\textbf{Across all \CmpDynBaselineRuns\ grid runs, our server throughput exceeds every baseline's ideal-scaling rate.} Mean fitting time $t$ per later frame on shared GPUs and ideal throughput $G/t$ per method and input setting over the 16 sequences; ours is the whole-server throughput over the 16 test sequences, and the last two columns divide it by each method's maximum.}
\label{tab:cmp_speed}
\end{table*}

\begin{table*}[t]
\centering
\small
\setlength{\tabcolsep}{5pt}
\begin{tabular}{llrrr}
\toprule
Format & Default setting & Encode time (s) & Modelled FPS (1 GPU) & Modelled FPS (4 GPUs) \\
\midrule
SPZ & v4 & 0.058 & 11.71 & 46.83 \\
PlayCanvas compressed PLY & compressed PLY & 0.286 & 11.71 & 46.83 \\
ksplat & level 1 & 0.279 & 11.71 & 46.83 \\
FCGS & $\lambda{=}10^{-4}$ & 4.202 & 0.24 & 0.95 \\
Draco & default & 0.072 & 11.71 & 46.83 \\
PlayCanvas SOG & default & 1.882 & 11.71 & 46.83 \\
\midrule
FCGS, encoding excluded & -- & 0 & 11.71 & 46.83 \\
\bottomrule
\end{tabular}
\caption{\textbf{Even granting 192 parallel CPU encoders and ideal overlap, format pipelines still send at least \BudgetFormatMinMbps\,Mbit/s, above a 100\,Mbit/s downlink.} Modelled throughput of each format of \cref{tab:cmp_codecs} at its default setting from recorded encoding times, ideal stage overlap and GPU scaling, and modelled head and pruning costs.}
\label{tab:cmp_format_speed}
\end{table*}

\paragraph{Reported operating points and feature coding.}
\Cref{tab:cmp_lit} lists streaming volumetric-video methods at the configuration their papers feature, and \cref{tab:cmp_human_speed} lists digital-human methods. \Cref{tab:cmp_same_gpu} times CodecSplat and HiReFF on an idle RTX~5090 with our inputs. CodecSplat runs its released DL3DV eight-view model with its official test configuration on the same eight input views and soft masks that our reconstructor receives, and is scored exactly like our rows. Of masked and unmasked inputs at $512\times512$ and $256\times256$, unmasked 256 scores highest at its top rate and is reported. We discard the Gaussians of pixels outside the input alpha, since its background Gaussians would otherwise cover the performer. \Cref{tab:cmp_feature} compares its top rate with our tiers.
\begin{table*}[t]
\centering
\small
\setlength{\tabcolsep}{5pt}
\begin{tabular}{llll}
\toprule
Method & Time per frame & Size per frame & Dataset \\
\midrule
\textbf{Ours} & feed-forward (\ThirtyFPSMean\ frames/s, \ThirtyFPSGPUs\ GPUs) & \LiteKBCmp--\UltraKBCmp\,kB & DNA-Rendering \\
\midrule
3DGStream~\cite{3dgstream} & 12\,s & 7.6\,MB & N3DV \\
HiCoM~\cite{hicom} & 6.7\,s & 0.7\,MB & N3DV \\
QUEEN-l~\cite{queen} & 7.9\,s & 0.75\,MB & N3DV \\
Instant Gaussian Stream~\cite{instantgs} & 2.67\,s & 7.90\,MB & N3DV \\
4DGC~\cite{fourdgc} & -- & 0.5\,MB & N3DV \\
MoRGS-l~\cite{morgs} & 4.0\,s & 3.80\,MB & N3DV \\
QuARC-GS~\cite{quarcgs} & 5.0\,s & 36\,kB & N3DV \\
ReCon-GS~\cite{recongs} & 6.4\,s & 0.40\,MB & N3DV \\
ComGS-s~\cite{comgs} & 37\,s & 0.049\,MB & N3DV \\
iFVC~\cite{ifvc} & 0.22\,min & 0.09\,MB & N3DV \\
StreamSTGS~\cite{streamstgs} & -- & 173.6\,kB & N3DV \\
StreamRF~\cite{streamrf} & 0.25\,min & 17.7\,MB & N3DV flame\_salmon \\
TeTriRF~\cite{tetrirf} & 0.65\,min & 62.5\,kB & DyNeRF \\
V$^3$~\cite{vcubed} & 0.82\,min & 0.532\,MB & ReRF (Kpop) \\
JointRF~\cite{jointrf} & -- & 0.47\,MB & ReRF \\
4DGCPro~\cite{fourdgcpro} & -- & 0.64\,MB & N3DV \\
HiFi4G~\cite{hifi4g} & 4\,min & 1.648\,MB & own capture \\
DualGS~\cite{dualgs} & 12\,min & 0.323\,MB & DualGS \\
\bottomrule
\end{tabular}
\caption{\textbf{Feed-forward streaming at the capture timescale versus seconds to minutes of fitting.} Each other streaming volumetric-video method at the configuration its paper features, transcribed from that paper with its own data and hardware, where time per frame is fitting time; ours is measured (\cref{sec:exp_production}). A dash in Time per frame means that no per-frame fitting time is reported for this configuration. N3DV and DyNeRF denote the same multi-view video dataset~\cite{n3dv}.}
\label{tab:cmp_lit}
\end{table*}

\begin{table*}[t]
\centering
\small
\setlength{\tabcolsep}{4pt}
\begin{tabular}{@{}p{0.19\textwidth}p{0.12\textwidth}p{0.25\textwidth}p{0.24\textwidth}p{\dimexpr0.20\textwidth-8\tabcolsep\relax}@{}}
\toprule
Method & Input views & Setup per subject or sequence & Per-frame time & GPU \\
\midrule
\multicolumn{5}{@{}l}{\textit{Per-sequence volumetric video}} \\
\raggedright 4K4D~\cite{fourk4d} & \raggedright 54 train (DNA-Rendering; 90\% of 60 cameras) & \raggedright point clouds + sequence fit ($\sim$24 h / 200 frames) & \raggedright $\sim$7.2\,min (training)$^\dagger$ & \raggedright RTX~4090 \tabularnewline
\raggedright HumanRF~\cite{humanrf} & \raggedright 124 train (160 rig) & \raggedright occupancy grids + sequence fit ($\sim$1 day / 1000 frames) & \raggedright $\sim$86.4\,s (training; $4\times$ downscaled)$^\dagger$ & \raggedright RTX~3090 (24~GB) \tabularnewline
\raggedright LongVolCap~\cite{longvolvideo} & \raggedright 19--21 (N3DV rig) & \raggedright sequence fit ($\sim$2 h / 1200 frames) & \raggedright $\sim$6\,s (training)$^\dagger$ & \raggedright RTX~4090 \tabularnewline
\raggedright HiFi4G~\cite{hifi4g} & \raggedright 81 & \raggedright meshing, tracking + per-frame fit & \raggedright $<7$\,min (preprocessing + training) & \raggedright RTX~3090 \tabularnewline
\raggedright DualGS~\cite{dualgs} & \raggedright 81 & \raggedright joint/skin Gaussians + per-frame fit & \raggedright 12\,min & \raggedright RTX~3090 \tabularnewline
\raggedright NeuS2~\cite{neus2} & \raggedright 45--91 train (DynaCap) & \raggedright first-frame fit ($\sim$40\,s) + per-frame fit & \raggedright 20\,s (training) & \raggedright RTX~3090 \tabularnewline
\midrule
\multicolumn{5}{@{}l}{\textit{Per-subject avatar}} \\
\raggedright Animatable Gaussians~\cite{animatablegaussians} & \raggedright 16--47 & \raggedright SMPL-X + template; avatar training $\sim$2 days & \raggedright -- & \raggedright RTX~4090 \tabularnewline
\midrule
\multicolumn{5}{@{}l}{\textit{Feed-forward}} \\
\raggedright GPS-Gaussian~\cite{gpsgaussian} & \raggedright 2 (adjacent) & \raggedright none & \raggedright 25 frames/s (TensorRT; one view from two $1024^2$ inputs)$^\ddagger$ & \raggedright RTX~3090 \tabularnewline
\raggedright HiReFF~\cite{hireff} & \raggedright 4 (uncalibrated) & \raggedright none & \raggedright 3.01 frames/s (2K)$^\ddagger$ & \raggedright RTX~4090 \tabularnewline
\raggedright Tele-Aloha~\cite{telealoha} & \raggedright 4 (upper body) & \raggedright none & \raggedright 30 frames/s (2K system)$^\ddagger$ & \raggedright RTX~4090 \tabularnewline
\raggedright Tele360~\cite{tele360} & \raggedright 4--6 (unposed) & \raggedright none & \raggedright $>$25 frames/s (2K system)$^\ddagger$ & \raggedright RTX~5090 (inference) \tabularnewline
\midrule
\multicolumn{5}{@{}l}{\textit{Live mesh system}} \\
\raggedright Live4D~\cite{live4d} & \raggedright 20 (secondary) & \raggedright none & \raggedright 24 frames/s (live mesh)$^\ddagger$ & \raggedright $4\times$ RTX~3090 + $1\times$ RTX~4090 \tabularnewline
\midrule
\sys (ours) & 8 & \raggedright session setup ($\sim$\SessionReadySeconds\,s) & \raggedright \SingleFPSMean\ frames/s (1~GPU); \ThirtyFPSMean\ (\ThirtyFPSGPUs~GPUs) & RTX~5090 \tabularnewline
\bottomrule
\end{tabular}
\caption{\textbf{\sys reaches \ThirtyFPSMean\,FPS while per-sequence human reconstruction takes \HumanSeqFitMin--\HumanSeqFitMax\ per frame.} Digital-human methods as reported by their authors: ``Setup per subject or sequence'' gives training or fitting time before display, and per-frame time excludes rendering except in the system rates marked $^\ddagger$. $^\dagger$Derived from a reported total. Live4D's camera and GPU counts are from Tele-Aloha~\cite{telealoha}; our row is from \cref{fig:production}.}
\label{tab:cmp_human_speed}
\end{table*}

  \begin{table*}[t]%
  \renewenvironment{table}[1][]{}{}%
  \begin{minipage}[t]{0.485\textwidth}\begin{table}[t]
\centering
\small
\setlength{\tabcolsep}{3pt}
\begin{tabular}{@{}llrr@{}}
\toprule
Method & Input views, resolution & Sender & Receiver \\
\midrule
CodecSplat~\cite{codecsplat} & 8, 256$\times$256 & \CmpCodecSplatSendMs\,ms & \CmpCodecSplatRecvMs\,ms \\
HiReFF~\cite{hireff} & 4, 2K & \CmpHiReFFSameGPUFPS\ FPS & -- \\
\sys (ours) & 8, 512$\times$512 & \SingleFPSMean\ FPS & \CmpOursClientSameGPUMs\,ms \\
\bottomrule
\end{tabular}
\caption{\textbf{On the same RTX 5090 and DNA-Rendering inputs, our client decodes \CmpCodecSplatRecvRatio$\times$ faster than CodecSplat's receiver, and our server outpaces HiReFF.} Released models at default settings, CodecSplat at the input setting of \cref{tab:cmp_feature}. CodecSplat's sender generates and compresses features and its receiver decodes them and predicts Gaussians; our sender includes online matting, and our receiver decodes a Pro packet, runs the hypernetwork on its tokens, and generates and prunes its Gaussians. CodecSplat's times and our server rate average the 16 test scenes (best of three runs for CodecSplat); HiReFF's rate is the best of three runs on 0012\_09, 0022\_10 and 0166\_04, and our receiver time is the median over the same three sequences.}
\label{tab:cmp_same_gpu}
\end{table}
\end{minipage}\hfill
  \begin{minipage}[t]{0.485\textwidth}\begin{table}[t]
\centering
\small
\setlength{\tabcolsep}{3pt}
\resizebox{\columnwidth}{!}{%
\begin{tabular}{@{}llrrrrr@{}}
\toprule
Method & Setting & kB\,$\downarrow$ & PSNR\,$\uparrow$ & SSIM\,$\uparrow$ & LPIPS\,$\downarrow$ & Decode (ms)\,$\downarrow$ \\
\midrule
CodecSplat~\cite{codecsplat} & $\lambda{=}1024$ & 45.4 & 23.14 & 0.743 & 0.268 & 199 \\
\midrule
Ours Lite & -- & 94.8 & 26.84 & 0.870 & 0.104 & -- \\
Ours Pro & -- & 208.2 & 27.22 & 0.880 & 0.094 & 3.9 \\
Ours Ultra & -- & 321.9 & 27.27 & 0.882 & 0.092 & -- \\
\bottomrule
\end{tabular}}
\caption{\textbf{CodecSplat plateaus 3.7\,dB below our lightest tier and decodes 51$\times$ slower.} DNA-Rendering, 16 test scenes, our scoring and inputs; CodecSplat runs its released model on our eight views at its best input setting, unmasked at 256$\times$256, and its kB omits the input masks that remove its background Gaussians. Decode gives receiver time per frame on one RTX~5090, timed as in \cref{tab:cmp_same_gpu}.}
\label{tab:cmp_feature}
\end{table}
\end{minipage}%
  \end{table*}

\paragraph{Additional reconstructors.}
Pre-head coding also wins on two other pixel-aligned feed-forward reconstructors, on which we repeat the split-point analysis of \cref{sec:exp_cut}. On CodecSplat's released model built on DepthSplat~\cite{depthsplat} across the \DSScenes\ test scenes (\cref{tab:depthsplat_cut}), splitting behind a shared receiver reproduces the original outputs bit for bit. PCA-coded 64-d pre-head features reach \DSTBestPSNR\,dB at \DSTBestKB\,kB, beating every coding of the 38 post-head Gaussian attributes (best \DSABestPSNR\,dB at \DSABestOverTBestBytes$\times$ the bytes; bypassing its co-trained entropy codec yields \DSBypassPSNR\,dB vs.\ \DSLearnedTopPSNR\,dB with it). On the released stage-2 GPS-Gaussian~\cite{gpsgaussian} model, we compare 32-d shared features $F$ before its three $3\times3$ parameter heads against 8-d pre-activation rotation, scale and opacity attributes $A$, with identical lossless XYZ and RGB streams and bit-exact pass-through at $25.07$\,dB native full-image PSNR. Raw and standardised attributes retain $99.27$--$100\%$ of variance at ranks 1--8 vs.\ $98.58$--$99.90\%$ for $F$, yet pre-head coding beats raw and standardised attribute coding by $1.34$ and $0.65$\,dB on average over the shared log-rate range. It reaches $25.21$\,dB at $(r,b){=}(1,2)$ with $0.91\times$ the bytes of the best attribute setting at $(8,8)$ ($25.06$\,dB).
\begin{table*}[t]
\centering
\scriptsize
\setlength{\tabcolsep}{5pt}
\begin{tabular}{llrrrr}
\toprule
Split point / codec & Setting & kB & PSNR & SSIM & LPIPS \\
\midrule
T (PCA) & PCA $r=1$, 4-bit & 127.5 & 14.73 & 0.265 & 0.651 \\
T (PCA) & PCA $r=2$, 4-bit & 153.0 & 15.48 & 0.270 & 0.658 \\
T (PCA) & PCA $r=4$, 4-bit & 310.1 & 15.64 & 0.268 & 0.654 \\
\midrule
A (raw) & PCA $r=4$, 4-bit & 379.7 & 13.51 & 0.226 & 0.670 \\
A (raw) & PCA $r=4$, 6-bit & 1115.8 & 13.58 & 0.234 & 0.668 \\
A (raw) & PCA $r=16$, 4-bit & 1259.0 & 15.38 & 0.233 & 0.649 \\
A (raw) & PCA $r=38$, 4-bit & 2982.4 & 15.39 & 0.233 & 0.641 \\
\midrule
A (standardised) & PCA $r=4$, 4-bit & 323.2 & 13.19 & 0.205 & 0.682 \\
A (standardised) & PCA $r=4$, 6-bit & 991.8 & 13.43 & 0.219 & 0.679 \\
A (standardised) & PCA $r=16$, 4-bit & 1293.9 & 15.26 & 0.231 & 0.656 \\
A (standardised) & PCA $r=16$, 6-bit & 4341.9 & 15.30 & 0.242 & 0.650 \\
A (standardised) & PCA $r=38$, 6-bit & 10631.7 & 15.31 & 0.242 & 0.637 \\
\midrule
Learned codec & $\lambda=16$ & 74.1 & 21.38 & 0.664 & 0.350 \\
Learned codec & $\lambda=128$ & 89.1 & 22.93 & 0.737 & 0.276 \\
Learned codec & $\lambda=1024$ & 111.4 & 23.14 & 0.743 & 0.268 \\
\midrule
Uncompressed & -- & -- & 15.28 & 0.242 & 0.637 \\
\bottomrule
\end{tabular}
\caption{\textbf{On a second, pixel-aligned reconstructor, coding before the head again beats coding the Gaussian attributes.} CodecSplat's released DepthSplat-based model on the 16 DNA-Rendering test scenes, coding its 64-dimensional features before the Gaussian head (T) or the 38 attributes after it, raw or standardised (A), with our PCA codec against its own learned feature codec. Rate--quality frontiers; the bypass row decodes the uncompressed features. Rates count whole packets, including the 1-bit foreground masks of the eight input views that remove background Gaussians, which \cref{tab:cmp_feature} leaves out of CodecSplat's bytes. The released head was trained behind the learned codec, which therefore also restores the features.}
\label{tab:depthsplat_cut}
\end{table*}

\paragraph{Unseen rigs.}
\Cref{tab:cmp_rig_controls} lists every tier with plain decoding, the hypernetwork, the hypernetwork plus NSC, and NSC on plain decoding; every deployed tier improves PSNR and SSIM over standard PointSplat512 with smaller packets. \Cref{tab:cmp_codecs_external_hifi_full,tab:cmp_codecs_external_ahq_full} apply all \CmpExtFullFormatSettings\ format settings to the unseen rigs. Every deployed tier improves PSNR and SSIM over every format setting applied to the uncalibrated Gaussians and over the uncompressed input. With its whole session state spread over the evaluated frames, even Ultra is \CmpExtHiFiMinCodecOverUltraAmort$\times$ and \CmpExtAHQMinCodecOverUltraAmort$\times$ smaller than the smallest format setting on HiFi4G and ActorsHQ.
\begin{table*}[t]
\centering
\small
\setlength{\tabcolsep}{5pt}
\begin{tabular}{llrrrrr}
\toprule
Decoder / method & Tier & PSNR & SSIM & LPIPS & kB/frame & Compression ratio \\
\midrule
\multicolumn{7}{l}{\textbf{HiFi4G}} \\
Hypernetwork + NSC (deployed) & Lite & 24.148 & 0.7414 & 0.2069 & 65.0 & 39.46 \\
Hypernetwork + NSC (deployed) & Pro & 24.270 & 0.7511 & 0.1953 & 141.6 & 18.11 \\
Hypernetwork + NSC (deployed) & Ultra & 24.306 & 0.7519 & 0.1931 & 219.2 & 11.70 \\
NSC, plain decoding (control) & Lite & 23.861 & 0.7278 & 0.2329 & 65.0 & 39.46 \\
NSC, plain decoding (control) & Pro & 24.059 & 0.7423 & 0.2104 & 141.6 & 18.11 \\
NSC, plain decoding (control) & Ultra & 24.184 & 0.7478 & 0.2011 & 219.2 & 11.70 \\
Hypernetwork only (control) & Lite & 22.771 & 0.6951 & 0.2334 & 65.0 & 39.46 \\
Hypernetwork only (control) & Pro & 22.851 & 0.7008 & 0.2208 & 141.6 & 18.11 \\
Hypernetwork only (control) & Ultra & 22.832 & 0.7008 & 0.2183 & 219.2 & 11.70 \\
Plain (control) & Lite & 22.396 & 0.6792 & 0.2903 & 65.0 & 39.46 \\
Plain (control) & Pro & 22.524 & 0.6899 & 0.2529 & 141.6 & 18.11 \\
Plain (control) & Ultra & 22.670 & 0.6947 & 0.2337 & 219.2 & 11.70 \\
Standard PointSplat512 & -- & 22.810 & 0.7017 & 0.2137 & 2564.6$^*$ & 1.00 \\
\midrule
\multicolumn{7}{l}{\textbf{ActorsHQ}} \\
Hypernetwork + NSC (deployed) & Lite & 25.721 & 0.8274 & 0.1809 & 177.9 & 59.76 \\
Hypernetwork + NSC (deployed) & Pro & 25.806 & 0.8320 & 0.1703 & 404.4 & 26.29 \\
Hypernetwork + NSC (deployed) & Ultra & 25.854 & 0.8333 & 0.1680 & 628.3 & 16.92 \\
NSC, plain decoding (control) & Lite & 25.343 & 0.8172 & 0.2016 & 177.9 & 59.76 \\
NSC, plain decoding (control) & Pro & 25.632 & 0.8272 & 0.1824 & 404.4 & 26.29 \\
NSC, plain decoding (control) & Ultra & 25.798 & 0.8315 & 0.1738 & 628.3 & 16.92 \\
Hypernetwork only (control) & Lite & 24.316 & 0.8084 & 0.1943 & 177.9 & 59.76 \\
Hypernetwork only (control) & Pro & 24.370 & 0.8118 & 0.1821 & 404.4 & 26.29 \\
Hypernetwork only (control) & Ultra & 24.411 & 0.8125 & 0.1792 & 628.3 & 16.92 \\
Plain (control) & Lite & 23.683 & 0.7962 & 0.2427 & 177.9 & 59.76 \\
Plain (control) & Pro & 23.898 & 0.8057 & 0.2140 & 404.4 & 26.29 \\
Plain (control) & Ultra & 24.195 & 0.8107 & 0.1938 & 628.3 & 16.92 \\
Standard PointSplat512 & -- & 24.438 & 0.8131 & 0.1749 & 10631.7$^*$ & 1.00 \\
\bottomrule
\end{tabular}
\caption{\textbf{On both unseen rigs, every deployed tier improves PSNR and SSIM over standard PointSplat512 with far fewer bytes.} Held-out cameras and all evaluated frames of the NSC study, where ``deployed'' is the hypernetwork plus NSC. The compression ratio divides the kB/frame of standard PointSplat512 by that of each row. $^*$Raw float32 Gaussians; our packets exclude session state.}
\label{tab:cmp_rig_controls}
\end{table*}

\begin{table*}[t]
\centering
\small
\setlength{\tabcolsep}{4pt}
\begin{tabular}{llrrrrr}
\toprule
Decoder / format & Tier / setting & Packet kB & Session kB/frame & PSNR & SSIM & LPIPS \\
\midrule
Hypernetwork + NSC & Lite & 65.777 & 6.451 & 23.929 & 0.7414 & 0.2096 \\
Hypernetwork + NSC & Pro & 143.330 & 6.998 & 24.048 & 0.7510 & 0.1976 \\
Hypernetwork + NSC & Ultra & 221.623 & 8.091 & 24.123 & 0.7522 & 0.1952 \\
\midrule
Plain & Lite & 65.777 & 1.132 & 22.141 & 0.6749 & 0.2921 \\
Plain & Pro & 143.330 & 1.679 & 22.266 & 0.6846 & 0.2547 \\
Plain & Ultra & 221.623 & 2.772 & 22.406 & 0.6892 & 0.2366 \\
\midrule
Hypernetwork only & Lite & 65.777 & 6.080 & 22.494 & 0.6905 & 0.2368 \\
Hypernetwork only & Pro & 143.330 & 6.627 & 22.579 & 0.6960 & 0.2242 \\
Hypernetwork only & Ultra & 221.623 & 7.720 & 22.564 & 0.6960 & 0.2215 \\
\midrule
NSC, plain decoding & Lite & 65.777 & 1.502 & 23.648 & 0.7263 & 0.2353 \\
NSC, plain decoding & Pro & 143.330 & 2.048 & 23.884 & 0.7422 & 0.2128 \\
NSC, plain decoding & Ultra & 221.623 & 3.142 & 24.008 & 0.7476 & 0.2033 \\
\midrule
Uncompressed float32 & -- & 2487.987 & 0.000 & 22.567 & 0.6966 & 0.2176 \\
SPZ v4 & -- & 578.908 & 0.000 & 22.575 & 0.6969 & 0.2181 \\
PlayCanvas compressed PLY & -- & 724.047 & 0.000 & 22.567 & 0.6966 & 0.2177 \\
PlayCanvas SOG & WebP effort 0 & 650.106 & 0.000 & 22.566 & 0.6966 & 0.2178 \\
PlayCanvas SOG & WebP effort 4 & 606.978 & 0.000 & 22.566 & 0.6966 & 0.2178 \\
PlayCanvas SOG & WebP effort 9 & 605.583 & 0.000 & 22.566 & 0.6966 & 0.2178 \\
ksplat & level 0 & 1959.967 & 0.000 & 22.584 & 0.6970 & 0.2175 \\
ksplat & level 1 & 1073.493 & 0.000 & 22.584 & 0.6970 & 0.2176 \\
ksplat & level 2 & 1073.493 & 0.000 & 22.584 & 0.6970 & 0.2176 \\
Draco & 8 bits & 466.166 & 0.000 & 22.591 & 0.6959 & 0.2248 \\
Draco & 12 bits & 777.045 & 0.000 & 22.567 & 0.6966 & 0.2177 \\
Draco & 16 bits & 1087.836 & 0.000 & 22.567 & 0.6966 & 0.2176 \\
FCGS~\cite{fcgs} & $\lambda{=}1.6\times10^{-3}$ & 494.278 & 0.000 & 22.415 & 0.6874 & 0.2407 \\
FCGS~\cite{fcgs} & $\lambda{=}8\times10^{-4}$ & 517.825 & 0.000 & 22.495 & 0.6915 & 0.2331 \\
FCGS~\cite{fcgs} & $\lambda{=}4\times10^{-4}$ & 550.084 & 0.000 & 22.475 & 0.6914 & 0.2311 \\
FCGS~\cite{fcgs} & $\lambda{=}2\times10^{-4}$ & 588.233 & 0.000 & 22.514 & 0.6934 & 0.2271 \\
FCGS~\cite{fcgs} & $\lambda{=}10^{-4}$ & 621.509 & 0.000 & 22.557 & 0.6943 & 0.2266 \\
Uniform + Zstd & 4 bits & 495.313 & 0.000 & 22.634 & 0.6885 & 0.2440 \\
Uniform + Zstd & 6 bits & 633.680 & 0.000 & 22.574 & 0.6963 & 0.2190 \\
Uniform + Zstd & 8 bits & 752.430 & 0.000 & 22.571 & 0.6966 & 0.2177 \\
FP16 + Zstd & level 1 & 1167.392 & 0.000 & 22.567 & 0.6966 & 0.2176 \\
FP16 + Zstd & level 5 & 1139.390 & 0.000 & 22.567 & 0.6966 & 0.2176 \\
FP16 + Zstd & level 9 & 1133.906 & 0.000 & 22.567 & 0.6966 & 0.2176 \\
Draco & position 11 / attributes 8 bits & 516.179 & 0.000 & 22.571 & 0.6966 & 0.2178 \\
PlayCanvas SOG & default & 607.919 & 0.000 & 22.566 & 0.6966 & 0.2178 \\
\bottomrule
\end{tabular}
\caption{\textbf{On HiFi4G, every deployed tier exceeds all format settings on uncalibrated Gaussians in PSNR and SSIM at smaller packet sizes.} First, middle and last evaluation frame of each session, all held-out cameras. Packet kB excludes the session state; session kB/frame spreads each decoder's PCA basis, hypernetwork update and NSC update over the evaluated frames of its own session. Formats encode the uncalibrated Gaussians.}
\label{tab:cmp_codecs_external_hifi_full}
\end{table*}

\begin{table*}[t]
\centering
\small
\setlength{\tabcolsep}{4pt}
\begin{tabular}{llrrrrr}
\toprule
Decoder / format & Tier / setting & Packet kB & Session kB/frame & PSNR & SSIM & LPIPS \\
\midrule
Hypernetwork + NSC & Lite & 180.606 & 6.709 & 25.722 & 0.8278 & 0.1809 \\
Hypernetwork + NSC & Pro & 408.830 & 7.277 & 25.793 & 0.8323 & 0.1704 \\
Hypernetwork + NSC & Ultra & 634.597 & 8.414 & 25.882 & 0.8345 & 0.1681 \\
\midrule
Plain & Lite & 180.606 & 1.177 & 23.777 & 0.7994 & 0.2401 \\
Plain & Pro & 408.830 & 1.746 & 24.035 & 0.8093 & 0.2124 \\
Plain & Ultra & 634.597 & 2.883 & 24.386 & 0.8157 & 0.1917 \\
\midrule
Hypernetwork only & Lite & 180.606 & 6.323 & 24.457 & 0.8126 & 0.1925 \\
Hypernetwork only & Pro & 408.830 & 6.891 & 24.518 & 0.8164 & 0.1796 \\
Hypernetwork only & Ultra & 634.597 & 8.028 & 24.606 & 0.8183 & 0.1764 \\
\midrule
NSC, plain decoding & Lite & 180.606 & 1.562 & 25.328 & 0.8176 & 0.2021 \\
NSC, plain decoding & Pro & 408.830 & 2.130 & 25.606 & 0.8276 & 0.1830 \\
NSC, plain decoding & Ultra & 634.597 & 3.267 & 25.849 & 0.8325 & 0.1737 \\
\midrule
Uncompressed float32 & -- & 7273.588 & 0.000 & 24.711 & 0.8203 & 0.1731 \\
SPZ v4 & -- & 1725.248 & 0.000 & 24.728 & 0.8201 & 0.1753 \\
PlayCanvas compressed PLY & -- & 2115.379 & 0.000 & 24.707 & 0.8199 & 0.1748 \\
PlayCanvas SOG & WebP effort 0 & 1827.164 & 0.000 & 24.673 & 0.8196 & 0.1745 \\
PlayCanvas SOG & WebP effort 4 & 1689.433 & 0.000 & 24.673 & 0.8196 & 0.1745 \\
PlayCanvas SOG & WebP effort 9 & 1687.178 & 0.000 & 24.673 & 0.8196 & 0.1745 \\
ksplat & level 0 & 5720.082 & 0.000 & 24.662 & 0.8194 & 0.1746 \\
ksplat & level 1 & 3128.467 & 0.000 & 24.662 & 0.8194 & 0.1746 \\
ksplat & level 2 & 3128.467 & 0.000 & 24.662 & 0.8194 & 0.1746 \\
Draco & 8 bits & 1298.538 & 0.000 & 24.678 & 0.8167 & 0.1850 \\
Draco & 12 bits & 2207.859 & 0.000 & 24.712 & 0.8203 & 0.1730 \\
Draco & 16 bits & 3117.056 & 0.000 & 24.711 & 0.8203 & 0.1730 \\
FCGS~\cite{fcgs} & $\lambda{=}1.6\times10^{-3}$ & 1315.263 & 0.000 & 24.140 & 0.8101 & 0.2020 \\
FCGS~\cite{fcgs} & $\lambda{=}8\times10^{-4}$ & 1377.575 & 0.000 & 24.358 & 0.8139 & 0.1919 \\
FCGS~\cite{fcgs} & $\lambda{=}4\times10^{-4}$ & 1486.304 & 0.000 & 24.376 & 0.8148 & 0.1879 \\
FCGS~\cite{fcgs} & $\lambda{=}2\times10^{-4}$ & 1575.558 & 0.000 & 24.457 & 0.8159 & 0.1836 \\
FCGS~\cite{fcgs} & $\lambda{=}10^{-4}$ & 1675.071 & 0.000 & 24.519 & 0.8163 & 0.1819 \\
Uniform + Zstd & 4 bits & 1409.117 & 0.000 & 24.186 & 0.7847 & 0.2284 \\
Uniform + Zstd & 6 bits & 1847.243 & 0.000 & 24.731 & 0.8198 & 0.1752 \\
Uniform + Zstd & 8 bits & 2160.051 & 0.000 & 24.713 & 0.8203 & 0.1733 \\
FP16 + Zstd & level 1 & 3393.563 & 0.000 & 24.713 & 0.8203 & 0.1731 \\
FP16 + Zstd & level 5 & 3242.388 & 0.000 & 24.713 & 0.8203 & 0.1731 \\
FP16 + Zstd & level 9 & 3208.188 & 0.000 & 24.713 & 0.8203 & 0.1731 \\
Draco & position 11 / attributes 8 bits & 1441.924 & 0.000 & 24.713 & 0.8203 & 0.1735 \\
PlayCanvas SOG & default & 1659.841 & 0.000 & 24.673 & 0.8196 & 0.1745 \\
\bottomrule
\end{tabular}
\caption{\textbf{On ActorsHQ, every deployed tier exceeds all format settings on uncalibrated Gaussians in PSNR and SSIM at smaller packet sizes.} First, middle and last evaluation frame of each session, all held-out cameras. Packet kB excludes the session state; session kB/frame spreads each decoder's PCA basis, hypernetwork update and NSC update over the evaluated frames of its own session. Formats encode the uncalibrated Gaussians.}
\label{tab:cmp_codecs_external_ahq_full}
\end{table*}

\paragraph{Synthetic X-Humans.}
\label{sec:supp_xhumans}
We evaluate the three 150-frame test sequences of X-Humans~\cite{xavatar} (00035\_test\_Take10, 00041\_test\_Take8, 00088\_test\_Take10), rendered at $1024\times1024$ ($4\times$ MSAA) from 48 fixed cameras on three elevation rings ($-12^\circ$, $6^\circ$, $24^\circ$; 16 azimuths each). Inputs are eight cameras at $45^\circ$ azimuth spacing on the $6^\circ$ ring, and all 40 remaining cameras are scored over frames 30--149 after calibration on frames 0--29 with the main protocol (\cref{sec:supp_nsc}). Pose normalisation follows the main protocol and takes its horizontal axis from the first camera's right axis.

\FloatBarrier
\section{Neural session calibration details}
\label{sec:supp_nsc}
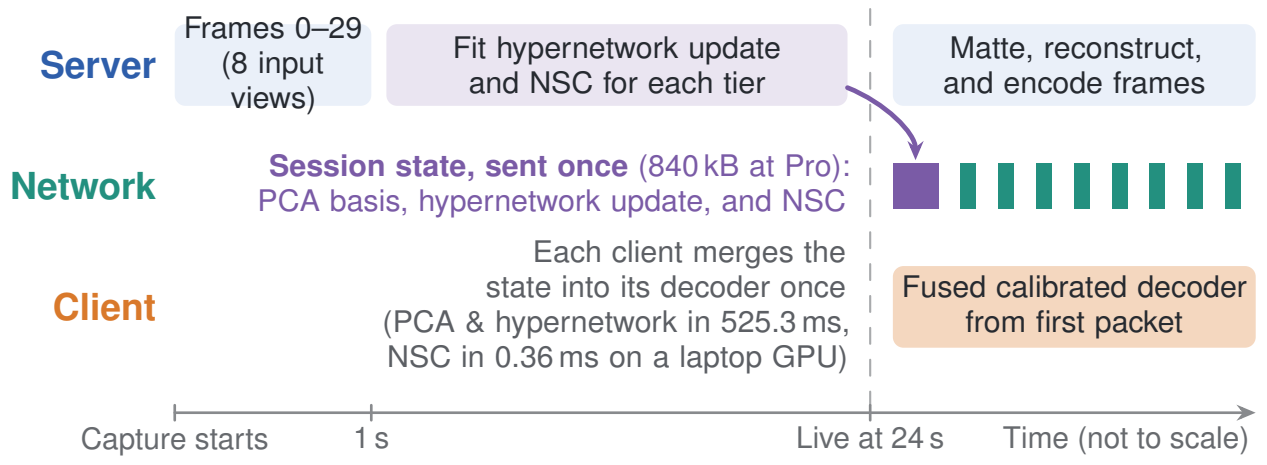
\begin{figure}[t]
\centering
\begin{tikzpicture}[hc,
  lane/.style={font=\sffamily\scriptsize\bfseries, anchor=east, inner sep=1pt},
  blk/.style={rounded corners=2pt, draw=none, font=\sffamily\fontsize{5.6}{6.6}\selectfont, align=center, inner sep=1.5pt},
]
\def\L{1.15}
\def\G{5.75}
\node[lane, text=srv] at (\L-0.08,-0.35) {Server};
\node[lane, text=pkt] at (\L-0.08,-1.15) {Network};
\node[lane, text=cli] at (\L-0.08,-1.95) {Client};
\fill[srvL, rounded corners=2pt] (\L,-0.08) rectangle (2.45,-0.62);
\node[blk, text=ink, text width=1.25cm] at ({(\L+2.45)/2},-0.35) {Frames 0--29\\(8 input views)};
\fill[ses!16, rounded corners=2pt] (2.55,-0.08) rectangle (5.6,-0.62);
\node[blk, text=ink, text width=2.9cm] at (4.075,-0.35) {Fit hypernetwork update\\and NSC for each tier};
\draw[ink!40, dashed, line width=0.4pt] (\G,0.02) -- (\G,-2.6);
\fill[srvL, rounded corners=2pt] (5.9,-0.08) rectangle (8.3,-0.62);
\node[blk, text=ink, text width=2.3cm] at (7.1,-0.35) {Matte, reconstruct,\\and encode frames};
\fill[ses] (5.9,-1.0) rectangle (6.2,-1.3);
\draw[->, draw=ses, line width=0.9pt] (5.6,-0.5) .. controls (5.85,-0.6) and (6.05,-0.8) .. (6.05,-0.98);
\foreach \x in {6.35,6.6,...,8.2} { \fill[pkt] (\x,-1.0) rectangle (\x+0.1,-1.3); }
\node[blk, text=ses, anchor=east, align=right, text width=4.4cm] at (\G-0.1,-1.15) {\textbf{Session state, sent once} (\SessionStateProKB\,kB at Pro):\\PCA basis, hypernetwork update, and NSC};
\node[blk, text=ink!75, anchor=east, align=right, text width=4.4cm] at (\G-0.1,-1.95) {Each client merges the state into its decoder once\\(PCA \& hypernetwork in \HyperFoldMs\,ms,\\NSC in \LDLaptopLoRAFold\,ms on a laptop GPU)};
\fill[cli!30, rounded corners=2pt] (5.9,-1.68) rectangle (8.3,-2.22);
\node[blk, text=ink, text width=2.3cm] at (7.1,-1.95) {Fused calibrated decoder\\from first packet};
\draw[->, draw=ink!55, line width=0.5pt] (\L,-2.65) -- (8.3,-2.65);
\foreach \x/\t in {\L/Capture starts, 2.45/1\,s, \G/Live at \SessionReadySeconds\,s} {
  \draw[ink!55, line width=0.5pt] (\x,-2.6) -- (\x,-2.7);
  \node[font=\sffamily\fontsize{5.6}{6.6}\selectfont, text=ink!70, anchor=north, inner sep=1pt] at (\x,-2.7) {\t};
}
\node[font=\sffamily\fontsize{5.6}{6.6}\selectfont, text=ink!70, anchor=north east, inner sep=1pt] at (8.3,-2.7) {Time (not to scale)};
\end{tikzpicture}
\caption{\textbf{Neural session calibration is a one-time setup step that adds zero per-frame overhead.} Analogous to camera calibration, the server fits the hypernetwork and NSC updates on the first second of capture before live streaming begins and transmits them once with the frame-zero PCA basis; each client merges them into its fused decoder and decodes the shared token stream. Server setup time is for Pro on four RTX~5090 GPUs.}
\label{fig:nsc_session}
\end{figure}

NSC is fitted once per session and tier from a one-second warm-up capture, frames 0--29 of the eight input views, before live streaming begins (\cref{fig:nsc_session}); live capture then proceeds at frame 30 with the frozen session state. Each fit runs 300 AdamW steps at learning rate $10^{-3}$ with 50 warm-up steps, weight decay $10^{-4}$ and gradient clipping at 0.5. A fit at $10^{-4}$ ends \NSCLRLossExcessMin--\NSCLRLossExcessMax\% higher in loss in all \NSCLRSessions\ DNA-Rendering sessions of the three tiers, so every session, including the unseen rigs, uses $10^{-3}$. The hypernetwork update and NSC together restore garment colour and printed detail that the generic hypernetwork blurs (\cref{fig:nsc_zoom}).
\input{fig/nsc_zoom}

\paragraph{Every tier on the unseen rigs.}
\Cref{tab:nsc_gap} extends \cref{fig:nsc_gap} to all tiers and both decoders, and \cref{tab:stats_rigs} gives every sequence with bootstrap intervals.
\begin{table}[t]
\centering
\small
\setlength{\tabcolsep}{3pt}
\resizebox{\columnwidth}{!}{%
\begin{tabular}{lrrrrrr}
\toprule
& \multicolumn{3}{c}{HiFi4G} & \multicolumn{3}{c}{ActorsHQ} \\
\cmidrule(lr){2-4}\cmidrule(lr){5-7}
Configuration & Lite & Pro & Ultra & Lite & Pro & Ultra \\
\midrule
Uncompressed, no pruning & \multicolumn{3}{c}{22.821} & \multicolumn{3}{c}{24.438} \\
Uncompressed, same pruning & \multicolumn{3}{c}{22.808} & \multicolumn{3}{c}{24.439} \\
Uncompressed, same pruning + NSC & \multicolumn{3}{c}{24.271} & \multicolumn{3}{c}{25.904} \\
\midrule
Plain & 22.396 & 22.524 & 22.670 & 23.683 & 23.898 & 24.195 \\
Hypernetwork & 22.771 & 22.851 & 22.832 & 24.316 & 24.370 & 24.411 \\
NSC, plain decoding & 23.861 & 24.059 & 24.184 & 25.343 & 25.632 & 25.798 \\
Hypernetwork + NSC (deployed) & 24.148 & 24.270 & 24.306 & 25.721 & 25.806 & 25.854 \\
\bottomrule
\end{tabular}}
\caption{\textbf{On unseen rigs, the codec costs little with the hypernetwork and session calibration brings large gains.} PSNR on held-out cameras over all frames after the one-second warm-up. ``Same pruning'' applies our token and client pruning without the codec, and ``no pruning'' applies neither. NSC is fitted separately for each tier on the input views, and without the codec for the uncompressed + NSC row.}
\label{tab:nsc_gap}
\end{table}

\begin{table*}[t]
\centering
\small
\begin{tabular*}{\textwidth}{@{\extracolsep{\fill}}llrrrrr@{}}
\toprule
Sequence & Tier & PSNR & $\Delta_{\mathrm{unc}}$ & $\Delta_{\mathrm{hyper}}$ & $n_f^+/n_f$ & $n_c^+/n_c$ \\
\midrule
Greeting & Lite & 23.667 & $+0.875\,[+0.713,+1.054]$ & $+1.037\,[+0.857,+1.212]$ & 120/120 & 8/8 \\
Greeting & Pro & 23.845 & $+1.053\,[+0.897,+1.209]$ & $+1.014\,[+0.871,+1.152]$ & 120/120 & 8/8 \\
Greeting & Ultra & 23.908 & $+1.115\,[+0.979,+1.264]$ & $+1.111\,[+0.955,+1.266]$ & 120/120 & 8/8 \\
\midrule
Umbrella & Lite & 24.101 & $+2.060\,[+1.665,+2.417]$ & $+2.101\,[+1.673,+2.492]$ & 120/120 & 8/8 \\
Umbrella & Pro & 24.246 & $+2.205\,[+1.820,+2.549]$ & $+2.187\,[+1.795,+2.538]$ & 120/120 & 8/8 \\
Umbrella & Ultra & 24.282 & $+2.241\,[+1.830,+2.611]$ & $+2.248\,[+1.831,+2.624]$ & 120/120 & 8/8 \\
\midrule
Changing clothes & Lite & 24.675 & $+1.083\,[+0.853,+1.336]$ & $+0.993\,[+0.833,+1.181]$ & 120/120 & 8/8 \\
Changing clothes & Pro & 24.720 & $+1.128\,[+0.900,+1.382]$ & $+1.058\,[+0.867,+1.284]$ & 120/120 & 8/8 \\
Changing clothes & Ultra & 24.728 & $+1.137\,[+0.905,+1.393]$ & $+1.064\,[+0.856,+1.305]$ & 120/120 & 8/8 \\
\midrule
Actor01 & Lite & 26.000 & $+1.545\,[+1.084,+1.992]$ & $+1.685\,[+1.285,+2.077]$ & 119/119 & 8/8 \\
Actor01 & Pro & 26.121 & $+1.666\,[+1.198,+2.123]$ & $+1.758\,[+1.311,+2.187]$ & 119/119 & 8/8 \\
Actor01 & Ultra & 26.180 & $+1.725\,[+1.276,+2.161]$ & $+1.750\,[+1.323,+2.167]$ & 119/119 & 8/8 \\
\midrule
Actor03 & Lite & 25.441 & $+1.018\,[+0.669,+1.367]$ & $+1.125\,[+0.788,+1.456]$ & 112/112 & 8/8 \\
Actor03 & Pro & 25.491 & $+1.068\,[+0.742,+1.388]$ & $+1.115\,[+0.795,+1.420]$ & 112/112 & 8/8 \\
Actor03 & Ultra & 25.527 & $+1.105\,[+0.787,+1.412]$ & $+1.136\,[+0.818,+1.442]$ & 112/112 & 8/8 \\
\bottomrule
\end{tabular*}
\caption{\textbf{NSC improves PSNR on every unseen-rig sequence and tier.} PSNR of the deployed tiers and their differences from our uncompressed reconstruction ($\Delta_{\mathrm{unc}}$) and from the hypernetwork without NSC ($\Delta_{\mathrm{hyper}}$), with 95\% intervals from a crossed bootstrap that resamples frames and cameras independently (10{,}000 draws); $n_f^+/n_f$ and $n_c^+/n_c$ count frames and held-out cameras on which NSC raises PSNR.}
\label{tab:stats_rigs}
\end{table*}

\paragraph{Calibration targets.}
On Pro restored tokens (\cref{tab:nsc_targets}), updating all 1.31\,M parameters of the three per-slot MLP layers overfits the 30-frame calibration window, costing \NSCProFullMLPHeldDrop\,dB on held-out views and \NSCProFullMLPTempDrop\,dB over subsequent frames. Removing the LPIPS term from the calibration loss degrades LPIPS by \NSCProNoLPIPSWorse. On HiFi4G and ActorsHQ, a 12-parameter colour-affine transform changes PSNR by \NSCColorHiFiPro\ and \NSCColorAHQPro\,dB, whereas our rank-5 per-slot MLP update (\NSCKB\,kB) reaches \NSCVsAdapterHiFi\% and \NSCVsAdapterAHQ\% of the gain of an equally sized per-frame feature adapter at zero per-frame cost. With the eight views split over four RTX~5090 GPUs, Pro goes live \SessionReadySeconds\,s after capture starts, counting the one second of capture, the frame-zero PCA, the hypernetwork update and NSC; a laptop client merges the PCA basis and hypernetwork update in \HyperFoldMs\,ms and NSC in \LDLaptopLoRAFold\,ms.
\begin{table*}[t]
\centering
\small
\setlength{\tabcolsep}{5pt}
\begin{tabular}{lrrrrrr}
\toprule
& & & \multicolumn{2}{c}{Held-out views} & \multicolumn{2}{c}{Subsequent frames} \\
\cmidrule(lr){4-5}\cmidrule(lr){6-7}
Calibration target / loss & Params & kB & PSNR & LPIPS & PSNR & LPIPS \\
\midrule
Per-slot MLP, $r{=}5$ (deployed) & 20{,}480 & 41.0 & 25.944 & 0.0916 & 24.827 & 0.1022 \\
Per-slot MLP, $r{=}2$ & 8{,}192 & 16.4 & 25.945 & 0.0909 & 24.865 & 0.1007 \\
Per-slot MLP, $r{=}8$ & 32{,}768 & 65.5 & 25.937 & 0.0916 & 24.775 & 0.1027 \\
Per-slot MLP, $r{=}32$ & 131{,}072 & 262.1 & 25.907 & 0.0930 & 24.636 & 0.1059 \\
Full per-slot MLP & 1.31\,M & 2625.5 & 25.816 & 0.0963 & 24.475 & 0.1115 \\
\midrule
Output layers only, $r{=}5$ & 25{,}670 & 51.3 & 25.829 & 0.0931 & 24.912 & 0.0991 \\
Per-slot MLP $r{=}5$ + all biases & 38{,}926 & 77.9 & 25.940 & 0.0915 & 24.794 & 0.1025 \\
Per-slot MLP + outputs, $r{=}5$ & 46{,}150 & 92.3 & 25.938 & 0.0917 & 24.806 & 0.1023 \\
\midrule
$L_1$ loss only (without SSIM and LPIPS) & 20{,}480 & 41.0 & 26.008 & 0.1070 & 24.771 & 0.1210 \\
Without LPIPS ($0.8L_1 + 0.2(1{-}\mathrm{SSIM})$) & 20{,}480 & 41.0 & 26.052 & 0.1034 & 24.869 & 0.1166 \\
\bottomrule
\end{tabular}
\caption{\textbf{On Pro tokens, a rank-5 update of the three per-slot MLP layers with perceptual supervision avoids full-MLP overfitting and perceptual degradation.} Fixed 300-step fits on frames 0--29 of the eight input views; kB counts FP16 parameter bytes. Held-out views scores the 52 non-input cameras at frame 30 on six scenes; subsequent frames scores eight held-out cameras over frames 40--89 of 0012\_09, 0022\_10 and 0166\_04.}
\label{tab:nsc_targets}
\end{table*}

\paragraph{Warm-up budget.}
\Cref{tab:nsc_budget} varies the warm-up frames $F$ and steps $S$ for NSC on plain decoding. A single frame with \NSCBudgetOneFrameSteps\ steps already recovers \NSCBudgetOneFrameShareHiFi\ and \NSCBudgetOneFrameShareAHQ\% of the full gain on HiFi4G and ActorsHQ. Deployment uses $F{=}\NSCBudgetFrames$ and $S{=}\NSCBudgetSteps$ as a universal zero-initialised protocol that needs no prior data from the rig.

\paragraph{The training domain.}
On the 16 DNA-Rendering test scenes, NSC adds \NSCDNADeltaLearnedLite/\NSCDNADeltaLearnedPro/\NSCDNADeltaLearnedUltra\,dB to the hypernetwork at Lite/Pro/Ultra, improving PSNR on all 16 scenes at every tier as well as SSIM and LPIPS (\cref{tab:nsc_dna}), so DNA-Rendering sessions deploy it as well.
\begin{table}[t]
\centering
\small
\setlength{\tabcolsep}{5pt}
\begin{tabular}{rccc}
\toprule
Warm-up frames $F$ & $S{=}30$ & $S{=}100$ & $S{=}300$ \\
\midrule
1 & 0.87 / 0.82 & 1.16 / 1.47 & 1.20 / 1.53 \\
3 & 0.87 / 0.86 & 1.14 / 1.50 & 1.27 / 1.61 \\
10 & 0.90 / 0.88 & 1.19 / 1.50 & 1.36 / 1.65 \\
30 & 0.98 / 1.05 & 1.36 / 1.46 & 1.50 / 1.63 \\
\bottomrule
\end{tabular}
\caption{\textbf{A single warm-up frame already recovers \NSCBudgetOneFrameShareHiFi\ and \NSCBudgetOneFrameShareAHQ\% of the full calibration gain.} Pro PSNR gain over the plain stream (dB, HiFi4G / ActorsHQ) for a rank-5 NSC fitted on the first $F$ frames for $S$ steps; this budget study fits one NSC on plain decoding for all three tiers, whereas deployed sessions fit one per tier. The full gain is that of the one-second warm-up ($F{=}\NSCBudgetFrames$, $S{=}\NSCBudgetSteps$).}
\label{tab:nsc_budget}
\end{table}

  \begin{table*}[t]%
  \renewenvironment{table}[1][]{}{}%
  \begin{minipage}[t]{0.485\textwidth}\begin{table}[t]
\centering
\small
\setlength{\tabcolsep}{4pt}
\begin{tabular}{llrrr}
\toprule
Tier & Decoder & PSNR & SSIM & LPIPS \\
\midrule
Lite & Plain & 24.950 & 0.8198 & 0.1809 \\
Lite & Hypernetwork & 25.772 & 0.8491 & 0.1239 \\
Lite & Hypernetwork + NSC & 26.840 & 0.8700 & 0.1043 \\
Pro & Plain & 25.582 & 0.8419 & 0.1459 \\
Pro & Hypernetwork & 26.065 & 0.8598 & 0.1121 \\
Pro & Hypernetwork + NSC & 27.219 & 0.8802 & 0.0941 \\
Ultra & Plain & 25.904 & 0.8528 & 0.1256 \\
Ultra & Hypernetwork & 26.095 & 0.8614 & 0.1097 \\
Ultra & Hypernetwork + NSC & 27.273 & 0.8818 & 0.0923 \\
\midrule
Pro & NSC, plain decoding & 26.758 & 0.8697 & 0.1091 \\
\bottomrule
\end{tabular}
\caption{\textbf{NSC adds \NSCDNADeltaLearnedLite/\NSCDNADeltaLearnedPro/\NSCDNADeltaLearnedUltra\,dB on top of the hypernetwork at Lite/Pro/Ultra.} Identical packets on the 16 test scenes; an NSC update per sequence and tier is fitted on frames 0--29 of the eight input cameras; per-frame bytes are unchanged, and each update is a one-time \NSCKB\,kB download.}
\label{tab:nsc_dna}
\end{table}
\end{minipage}\hfill
  \begin{minipage}[t]{0.485\textwidth}\begin{table}[t]
\centering
\small
\setlength{\tabcolsep}{3pt}
\begin{tabular}{lrrrrrr}
\toprule
Tier & kB & Plain & Generic & \shortstack{+ hyper-\\network\\update} & \shortstack{Gain\\(dB)} & \shortstack{Recovered\\(\%)} \\
\midrule
Lite & 94.8 & 24.950 & 25.343 & 25.772 & +0.822 & 66.7 \\
Pro & 208.2 & 25.582 & 25.815 & 26.065 & +0.483 & 80.4 \\
Ultra & 321.9 & 25.904 & 25.971 & 26.095 & +0.191 & 68.5 \\
\bottomrule
\end{tabular}
\caption{\textbf{Both the pretrained generic hypernetwork and the hypernetwork update recover coding loss at zero per-frame bytes.} PSNR on the 16 test scenes of identical packets decoded plainly, with the generic hypernetwork alone, and with its one-time hypernetwork update, where Recovered is the share of the coding loss (uncompressed with matched pruning minus plain stream) recovered.}
\label{tab:learned_decoding}
\end{table}
\end{minipage}%
  \end{table*}

\section{Implementation details}
\label{sec:supp_handoff}
\paragraph{Hypernetwork.}
\Cref{tab:learned_decoding} separates plain PCA decoding, the pretrained generic hypernetwork and its one-time hypernetwork update across all three tiers on DNA-Rendering. The generic weights alone recover \HyperGenericRecoveredLite/\HyperGenericRecoveredPro/\HyperGenericRecoveredUltra\% of the coding loss at Lite/Pro/Ultra, and the hypernetwork update raises this to \LDRecoveredLite/\LDRecoveredPro/\LDRecoveredUltra\%, adding \HyperSessionDeployedLite/\HyperSessionDeployedPro/\HyperSessionDeployedUltra\,dB even with NSC fitted on each decoder. Controlled ablations on the 16 test scenes support both spatial context and head-metric supervision. Adding the 8-nearest-neighbour context improves PSNR by \MechNeighGainLitedB/\MechNeighGainProdB\,dB at Lite/Pro, and randomly permuting neighbour correspondence erases that gain (\MechPermGainLitedB/\MechPermGainProdB\,dB below spatial neighbours). Supervising in the head metric $W^\top W$ explains \MechMetricExpGainLite/\MechMetricExpGainPro\ percentage points more head-space residual energy than unweighted token MSE and improves LPIPS by \MechMetricLPIPSDeltaLite/\MechMetricLPIPSDeltaPro. The generic hypernetwork trains for 30{,}000 steps with AdamW at peak learning rate $3\times10^{-4}$ (1000 warm-up steps, cosine decay to $3\times10^{-5}$, weight decay 0.01) on batches of 4096 tokens from four scenes, across ten rank and bit-depth settings on \HyperTrainScenes\ training scenes and validated on \HyperValActors\ held-out actors. The hypernetwork update trains 1000 steps at learning rate $10^{-3}$ on frames 0--29 in \HyperFitSecondsMin--\HyperFitSecondsMax\,s per tier on one RTX~5090. In FP16 on the laptop, the hypernetwork preserves the exact FP32 selection of the 20\% pruned Gaussians across all \LaptopFPCheckFrames\ decoded frames and changes the recovered share of the coding loss by at most \LaptopFPExplainedMax.

\paragraph{Timing.}
BiRefNet Lite is fine-tuned at 512 from the official Lite checkpoint on mask-annotated DNA-Rendering data and runs with backbone CUDA graphs at batch size four. Alpha is quantised to 8 bits and multiplied by the area-averaged valid region before reconstruction. Three producer processes per GPU share it through a private CUDA MPS context, and every multi-GPU server uses this setting. Frames are replayed from memory, already undistorted. Steady throughput counts source frames 36--131 as $96/(t_{131}-t_{35})$, where $t$ is the time a packet is ready, taking the best of three runs per scene and averaging over scenes. Throughput does not depend on the tier: matting and reconstruction dominate each frame, and rank-48 (Pro) and rank-80 coding run at a \ServerPairedRatio\ throughput ratio on the same GPUs and scenes. Client decoding, the hypernetwork, the head and pruning are timed from a packet in memory to the Gaussians passed to the renderer, on the real packets of \ClientRealScenes\ test sequences (\ClientRealTokensMean\ token groups per Pro packet on average), with the hypernetwork in FP16 and the head in BF16. Medians over all timed passes are \LDLaptopClientLite/\LDLaptopClientPro/\LDLaptopClientUltra\,ms at Lite/Pro/Ultra on the laptop RTX~3050~Ti with a fused Triton 8-nearest-neighbour kernel (\LDLaptopDecodePro/\LDLaptopOverheadPro/\LDLaptopHeadPro\,ms for the median Pro frame), and \CmpOursClientSameGPUMs\,ms at Pro on an RTX~5090 with the Torch voxel search selected at startup.

\paragraph{Live broadcast.}
In 64-second (1{,}920-frame) live broadcasts at 30\,FPS after a 30-frame warm-up, five RTX~5090 GPUs stream Pro packets over Wi-Fi to the laptop RTX~3050~Ti with zero drift. End-to-end latency from capture to rendered frame averages \GoLiveRenderMinMs--\GoLiveRenderMaxMs\,ms across the three sequences: \GoLiveServerMinMs--\GoLiveServerMaxMs\,ms from capture to transmission on the server, \GoLiveWiFiMinMs--\GoLiveWiFiMaxMs\,ms Wi-Fi transfer, \GoLiveClientMinMs--\GoLiveClientMaxMs\,ms client decoding to Gaussians and \GoLiveGsplatMinMs--\GoLiveGsplatMaxMs\,ms \texttt{gsplat}~\cite{gsplat} rasterisation. Replacing BiRefNet Lite with the video matting network RVM~\cite{rvm} and refitting the session state on its mattes lowers Pro's PSNR on all 16 test sequences, by \MatRVMProLoss\,dB on average.

\paragraph{Reconstructor choice.}
We build on PointSplat~\cite{pointsplat} because it is a feed-forward reconstructor designed for multi-view human capture whose per-token Gaussian head gives a natural split point. Its Point-Image Transformer fuses all input views into one shared token set around a coarse point proxy of the performer, each token driving 16 Gaussian slots, so each spatial token is coded once regardless of how many views are fused, unlike pixel-aligned reconstructors that emit a Gaussian map for every input view. At eight views it also leads the feed-forward reconstructors of \cref{tab:cmp_feature,tab:cmp_feedforward} with \FFPointPSNR\,dB, against \CmpCodecSplatTopPSNRFine\,dB for CodecSplat (\cref{tab:cmp_feature}) and \FFNewMinPSNR--\FFBestNewPSNR\,dB for GPS-Gaussian, DA3 and WorldMirror~2.0.

\paragraph{Stateless streaming.}
Offline replay of the real packets shows four streaming properties of stateless coding. (i)~A viewer joining at frame 30, 60 or 89 downloads a constant 1.05--1.06\,MB at Pro (0.84\,MB session state plus one packet). Stateful streams accumulate every prior update, reaching 11.7/251.5/483.3\,MB for 3DGStream, 22.9/114.4/204.0\,MB for HiCoM and 127.8/1710.6/3237.8\,MB for QUEEN-l across the 16 scenes. (ii)~Losing frame 45 on 0012\_09 affects only that frame for Pro, which recovers bit-exact rendering at frame 46. 3DGStream's relative neural transformation cache propagates the error across all 25 subsequent frames (2.94\,dB below its clean stream at frame 70). (iii)~With the three merged per-tier decoders resident on the client (137\,MB total), switching tiers every 10 frames or randomly per frame over frames 30--89 of 0012\_09 and 0022\_10 decodes Gaussians bit-identical to each tier's fixed stream: an element-wise comparison of all decoded means, scales, rotations, opacities and colours finds no differing frame and a maximum absolute difference of 0, and the rendered PSNR and LPIPS of the 10-frame schedule equal those of the fixed streams exactly. (iv)~In trace-driven 60\,s FIFO link simulations at 30\,FPS with a 200\,ms deadline (step, square-wave and random-walk bandwidths), causal per-frame tier adaptation reduces the share of frames missing the deadline from 74.9\% to 5.0\% on the step trace. Clearing queued unsent frames once the oldest unsent frame exceeds the deadline and sending the latest frame lowers the share of skipped or late frames to 0.85\% (0.67\% skipped, 29.8\,FPS delivered). All three per-sequence baselines miss 100\% of deadlines even at a constant 100\,Mbit/s.

\section{Feed-forward reconstructors}
\label{sec:supp_ff}
\begin{table*}[t]
\centering
\small
\setlength{\tabcolsep}{6pt}
\begin{tabular}{@{}lrrrrrrr@{}}
\toprule
Method & Gaussians & kB\,$\downarrow$ & Mbit/s\,$\downarrow$ & FPS\,$\uparrow$ & PSNR\,$\uparrow$ & SSIM\,$\uparrow$ & LPIPS\,$\downarrow$ \\
\midrule
GPS-Gaussian~\cite{gpsgaussian} & 533k & 29{,}860.3 & 7{,}166.5 & 34.6 & 18.38 & 0.602 & 0.297 \\
DA3~\cite{da3} & 176k & 26{,}714.3 & 6{,}411.4 & 6.6 & 18.98 & 0.581 & 0.342 \\
WorldMirror~2.0~\cite{worldmirror2} & 84k & 4{,}731.8 & 1{,}135.6 & 6.9 & 20.47 & 0.631 & 0.302 \\
PointSplat512~\cite{pointsplat} & 85k & 4{,}753.1 & 1{,}140.8 & -- & 26.17 & 0.865 & 0.103 \\
\midrule
Ours Lite & \textbf{65k} & \textbf{94.8} & \textbf{22.8} & -- & 26.84 & 0.870 & 0.104 \\
Ours Pro & \textbf{65k} & 208.2 & 50.0 & \ThirtyFPSMean & 27.22 & 0.880 & 0.094 \\
Ours Ultra & \textbf{65k} & 321.9 & 77.3 & -- & \textbf{27.27} & \textbf{0.882} & \textbf{0.092} \\
\bottomrule
\end{tabular}
\caption{\textbf{Every tier of \sys exceeds GPS-Gaussian, DA3, WorldMirror~2.0 and PointSplat512 in PSNR and SSIM with at least \FFMinBytesRatio$\times$ fewer bytes.} DNA-Rendering, 16 test scenes, frame 30, the eight input views and scoring of \cref{tab:official_quality_full}; kB counts each frame's Gaussians as uncompressed float32 for the reconstructors and the packet for our tiers. Mbit/s is the payload rate at 30 frames/s; FPS gives four RTX~5090 GPUs, $4/t$ from the reconstructors' one-GPU inference time $t$ with inputs and alpha already on the GPU, and our server throughput.}
\label{tab:cmp_feedforward}
\end{table*}

\input{fig/ffqual}
General-purpose reconstructors trained on scenes or other capture rigs lose the performer's detail at eight views (\cref{tab:cmp_feedforward}) and blur, ghost or fracture the face and limbs (\cref{fig:ff_qual}): the strongest of the three reconstructors added in \cref{tab:cmp_feedforward}, \FFBestNewName, reaches \FFBestNewPSNR\,dB, \FFLiteOverBestNewDB\,dB below our lightest tier at \FFBestNewBytesOverPro$\times$ the bytes of Pro. We run the released models of three further feed-forward reconstructors on the eight input views, soft masks and scoring of \cref{tab:official_quality_full}: Depth Anything~3~\cite{da3} with its Gaussian head, WorldMirror~2.0~\cite{worldmirror2} and GPS-Gaussian~\cite{gpsgaussian}. DA3 and WorldMirror~2.0 receive the calibrated cameras in their own conventions and are aligned to the capture frame through the eight input cameras; WorldMirror~2.0 unprojects its predicted depth with these cameras through its released \texttt{gsdepth+gtcamera} path. GPS-Gaussian forms disjoint stereo pairs from the eight input cameras by calibrated geometry, rectifies them at its native $1024^2$ resolution with the principal points placing a reference point of the calibrated optical axes at zero disparity, and merges their Gaussians. As for CodecSplat, DA3 and WorldMirror~2.0 run on masked and unmasked inputs and report the variant with the higher scene-averaged PSNR, and GPS-Gaussian takes masked inputs as in its release. Every reconstructor keeps the Gaussians inside the input alpha.
The feed-forward clips of the supplementary video (0811\_06 and 0438\_02) label each method with its frame-30 kB/frame and PSNR on that sequence from the evaluations of \cref{tab:cmp_dyn_more_ours_b,tab:cmp_feature,tab:cmp_feedforward}, and GPS-Gaussian, DA3, WorldMirror~2.0 and Pro with their FPS from \cref{tab:cmp_feedforward}. PointSplat512 is the reconstructor our server runs, so its clips carry our \ThirtyFPSMean\,FPS four-GPU rate; CodecSplat needs both its sender and its receiver to produce Gaussians and is labelled `$<$\,\FFCodecSplatFPSBound\,FPS', since four divided by its \CmpCodecSplatSendMs\,+\,\CmpCodecSplatRecvMs\,ms sender plus receiver time on one RTX~5090 gives \FFCodecSplatFourGPUFPS\,FPS (\cref{tab:cmp_same_gpu}). The other clip labels (kB/frame, PSNR in dB on 0811\_06 / 0438\_02) are: \FFClipLabels.

\paragraph{Inputs on one camera ring.}
GPS-Gaussian matches neighbouring views by stereo, and Tele360~\cite{tele360} evaluates it with inputs sampled uniformly from one camera ring. We therefore also take the eight inputs from one ring at 45$^\circ$ spacing, where GPS-Gaussian merges the eight cyclic adjacent pairs, and rerun GPS-Gaussian, PointSplat512, and all three tiers with their session state on the same 16 scenes. \Cref{tab:cmp_ring8} shows that \sys keeps its lead: Pro reaches \RingProPSNR\,dB, \RingProOverPointDB\,dB above PointSplat512 with \RingPointBytesOverPro$\times$ fewer bytes and \RingProOverGPSDB\,dB above GPS-Gaussian.
\begin{table}[t]
\centering
\small
\setlength{\tabcolsep}{4pt}
\begin{tabular}{@{}lrrrr@{}}
\toprule
Method & kB\,$\downarrow$ & PSNR\,$\uparrow$ & SSIM\,$\uparrow$ & LPIPS\,$\downarrow$ \\
\midrule
GPS-Gaussian~\cite{gpsgaussian} & 95{,}757.8 & 16.52 & 0.493 & 0.375 \\
PointSplat512~\cite{pointsplat} & 5{,}495.7 & 26.58 & 0.870 & 0.099 \\
\midrule
Ours Lite & \textbf{94.6} & 27.12 & 0.876 & 0.099 \\
Ours Pro & 207.4 & 27.51 & 0.886 & 0.089 \\
Ours Ultra & 320.4 & \textbf{27.59} & \textbf{0.888} & \textbf{0.087} \\
\bottomrule
\end{tabular}
\caption{\textbf{With the eight inputs on one camera ring, every tier of \sys still exceeds GPS-Gaussian and PointSplat512 in PSNR and SSIM.} DNA-Rendering, 16 test scenes, frame 30, cameras 1, 7, \ldots, 43 as inputs; scoring and kB as in \cref{tab:cmp_feedforward}.}
\label{tab:cmp_ring8}
\end{table}

\end{document}